\documentclass[10pt, a4paper]{article}
\usepackage[dvipdfmx]{}
\usepackage{lineno,hyperref,dsfont,mathtools}
\hypersetup{ colorlinks=true,urlcolor = blue,linktocpage=true}
\usepackage[ruled,lined]{algorithm2e}
\usepackage{cuted}
\usepackage{graphicx}
\usepackage{bm}
\usepackage{multicol}
\usepackage{caption}
\usepackage{subcaption}
\usepackage{tikz}  
\usetikzlibrary{spy,calc}
\usepackage{pgfplots}
\usepackage{amsmath}
\usepackage{tabularx} 
\usepackage{multirow}
\usepackage{adjustbox}
\usepackage{amsmath,amsfonts,amssymb}
\usepackage{authblk}

\usepackage{geometry}
\hypersetup{
    colorlinks=true      
}

\newcommand{\matr}[1]{\mathbf{#1}}

\newcolumntype{C}{ >{\centering\arraybackslash} }

\date{}
\begin{document}
\title{Reflectance Intensity Assisted Automatic and Accurate Extrinsic Calibration of 3D LiDAR and Panoramic Camera Using a Printed Chessboard}

\author[1]{Weimin Wang\thanks{Corresponding Author: weimin@ucl.nuee.nagoya-u.ac.jp}}                   
\author[1]{Ken Sakurada\thanks{sakurada@nagoya-u.jp}}           
\author[1]{Nobuo Kawaguchi\thanks{kawaguti@nagoya-u.jp}}           %

\affil[1]{Nagoya University, Japan}       
\maketitle





\begin{abstract}
This paper presents a novel method for fully automatic and convenient extrinsic calibration of a 3D LiDAR and a panoramic camera with a normally printed chessboard. The proposed method is based on the 3D corner estimation of the chessboard from the sparse point cloud generated by one frame scan of the LiDAR. To estimate the corners, we formulate a full-scale model of the chessboard and fit it to the segmented 3D points of the chessboard. The model is fitted by optimizing the cost function under constraints of correlation between the reflectance intensity of laser and the color of the chessboard's patterns. Powell's method is introduced for resolving the discontinuity problem in optimization. The corners of the fitted model are considered as the 3D corners of the chessboard. Once the corners of the chessboard in the 3D point cloud are estimated, the extrinsic calibration of the two sensors is converted to a 3D-2D matching problem. The corresponding 3D-2D points are used to calculate the absolute pose of the two sensors with Unified Perspective-n-Point (UPnP). Further, the calculated parameters are regarded as initial values and are refined using the Levenberg-Marquardt method. The performance of the proposed corner detection method from the 3D point cloud is evaluated using simulations. The results of experiments, conducted on a Velodyne HDL-32e LiDAR and a Ladybug3 camera under the proposed re-projection error metric, qualitatively and quantitatively  demonstrate the accuracy and  stability of the final extrinsic calibration parameters.
\end{abstract}








\section{Introduction}
A combination of the Light Detection And Ranging (LiDAR) sensor and the panoramic camera has been widely utilized for deriving the benefits of color as well as depth information. Two typical examples in which the combination is used are 3D mapping and model generation, which use the LiDAR sensor and the color and the texture information of images  \cite{Mastin_2009,paparoditis2012stereopolis}, and the improvement of the pedestrian detection accuracy in images using the distance information obtained from the LiDAR~\cite{Szarvas,Premebida_2013,Schlosser_2016}.  The first and critical step for fusing multi-modal data from the two devices is the accurate and convenient extrinsic calibration. 

The process of the extrinsic calibration between the LiDAR and the camera involves the calculation of a proper transformation matrix to align the coordinate systems of the two sensors. This process has been studied for many years in the fields of both robotics and computer vision. Extrinsic calibration methods can be classified into target-based methods and non-target methods. The focus of the target-based methods is to find corresponding features of the common target from multi-modal data. Non-target methods estimate the transformation matrix by maximizing the correlation of mutual information in multi-modal data, such as edges in images and discontinuity of the scanline in the point cloud~\cite{Levinson_2013, Taylor_2014}, as well as luminance of images and reflectance of the point cloud~\cite{Pandey_2014}. In this work, we~focus on the target-based extrinsic calibration.

For target-based calibration, the conventional method involves finding the vertices of a polygonal board, which can be a chessboard or a triangular board, both in the point cloud obtained by the LiDAR and the image captured by the camera either manually or automatically \cite{Mirzaei_2012,Park_2014,Moreno2014782}. The vertices are estimated by constructing the convex hull of the extracted board's point cloud. However, the vertices only include the geometric features of the target board's outer contour, which are formed by the discontinuity of the scanlines, while the information inside the counter is not used. To make full use of the information acquired by the LiDAR, the ``inner texture'' of the point cloud, i.e., the~reflectance intensity, which is also derived from the LiDAR sensor, is used. Unlike the approach in \cite{Pandey_2014}, we utilize the reflectance intensity to estimate the corners of the chessboard from the 3D point cloud. If the corners of the 3D point cloud are identified, the extrinsic calibration is converted to a 3D-2D matching~problem.

\begin{figure}[h!]
\centering
    \subcaptionbox{\label{intro_pcd_by_inte}}{\includegraphics[width=0.48\textwidth]{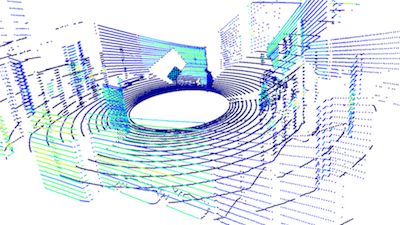}} 
    \subcaptionbox{\label{intro_zoomed_chessboard}}{\includegraphics[width=0.47\textwidth]{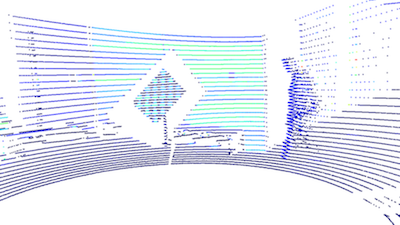}}\\
    \subcaptionbox{\label{intro_pano_img}}{\includegraphics[width=0.97\textwidth]{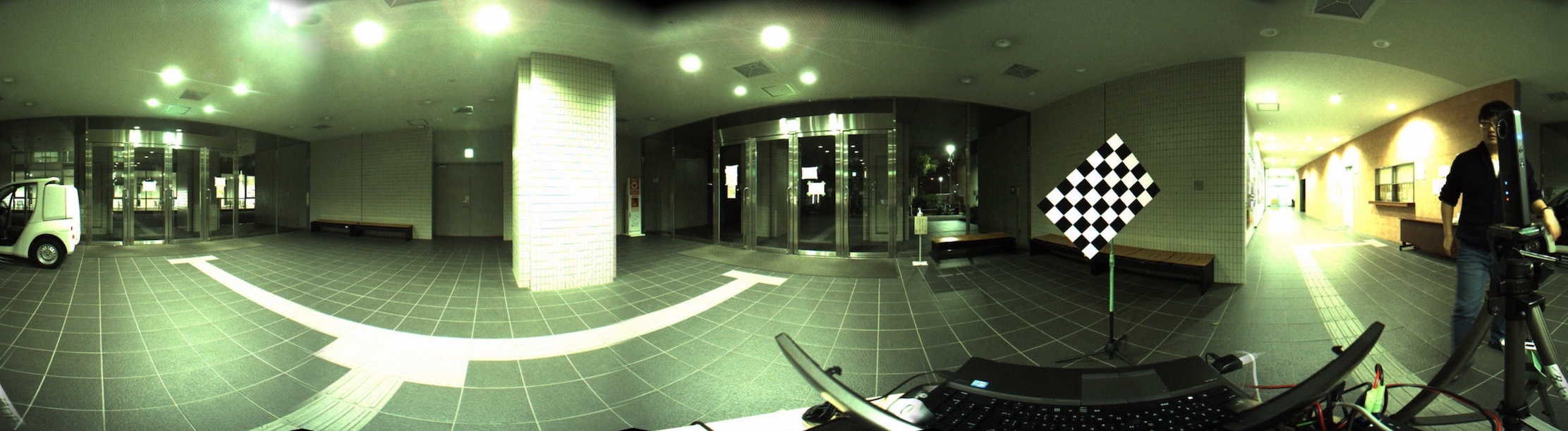}}
  \caption{Data from an identical scene captured by the LiDAR sensor and the panoramic camera. (\textbf{a})~The points are colored by the reflectance intensity (blue indicates low intensity, red indicates high intensity); (\textbf{b}) The zoomed chessboard. We can see the changes in reflection intensity of the point cloud between the white and black patterns; (\textbf{c}) The panoramic image of the same scene. }
  \label{intro_fig}
\end{figure}

The idea of the estimation of corners from the point cloud is based on the fact that the intensity of white patterns differs from that of black patterns, as seen in Figure \ref{intro_fig}. Figure \ref{intro_fig}a shows a point cloud; colors are representatives of the intensity. Figure \ref{intro_fig}b shows the zoomed-in part of the chessboard shown in Figure \ref{intro_fig}c, in which the change in reflectance intensity between the different patterns can be observed.  However, the noise and sparsity of the point cloud present a challenge. To address this challenge, we propose a novel method that fits a chessboard model to the chessboard's point cloud, and we also introduce Powell's method \cite{Powell_1964} for optimizing a defined cost function to obtain the fitting solution. After the corners of the chessboard in the 3D point cloud are estimated, the initial value of the transformation matrix is calculated using the Unified Perspective-n-Point (UPnP) method \cite{Kneip_2014}. The initial value is then applied to refine the result by using the nonlinear least squares optimization with the Levenberg-Marquardt (LM) method \cite{Levenberg_1944,marquardt1963algorithm}. We evaluate the proposed method by using data obtained from a {Velodyne HDL-32e LiDAR} sensor and a {FLIR Ladybug3 panoramic camera} under the proposed reflectance intensity-based re-projection error metrics.

Three main contributions of this work are as follows. First, we propose a method that utilizes the reflectance intensity of the point cloud to estimate the corners of the chessboard's point cloud. Further, we introduce Powell's method to optimize the corner detection problem. To the best of our knowledge, this is the first published work that utilizes the intensity information to estimate the corners from the point cloud. Second, the quantitative error evaluation of the proposed method is conducted using simulations. In addition, we analyze the relationship between the estimation error and other conditions such the measurement error and the distance of the chessboard. Finally, we provide the Python implementation of the proposed method, which can be downloaded from \href{https://github.com/mfxox/ILCC}{https://github.com/mfxox/ILCC}.


The rest of this paper is structured as follows. Related works are revisited in Section \ref{sec:related_works}. The~ overview of the proposed method and notations used in this work are described in Section~\ref{sec:overview}. In~ Section~\ref{sec:corner_est}, detailed procedures from automatic detection of the chessboard to the corner estimation and optimization are explained. The corner detection on the panoramic image and the correspondence grouping of the detected corners in modal data for the two devices is described in Section \ref{sec:ext_cal}.  In~addition, the process of the final extrinsic calibration after the correspondence grouping of the corners is described in this section. Simulation results for corners estimation in the point cloud and experimental results are described in Section \ref{sec:exp}. Discussions about this work and some tips for better extrinsic calibration with the proposed method are made in Section \ref{sec:dis}. Finally, we present our conclusions and scope for future work in Section \ref{sec:con}.


\section{Related Works}
\label{sec:related_works}
Extrinsic calibration of two sensors in the fields of computer vision as well as robotics is usually achieved by solving an optimization problem under certain constraints. These constraints can be geometric relationships, the correspondence of common features,  or the correlation of mutual information between two sensor modalities. Related works we revisited during the course of this study are classified on the basis of these types of constraints.

\subsection{Multiple Views on a Planar Checkerboard}
Zhang (2004) first published the work about the calibration for a system comprising a 2D LiDAR and a monocular camera \cite{Zhang_2004}. He extrinsically calibrated the two sensors under the constraint that in the camera coordinate system, the scalar projection of a vector between the origin and a point on the plane onto the normal of the plane equals the distance between the origin and the plane. A~system for 3D LiDAR sensors with constraints similar to those in Zhang's method was proposed and implemented as a {MATLAB toolkit} in \cite{unnikrishnan2005fast}. This system also made several assumptions, key among those being that the point cloud is dense enough, which is challenging when using single frame data obtained by mobile LiDAR sensors such as the Velodyne HDL-32e.  Pandey et al. extended the method to a system of {Velodyne HDL-64E LiDAR} and Ladybug3 panoramic camera, which is similar to the set we used for this work \cite{Pandey_2010}. This method was applied to the Ford Campus Dataset \cite{Pandey_2011}. Mirzaei et al. (2012) subsequently proposed a method for both intrinsic and extrinsic calibration in \cite{Mirzaei_2012}.

\subsection{Multiple Geometry Elements}
The extrinsic calibration can also be performed if common geometry features like points, line~segments, or planes are extracted from the data obtained by the two sensors and the correspondence of features from the two modalities are built. Scaramuzza et al. proposed a~method based on the point correspondences \cite{Scaramuzza_2007}. Once the point correspondences are known, the transformation matrix can be calculated using the methods for PnP (Perspective n Points) problems. However, it is difficult to  manually identify the corresponding 3D point features accurately in point cloud, especially for a sparse point cloud. To overcome this drawback, Moghadam et al. proposed a method based on automatically extracted 2D and 3D lines \cite{Moghadam_2013}. However, this approach requires the user to manually determine the correspondence of lines. Gong et al. proposed a method under the plane-to-plane constraints using a trihedral object, which also requires human intervention for plane selection \cite{s130201902}.
There are also some methods that automatically extract feature points, such as vertices of a polygonal planar checkerboard, from the LiDAR data. Nevertheless, these approaches require either manual operation for feature points selection from the image \cite{Park_2014} or customized checkerboard for feature points generation in the point cloud \cite{Moreno2014782}.
Geiger et al. proposed an automatic method for extrinsic calibration with one shot of multiple checkerboards in \cite{Geiger_2012}. This approach recovers the 3D structure from the detected corners in images. Then the calibration is performed under the constraints that planes of the chessboards recovered from the images should coincide with the planes detected from the LiDAR data. This method was applied for extrinsic calibration between the cameras and the LiDAR sensor in KITTI Dataset \cite{Geiger_2013}. However, to recover the 3D structure from corners of different chessboards, the cameras require stereo configuration for sufficient common field of view of the chessboards. This is challenging for panoramic camera like Ladybug3 we use in this work. Geiger's method for corner detection in images  showed robustness from the experimental results, and hence we also apply it for corners detection from the panoramic images in this work.

\subsection{Correlation of Mutual Information}
The abovementioned methods require either artificial observation objects (chessboards or triangular boards) or user intervention (features or correspondence selection).  This is inconvenient for applications that need frequent extrinsic calibration like autonomous driving cars, in which case the relative pose may drift due to the vibration. To solve this problem, several online calibration methods are proposed. These methods are generally based on the maximization of the mutual information (MI), such as edges in images and discontinuity of the scanline in the point cloud or luminance of images and reflectance intensity of the point cloud \cite{Levinson_2013,Pandey_2014, Taylor_2014}.
However, online calibration methods remain difficult to apply with poor initializations. This means that the pre-knowledge of a roughly accurate initial guess, which maybe estimate using the off-line calibration method or manual measurement, is~necessary.

\subsection{Our Approach}
In our proposed method for extrinsic calibration, we also applied constraints by obtaining corresponding feature points on a printed chessboard. This method belongs to the constraints on multiple geometry elements. However, unlike the approaches mentioned above, we use the corners instead of the vertices of the chessboard's point cloud. To estimate the corners of the point cloud, we formulate a cost function under the constraint that the 3D points with high reflectance intensity must lie on the white patterns of the chessboard and vice versa. Once the corners are detected in both multi-modal data, the correspondences are further made in a predefined order. Initial parameters can be obtained using the UPnP method \cite{Kneip_2014} on the corresponding corners and further refined using the Levenberg-Marquardt method \cite{Levenberg_1944,marquardt1963algorithm}.  
Our method is fully automatic and thus dose not require user intervention in the whole process from the detection of the chessboard, corner detection, the~correspondence of the corners, to the final optimization.

A combination of a Velodyne HDL-32e LiDAR sensor and a Ladybug3 panoramic camera is used for experiments in this work. There are several studies on the intrinsic calibration of the 3D LiDAR sensor  \cite{Atanacio_Jim_nez_2011,Mirzaei_2012} and the panoramic camera \cite{Zhang_2000,Scaramuzza_2006}. In the remainder of this paper, we assume that before the process of the extrinsic calibration, both the LiDAR and the panoramic cameras have been intrinsically calibrated from the factory setting.


\section{Overview and Notations}
\label{sec:overview}
\vspace{-6pt}

\subsection{Overview}
\begin{figure}[h!]
   \centering
   \includegraphics[width=0.9\textwidth]{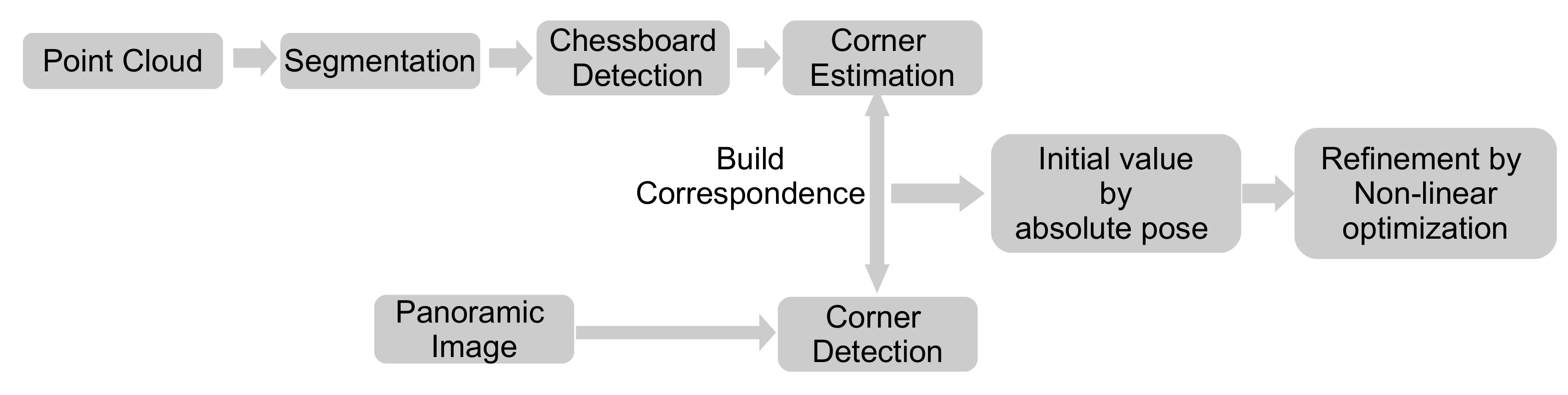}
   \caption{Overview of the proposed method.}
   \label{overview}
\end{figure}

The overview of the proposed method is illustrated in Figure \ref{overview}. First, the point cloud obtained from the LiDAR is segmented into multiple parts. The point cloud of the chessboard is identified from within the segments based on the characters of the segment. The corners of the chessboard in the point cloud are estimated by minimizing a defined cost function. On the other hand, corners of the chessboard in the image are detected using an existing method. Correspondence of the corners is built based on the predefined counting order. The corresponding pairs are then used to estimate  an initial value of the transformation matrix by solving an absolute pose problem. Finally, the value is refined by optimizing a proposed nonlinear cost function.
\subsection{Notations}
For the convenience of explanation, the following notations are used in this paper.
\begin{itemize}
\item $\bm{p}_i=(x_i,y_i,z_i)^T$: coordinates of a 3D point.
\item $\textit{P}=\{ \bm{p}_1, \bm{p}_2, \ldots,\bm{p}_n \}$: set of n 3D points.
\item $\bm{ \theta}=(\theta_x,\theta_y,\theta_z)^T$:  rotation angle vector whose element corresponds to the rotation angle along $x$-, $y$-, $z$-axis respectively.
\item $\bm{t}=(t_x,t_y,t_z)^T$: the translation vector. 
\item $\matr{R}(\bm{ \theta})=\matr{R}_z(\theta_z)\matr{R}_y(\theta_y)\matr{R}_x(\theta_x)$: rotation matrix. 
\item $T_r(\bm{ \theta},\bm{t},\bm{p_i})=\matr{R}(\bm{ \theta})\bm{p_i}+\bm{t}$: function that transforms the 3D point $\bm{p}_i$ with the angle vector $\bm{\theta}$ and translation vector $\bm{t}$.
\item $\bm{\hat{p}}_i=T_r(\bm{ \theta},\bm{t},\bm{p}_i)$: transformed point of $\bm{p}_i$.
\item $\textit{P}^c=\{ \bm{p}^c_1, \bm{p}^c_2, \ldots,\bm{p}^c_N \}$: set of estimated 3D corner points of the chessboard from the point cloud. $N$ is the number of the corners in the chessboard.
\item $\bm{\mathtt{x}}_i=(u_i,v_i)^T$: coordinates of 2D pixel.
\item $\mathtt{X}^c=\{ \bm{\mathtt{x}}^c_1, \bm{\mathtt{x}}^c_2, \ldots,\bm{\mathtt{x}}^c_N \}$: set of detected 2D corner pixels of the chessboard from the image.

\end{itemize}

\section{Corner Estimation from the Point Cloud}
This section explains the detailed process of corner estimation from one frame of the point cloud data obtained by the LiDAR sensor. All concerned coordinates are located in the LiDAR coordinate system in this section.
\label{sec:corner_est}
\subsection{Automatic Detection of the Chessboard}
This subsection describes the procedure of automatic extraction of the points that are reflected from the chessboard. The points discussed in this subsection refer to all points in one frame of the point cloud.
\subsubsection{Segmentation of the Point Cloud}
Region growing \cite{rabbani2006segmentation} is often used to segment the point cloud.  Region growing estimates the curvature value and the normal vector of each point based on the plane constructed by its kNN points. Then the method clusters the points according to the Euclidean distance and the angle of normal vectors of points. RANdom Sample Consensus (RANSAC) \cite{RANSAC} is also used for the shape extraction. For example, the RANSAC of the planar model is applied for the plane fitting and extraction from the point cloud.  However, both methods encounter challenges while processing the sparse and non-uniformly distributed point cloud, which may be generated by a single frame scanning of the mobile LiDAR sensor like Velodyne HDL-32e in this work. 

The scanline-based segmentation methods are suitable for processing this kind of point cloud, such as the method in \cite{Wang_2016}. This method first cluster a single frame of the point cloud into scanline segments according to change of distance and direction between successive points along the scan direction. Then, scanline segments are agglomerated into object segments based on their similarity. This method showed stable segmentation result from the experimental results and thus we apply it for segmentation in this work. 


\subsubsection{Finding the Chessboard from the Segments}
\label{subsec:find_chessboard}
After the segmentation of the point cloud, the segment of the chessboard needs to be correctly identified. We use characteristics such as the planarity, 
bounds, and points distribution of segments as the conditions for filtering the segments automatically.

\begin{figure}[h!]
        \centering
          \begin{minipage}{0.5\textwidth}
        \captionbox{Angular resolution of the used LiDAR in this work. The left figure is the top view and the right one is the side view of the LiDAR and the chessboard . \label{fig:theo_num}}{\includegraphics[width=0.95\textwidth]{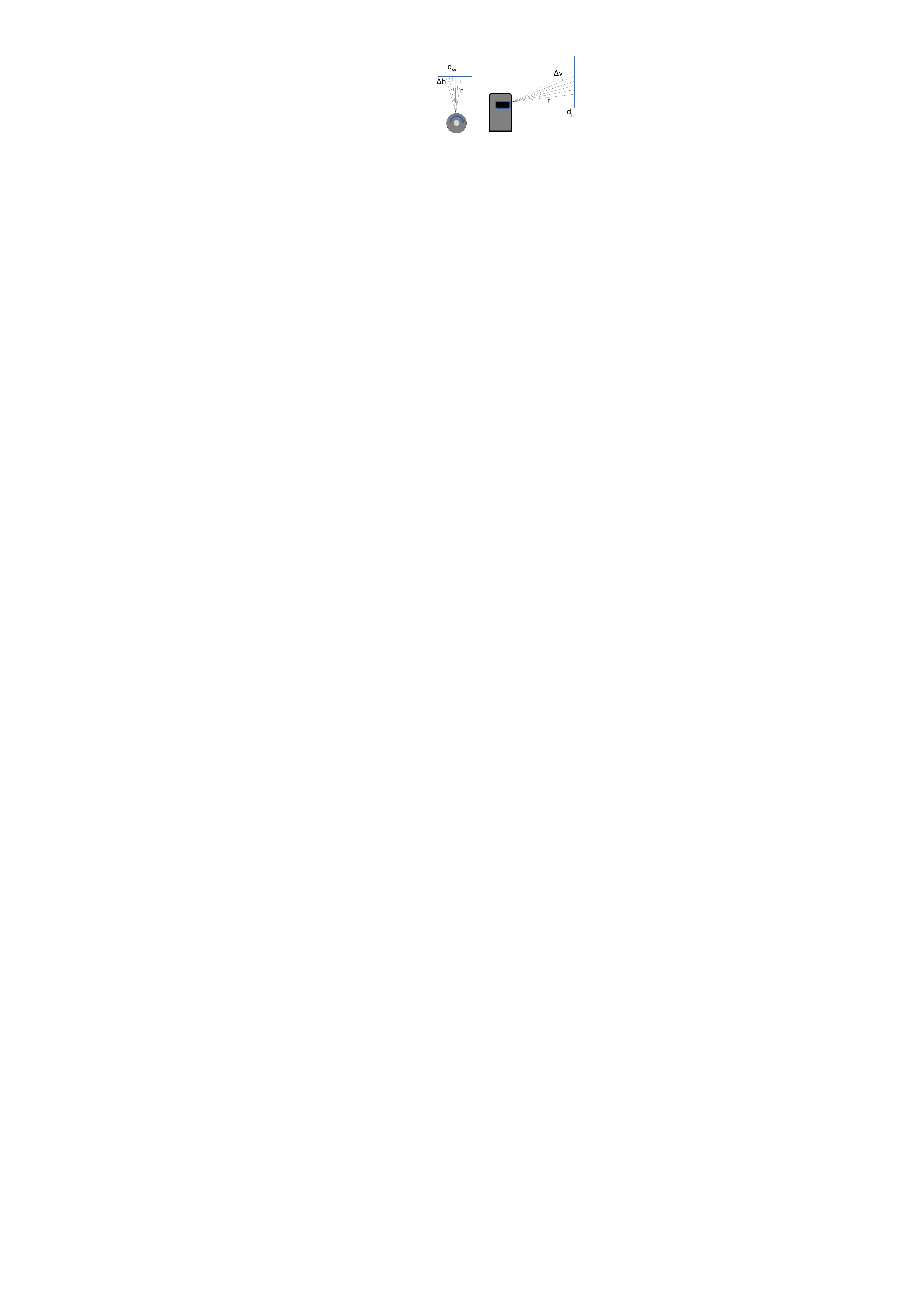}}
        \end{minipage}
        \hspace{-1.5cm}
        \begin{minipage}{0.58\textwidth}\centering
        \subcaptionbox{\label{fig:find_corner:distr1}}{\includegraphics[width=0.34\textwidth]{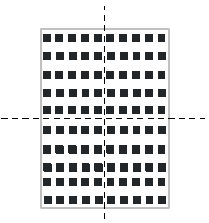}}
        \subcaptionbox{\label{fig:find_corner:distr2}}{\includegraphics[width=0.3\textwidth]{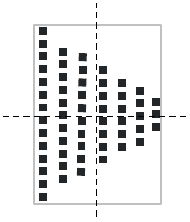}}
        \caption{Uniformity of the points distribution.}
        \label{fig:find_corner:distr}
\end{minipage}
\end{figure}

To reduce the computational cost, we first filter out some improbable segments based on the theoretical number $n_{theo}$ of points of a segment defined in Equation \eqref{eq:points_num}. $n_{theo}$ represents the theoretically maximum number of points and calculated from the vertical and horizontal angle of the LiDAR when the chessboard is parallel to the rotation axis of Velodyne, as shown in {Figure} \ref{fig:theo_num}. Segments in which the number of points fall within the interval $[\epsilon_{theo}n_{theo},n_{theo}]$ are further processed, where $\epsilon_{theo}$ is a~ coefficient and empirically set to 0.5 for this study. 
\begin{equation}
\label{eq:points_num}
             n_{theo}\approx\lfloor\frac{d_W}{2r\sin(\frac{\Delta h}{2})}\rfloor\lfloor\frac{d_H}{2r\sin(\frac{\Delta v}{2})}\rfloor \\
\end{equation}
where $d_W$ and $d_H$ are the width and height of the chessboard, $r$ is the Euclidean distance from the segment's centroid to the LiDAR sensor, $\Delta h$ and $\Delta v$ represent horizontal and vertical angular resolution, which are $0.16^\circ$ and $1.33^\circ$ for Velodyne HDL-32e \cite{velodyne2012hdl} respectively, $\lfloor \rfloor$ represents the ceiling truncation of a real number.

The planarity of the segment is verified using the Principle Components Analysis (PCA) \cite{jolliffe2002principal} method. The matrix $\textit{M}_{n\times 3}$ consisting of all points in the segment is decomposed  along three basis vectors $\textit{M}_b=({\bm{\mu}_1}, {\bm{\mu}_2}, {\bm{\mu}_3})^T$ with components ratios $\lambda_1$, $\lambda_2$, $\lambda_3$ on each basis vector. The segment whose least ratio $\lambda_3$ is less than 0.01 is considered as a planar object.

For a planar segment, all points in this segment are projected to the estimated plane formed by the RANSAC method \cite{RANSAC} for further verification. Fitted points are denoted as $\textit{M}_{n\times 3,f}$. As the final step, the range of the bounding box and the uniformity of the distribution are checked. For manipulation convenience, we rotate coordinates of all points to the chessboard plane and align $\lambda_1$, $\lambda_2$ to the $x$-axis and $y$-axis with the Equation \eqref{eq:rotate}. Then the centroid of the chessboard's points $\textit{M}_{n\times 3,fr}$ is translated to the origin using of Equation \eqref{eq:trans}. 
\begin{equation}
\label{eq:rotate} 
 \textit{M}_{n\times 3,fr}={(\textit{M}_b\cdot M_{n\times 3,f}^\intercal)}^\intercal
\end{equation}
\begin{equation}
\label{eq:trans} 
 \textit{M}_{n\times 3,frt}=\textit{M}_{n\times 3,fr}-mean(\textit{M}_{n\times 3,fr})
\end{equation}

After the transformation by Equations \eqref{eq:rotate} and \eqref{eq:trans}, the planar segment is transformed to the XOY plane and the centroid of the segment coincides with the origin point. 
The segment with the bounding box within [0.8$d_W$, 1.6$d_W$] and [0.8$d_H$, 1.6$d_H$] 
is considered as a potential chessboard. Uniformity of the point distribution is determined by the difference between the points distribution in four equally divided regions, which is illustrated in Figure \ref{fig:find_corner:distr}. Two sample segments that fall in the range of the threshold bounding box
are shown in Figure \ref{fig:find_corner:distr}. The dashed lines divide the theoretical region of the chessboard into four equal parts. The difference of the points' number in each part is used for uniformity check. For example, as the difference of points' number in each part in Figure \ref{fig:find_corner:distr}a is smaller then that in Figure \ref{fig:find_corner:distr}b, Figure \ref{fig:find_corner:distr}a has better uniformity than Figure \ref{fig:find_corner:distr}b.
Let us assume that the maximum number of points in the regions is $n_{max}$, and minimum is $n_{min}$. The uniformity of the distribution is calculated as $\epsilon_{norm}=1-\frac{n_{max}-n_{min}}{n_{all}} \in [0,1)$, where $n_{all}$ is the total number of points in the segment. A~large $\epsilon_{norm}$ value indicates that the points are distributed normally. The threshold for uniformity $\epsilon_{norm}$ is set to 0.85.

If more than one segment satisfies the above conditions, the segment with greater uniformity is selected. The set of points in the detected chessboard's segment is denoted as $\textit{P}^M = \{ \bm{p}^M_1, \bm{p}^M_2, \ldots,\bm{p}^M_n \}$ (The superscript ``$M$'' refers to the chessboard Marker).  



\subsection{Corner Estimation}
This subsection explains the principle and detailed process for estimation of the corners from the point cloud of the chessboard. The points discussed in this subsection refers to points of the chessboard~only.
\subsubsection{Model Formulation}
\label{subsubsec:transform_xoy}
 
\begin{figure}[h!]
\centering
    \subcaptionbox{\label{model_grid}}{\includegraphics[width=0.16\textwidth]{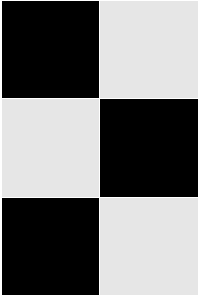}}\hspace{0.05\textwidth}
    \subcaptionbox{\label{model_pcd}}{\includegraphics[width=0.16\textwidth]{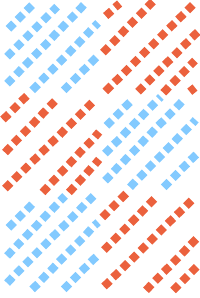}}  \hspace{0.05\textwidth}
    \subcaptionbox{\label{model_match}}{\includegraphics[width=0.16\textwidth]{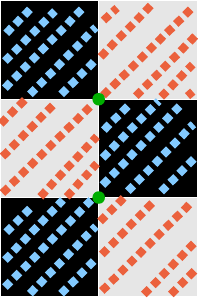}}\hspace{0.05\textwidth}
        \subcaptionbox{\label{model_corners}}{\includegraphics[width=0.16\textwidth]{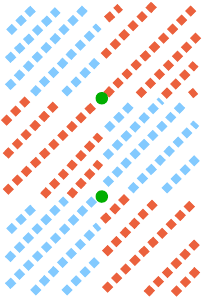}}
  \caption{The principle used to estimate corners in the points. {(\textbf{a})} The chessboard model; {(\textbf{b})} The scanned point cloud of the chessboard.  Colors indicate the intensity (blue for low and red for high reflectance intensity); {(\textbf{c})} Find a matrix that translates the most 3D points on the corresponding patterns. Green points are estimated corners; {(\textbf{d})} Consider the corners of the chessboard model as the corners of the point cloud.}
  \label{marker_model_fig}
\end{figure}

After the chessboard points $\textit{P}^M$ are automatically extracted, we utilize the intensity of points to estimate the corners. Without loss of generality, we use a model chessboard with a 2 $\times$ 3 pattern as shown in Figure \ref{marker_model_fig}a. Figure \ref{marker_model_fig}b illustrates theoretical laser-reflected points from the chessboard by lasers. The blue points indicate low intensity reflected from black patterns while red points indicate high intensity reflected from white patterns. If we find a pose of the model that make the reflected points best fit the model (Figure \ref{marker_model_fig}c), we can use the corners (green points in Figure \ref{marker_model_fig}d) of the model to represent the corners of the points. 

\begin{figure}[h!]
        \centering
        \captionbox{Directions of the basis vectors relative to the LiDAR coordinate system. Blue arrow lines in the left figure represent the basis vectors decomposed by PCA. After transformation with the basis vectors, chessboard's points are mapped to the $X^POY^P$ (chessboard plane). Then we can apply the model described in Figure \ref{marker_model_fig}.  \label{fig:corner_est:dir}}{\includegraphics[width=1\textwidth]{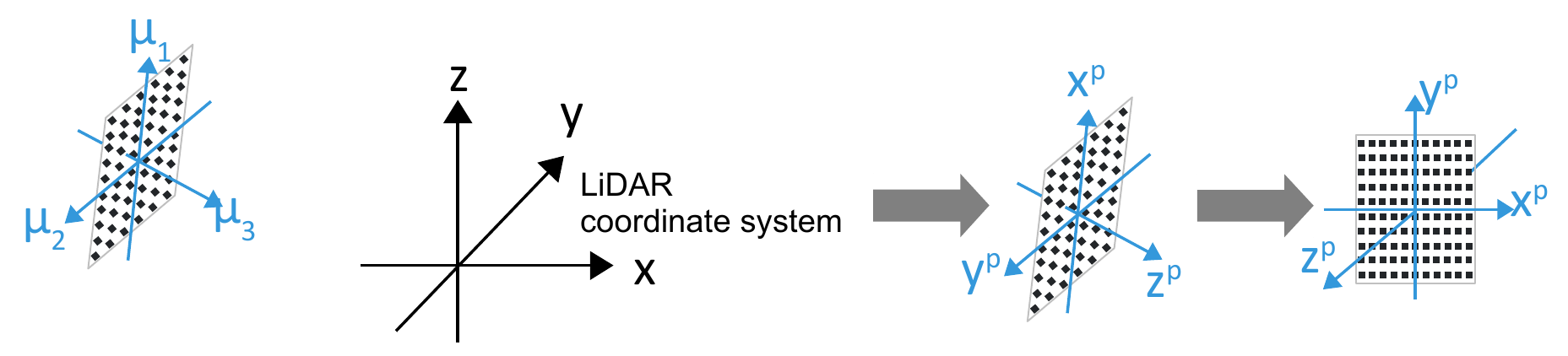}}
\end{figure}

In addition, similar to the process in Section \ref{subsec:find_chessboard}, the coordinate system of the chessboard is transformed to the chessboard plane by rotating with the matrix consisting of three PCA vectors and subtracting the mean of the rotated points, as shown in Figure \ref{fig:corner_est:dir}. Then, we can treat the points as illustrated in Figure \ref{marker_model_fig}. However, for the consistency of further correspondence with corners from images, the directions of principle vectors ${\bm{\mu}_1},{\bm{\mu}_2},{\bm{\mu}_3}$ are deliberately determined. The conditions for the basis vectors are illustrated in Figure \ref{fig:corner_est:dir}:
\begin{itemize}
    \item directions of ${\bm{\mu}_1},{\bm{\mu}_2},{\bm{\mu}_3}$ are defined to obey to the right hand rule.
    \item direction of ${\bm{\mu}_3}$ (the normal of the chessboard) is defined to point to the side of origin of the LiDAR coordinate system.
    \item angle between ${\bm{\mu}_1}$ and $x$ axis of the LiDAR coordinate system is not more than $90^\circ$
\end{itemize}

The rotation matrix $\textit{M}_{XOY^P}^M$ is defined as $({\bm{\mu}_1},{\bm{\mu}_2},{\bm{\mu}_3})^T$, where directions of ${\bm{\mu}_1},{\bm{\mu}_2},{\bm{\mu}_3}$ satisfy the aforementioned conditions. After the translation ${t}_{XOY^P}^M$ by subtracting the mean of the rotated points, ${\bm{\mu}_1}$, ${\bm{\mu}_2}$, ${\bm{\mu}_3}$ and the center of the original point cloud are transformed to $x^P$-, $y^P$-, $z^P$-axis and the origin plane coordinates respectively.

\subsubsection{Correspondence of Intensity and Color}
\begin{figure}[h!]\centering
\label{fig:intens}
    \subcaptionbox{\label{fig:intens:scatter}}{\includegraphics[width=0.48\textwidth]{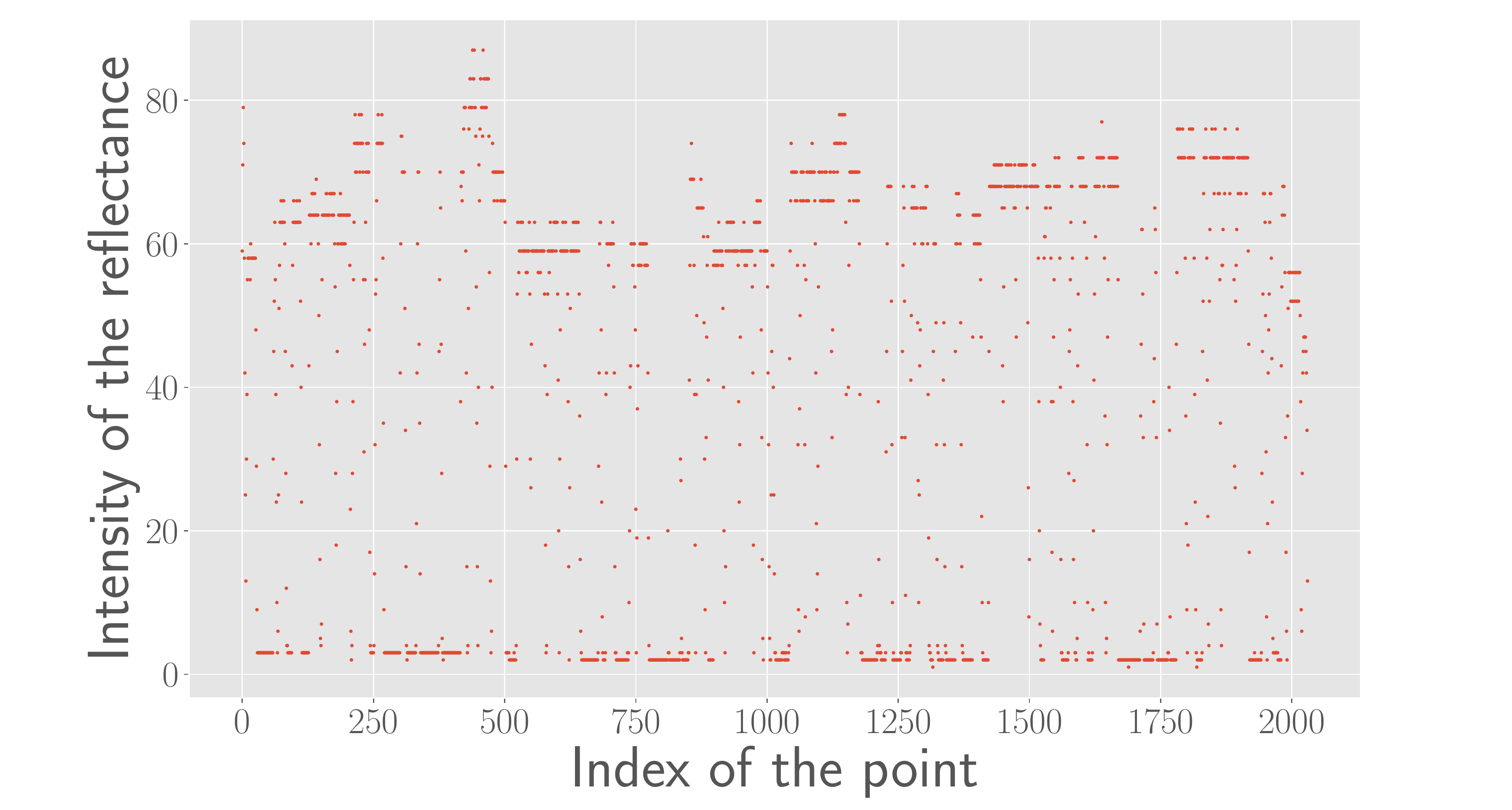}}
    \subcaptionbox{\label{fig:intens:hist}}{\includegraphics[width=0.48\textwidth]{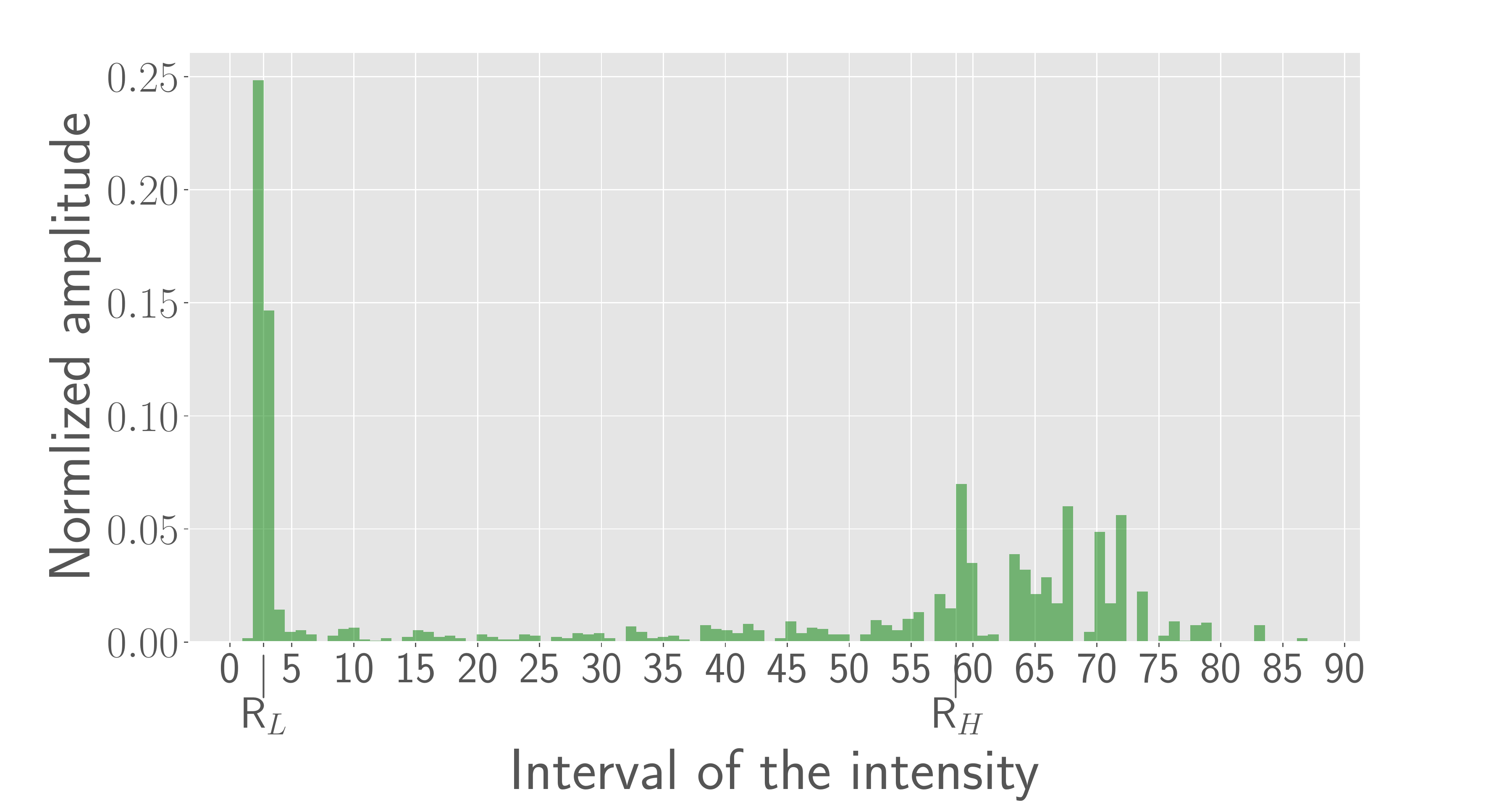}} 

    \caption{Estimated parameters for each frame. (\textbf{a}) Scatter diagram of the intensity for all points in the chessboard; (\textbf{b}) The histogram of the intensity. $R_L,R_H$ can be found at the peaks of the two sides.}
    \label{fig7}
\end{figure}

Another problem to solve before the formulation of the cost function is the correspondence between the reflectance intensity and the color of the pattern. There are only two colors in the pattern, black and white. However, the reflectance intensity values distribute discretely, mainly due to the divergence of the laser beam. Figure \ref{fig7}a shows the scatter plot of the intensity values of a real chessboard's point cloud. Further, the absolute value of the reflectance intensity is also affected by the~distance.

To process the intensity data adaptively, we define a range called the gray zone denoted as $[\tau_l,\tau_h]$. The points with the intensity less than $\tau_l$ are considered as reflected from black patterns and those greater than $\tau_h$ are considered as reflected from white patterns. To evaluate the values of $\tau_l$ and $\tau_h$, the~histogram is created and the bins ($R_L,R_H$) of peaks in both sides of the mean intensity are detected automatically, as shown in Figure \ref{fig7}b for illustration. The gray zone $[\tau_l,\tau_h]$ is then defined as follows,
\begin{equation}
\label{eq:grayzone}
\left\{  
             \begin{array} {ll} 
             \tau_l=& ((\epsilon_g - 1) R_L + R_H)/\epsilon_g \\  
             \tau_h=&(R_L + (\epsilon_g - 1)R_H)/\epsilon_g

             \end{array}  
\right.
\end{equation}
where $\epsilon_g\geq2$ is a constant.  $\epsilon_g$ is set to $2$ for corner estimation with enough points (gray zone will be zero and all points are used) and $4$ for error evaluation with enough confidence of the pattern color from the reflectance intensity in this~study.

\subsubsection{Cost Function and Optimization}
\label{subsubsec:cost_fun_corner}
After completing the processes explained above, we formulate the cost function for corner estimation. The cost function is formulated based on the constraints of the correspondence between the intensity and color, and defined in Equation \eqref{marker_cost_func}.
\begin{multline}\label{marker_cost_func}
    C_m=\sum_{i}^{n} f_g(r_i)\{ f_\text{in}(\bm{\hat{p}}_i,\bm{G})\left |c_{i}-\hat{c}_{i} \right | f_d(\bm{\hat{p}}_i,\bm{V_i}) 
    + [1-f_\text{in}(\bm{\hat{p}}_i,\bm{G})]f_d(\bm{\hat{p}}_i,\bm{G})  \}, \  \bm{\hat{p}}_i\in{\hat{\textit{P}}}^M
\end{multline}
where ${\hat{\textit{P}}}^M=\{T_r({\bm{\theta}}^M,\bm{t}^M,{\bm{p}_i}^M), \, {\bm{p}_i}^M\in\textit{P}^M\}$ is the set of points that are transformed to the XOY plane using the aforementioned processes, and~{$z$}~coordinates of all points in $\textit{P}^M$ is 0. Namely 3D points are degenerated to 2D after rotation by the matrix of three PCA vectors. Thus, the transformation parameters along the plane are ${\bm{\theta}}^M=[0,0,{\theta}_z^M]^T$ and $\bm{t}^M=[t_x^M,t_y^M,0]^T$. $r_i$ is the reflectance intensity of $\textit{i}$-th point. $f_{g}(r_i)$ is used to determine whether a point falls into the gray zone and is defined in Equation \eqref{gray_zone_func}. $\bm{G}$ represents the four corners of the chessboard and $\bm{V}_i$ represents the four vertices of the grid corresponding to the $\textit{i}$-th point.
\begin{equation}
\label{gray_zone_func} 
f_{g}(r_i)=
\left\{  
             \begin{array} {ll} 
             1 : & r_i \notin [\tau_l,\tau_h]\\  
             0 : & \text{else}

             \end{array}  
\right.
\end{equation}
\begin{equation}
f_{in}(\bm{\hat{p}}_i,\bm{V}_i)=
\left\{  
             \begin{array} {ll} 
             1 : & \text{if } \bm{\hat{p}}_i \text{ in the  polygon with vertices }  \bm{G} \\  
             0 : & \text{else}

             \end{array}  
\right.
\end{equation}

\begin{equation}\label{fd1}
f_d=\min(\Delta  {x}_1,\Delta  {x}_2)+\min(\Delta  {y}_1,\Delta  {y}_2)
\end{equation}

\begin{figure}[h!]\centering
    \subcaptionbox{\label{marker_cost_1}}{\includegraphics[width=0.3\textwidth]{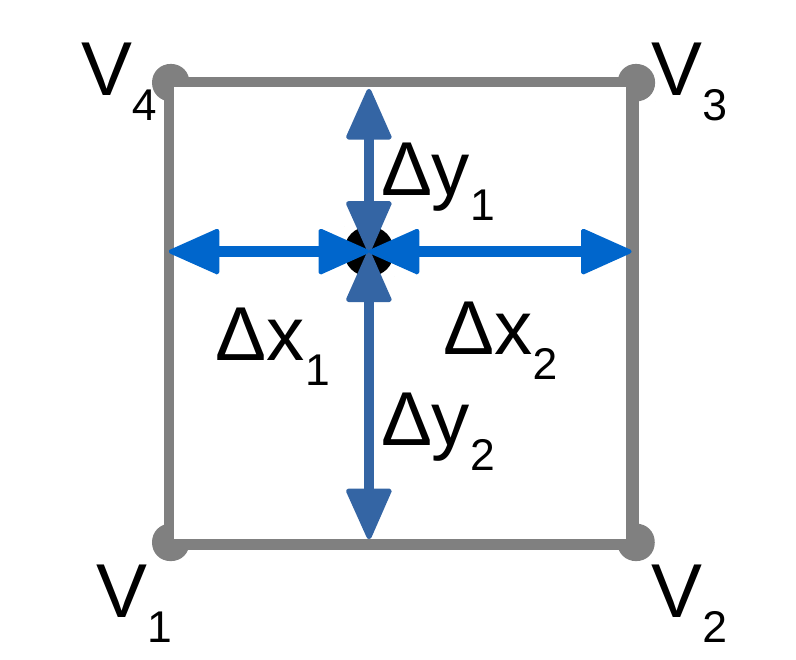}} 
    \subcaptionbox{\label{marker_cost_2}}{\includegraphics[width=0.27\textwidth]{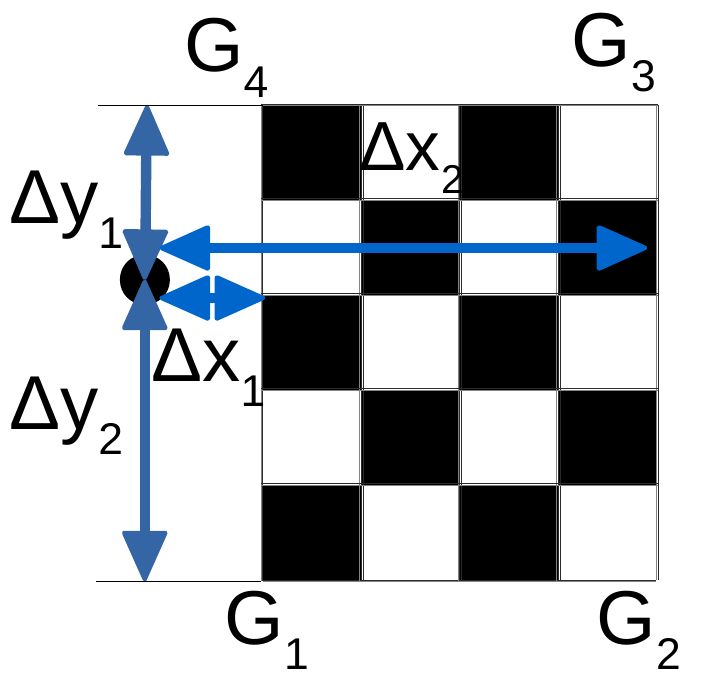}}
  \caption{Cost definition for corner detection in the point cloud. (\textbf{a}) An example of the point falling into the wrong pattern. The square represents a white pattern of the chessboard; (\textbf{b}) An example of the point falling out of the chessboard model; (\textbf{a}) describes the first term and (\textbf{b}) describes the second term of the cost function (Equation \eqref{marker_cost_func}).}
  \label{marker_cost_fig}
\end{figure}

$f_\text{in}(\bm{\hat{p}}_i,\bm{V_i})$ indicates whether the polygon with the vertices $\bm{V}_i$ contains the point $\bm{\hat{p}}_i$. $c_i$ is the estimated color from reflectance intensity $r_i$. We define $c_i=0$ if $r_i<\tau_l$ and $c_i=1$ if $r_i>\tau_h$. $\hat{c}_i$~represents the color of the pattern the $\hat{p}_i$ falls in. It is defined as 0 for black and 1 for white. 
We~experimentally use $L1$ distance to calculate the cost for  points that fall in the chessboard or out the chessboard as shown in Figure \ref{marker_cost_fig}, and the function $f_{d}$ is defined in Equation \eqref{fd1}. Figure \ref{marker_cost_fig}a shows the situation that a point $\hat{p}_i$ falls in the wrong pattern constructed by $\bm{V}_i$. $\Delta  {x}_1,\Delta  {x}_2$ represent the distances from the point $\hat{p}_i$ to the two sides of the pattern and $\Delta  {y}_1,\Delta  {y}_2$  represent the distances to the other two sides.  Figure \ref{marker_cost_fig}b shows the situation that a point falls out of the chessboard's region. Similarly, $\Delta  {x}_1,\Delta  {x}_2$ represent the distances from the point $\hat{p}_i$ to the two sides of the chessboard and $\Delta  {y}_1,\Delta  {y}_2$ 
represent the distances to the other two sides of the chessboard in Figure \ref{marker_cost_fig}b.

Since the cost function $C_m$ is discontinuous, it is impractical to derive for optimization. Thus, we~utilize the Powell optimization method \cite{Powell_1964} that needs no derivative. We used the implementation by SciPy \cite{scipy} in this work. As mentioned before, the set of chessboard's points are almost fitted to the chessboard model after the transformation. Therefore, all three parameters ${\theta}_z^M$, $t_x^M$, $t_y^M$ are initialized with zero when the Powell's method is applied.



\section{Extrinsic Calibration Estimation}
\label{sec:ext_cal}
   \vspace{-6pt}
\subsection{Corner Estimation from the Image}

There exist many methods for corner  detection in perspective images or even in distorted or blurred images \cite{Rufli_2008,Geiger_2012}. In this work, we utilize the method in \cite{Geiger_2012} to detect the corners from the panoramic images directly. 


\subsection{Correspondence of the 3D-2D Corners}

After the corners on the images are detected, the detected corners both on the image and in the point cloud are corresponded by defining the common counting order that starts from left-down side of the chessboard.

\subsection{Initial Value by PnP}
With these corresponding 3D-2D pair points, an initial value for nonlinear optimization can be obtained by estimating the central absolute pose with the UPnP (Unified Perspective-n-Point) method~\cite{Kneip_2014}. We used the implementation by OpenGV \cite{6906582}.
\subsection{Refinement with Nonlinear Optimization}
We use the difference of inclination angle and azimuth angle in the spherical coordinate system as the error metric for optimization to be independent of the panoramic image projection models. For the i-th 3D-2D pair, the residual is calculated as $E_{err}=\sqrt{f_{p2a}^2(\hat{\bm{p}}_i)-f_{\mathtt{x}2a}^2(\bm{\mathtt{x}}_i)}$, where $f_{p2a}$ and $f_{\mathtt{x}2a}$ convert a 3D point and a pixel to the inclination angle and azimuth angle, respectively, in the spherical coordinate system.

\section{Experimental Results and Error Evaluation}
\label{sec:exp}
The setup of the sensors and the measurement environment in this work are introduced in Section~\ref{sec:sub:setup}. To evaluate the accuracy of 3D corner detection with the proposed method, simulation results under different conditions are presented in Section \ref{sec:sub:sim_res}. Then the method is applied to the real data for 3D corner detection of the chessboard's point cloud. Some results of detected 3D corners as well as 2D corners of the panoramic images are shown in Section \ref{sec:sub:corners}. Section \ref{sec:sub:extrinsic_paras} presents the extrinsic parameters estimated with the 3D-2D corner correspondences. Finally, quantitative and qualitative evaluations are performed in Sections \ref{sec:sub:reproj_err} and \ref{sec:sub:reproj_res} respectively.

\subsection{Setup}
\label{sec:sub:setup}

\begin{figure}[h!]
   \centering
   \subcaptionbox{\label{fig:setup:1}}{\includegraphics[width=0.22\textwidth]{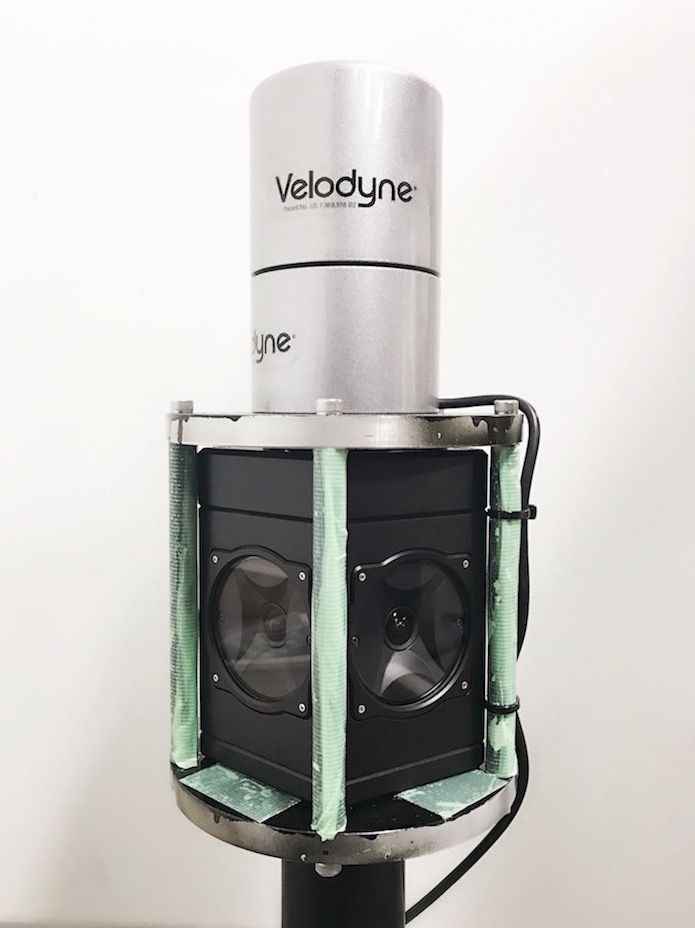}}
   \hspace{0.3cm}
   \subcaptionbox{\label{fig:setup:2}}{\includegraphics[width=0.36\textwidth]{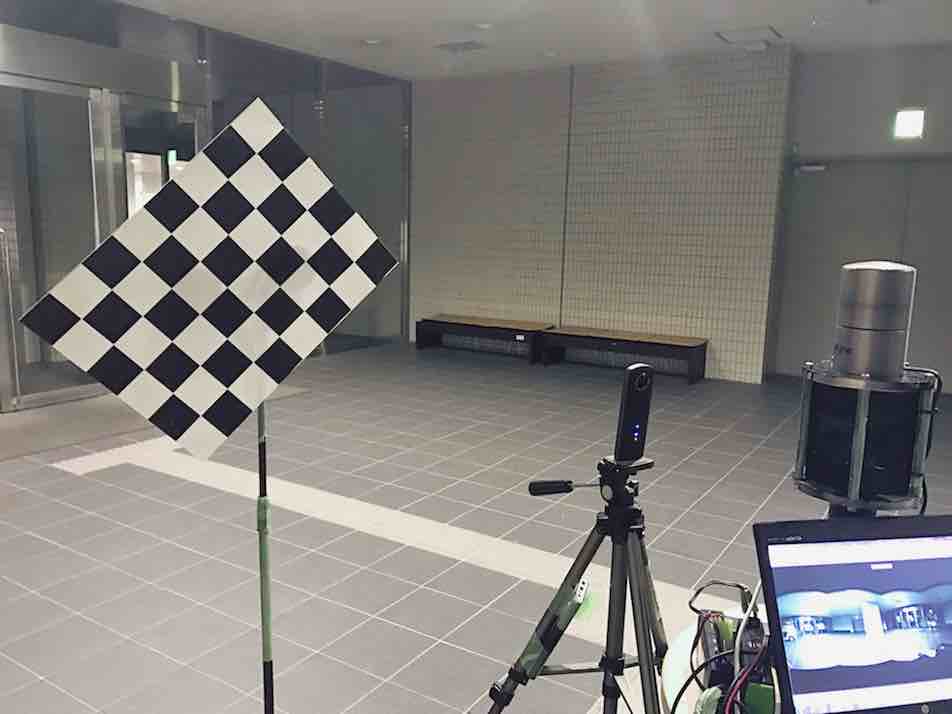}}

   \caption{Setup. (\textbf{a}) the image of the setup for two sensors; (\textbf{b}) a scene for the data acquisition. }
   \label{setup}
\end{figure}

We setup the system with a Velodyne HDL-32e LiDAR sensor mounted atop a Ladybug3 camera as shown in Figure \ref{setup}a. The chessboard used in this work is constructed by $6\times8$ grids with a side length of 7.5 cm as shown in  Figure \ref{setup}b. Velodyne rotates at 600 rpm and the resolution of the panoramic image output by Ladybug3 is set to 8000 $\times$ 6000.

\begin{figure}[h!]\centering
    \subcaptionbox{\label{fig:dataset:top}}{\includegraphics[height=4.5cm,keepaspectratio]{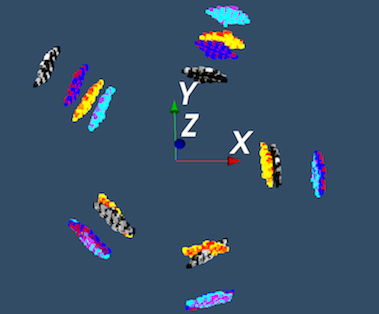}}
    \hspace{1cm}
    \subcaptionbox{\label{fig:dataset:side}}{\includegraphics[height=4.5cm]{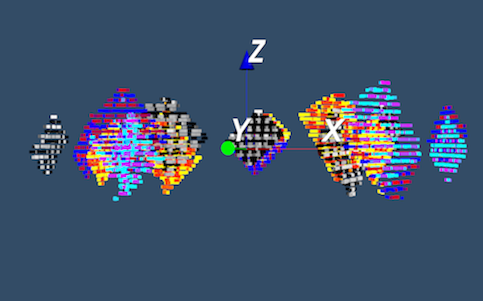}} 
  \caption{Distribution of the 20 chessboard positions. The chessboard is captured by the LiDAR from different heights and angles. The length of the coordinate axis is 1 m. Four groups of colors represent four positions of the chessboard for each horizontal camera. (\textbf{a}) Top view of the point clouds of the chessboard; (\textbf{b}) Side view of the point clouds of the chessboard.}
  \label{fig:dataset}
\end{figure}

In total, we captured 20 frames of the chessboard. For each horizontal camera of the Ladybug3, the chessboard is placed at 4 different places. The top and side views of the 20 frames are shown in~Figure~\ref{fig:dataset}. 

\subsection{Simulation for Corner Detection Error in the Point Cloud}
\label{sec:sub:sim_res}
To evaluate the error of estimated corners from the point cloud,  the ground truth is indispensable. However, it is difficult to obtain ground truth from the real data. Thus, we simulate the scanned point cloud of a generated chessboard model and record the corners' coordinates of the chessboard as the ground truth for evaluation. Subsequently, the error is evaluated by comparing the estimated results with the ground truth. 

\subsubsection{Simulation of the Point Cloud}
\begin{figure}[h!]\centering
    \subcaptionbox{\label{fig:dis_vs_ang:vertical_angs}}{\includegraphics[width=0.47\textwidth]{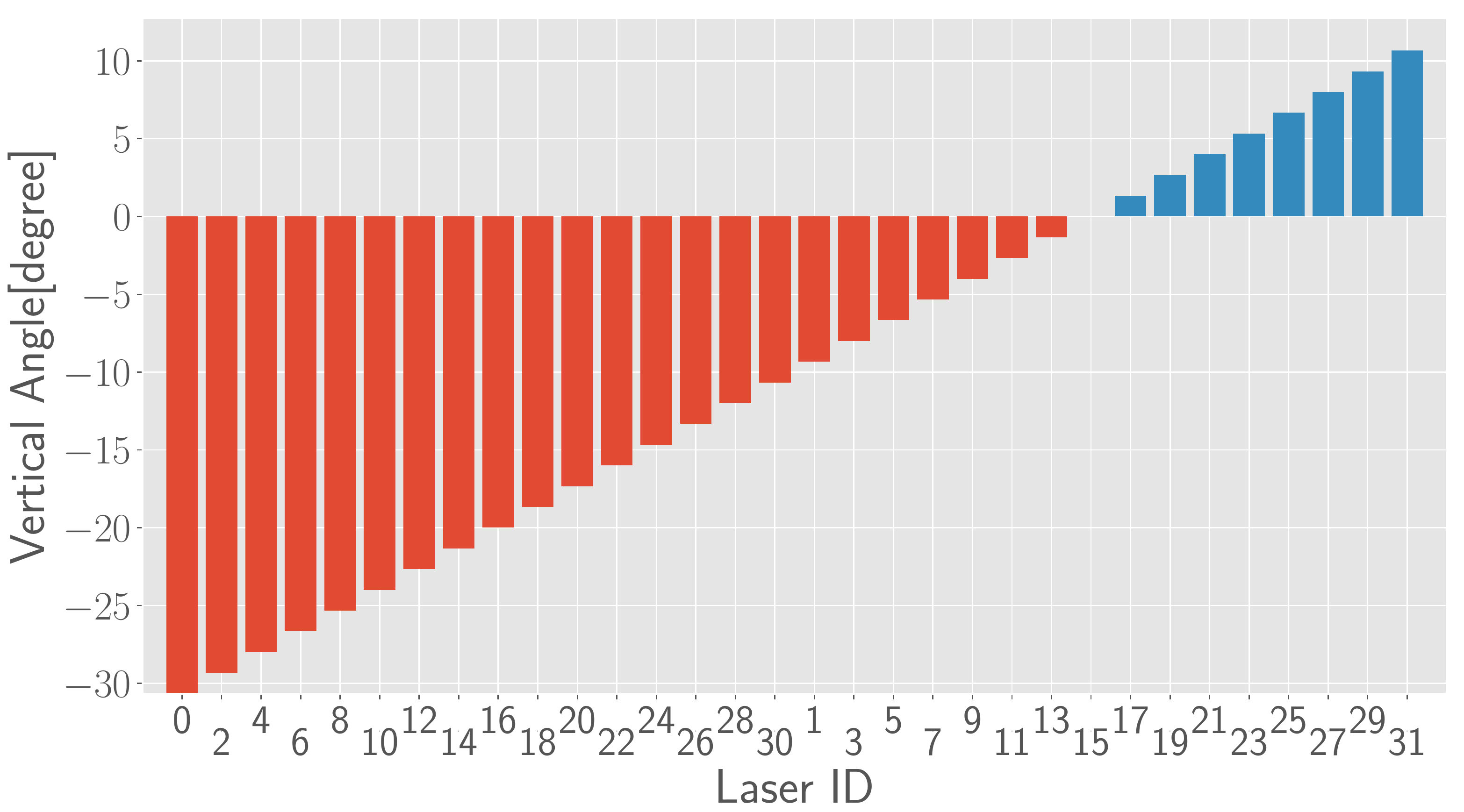}}
    \subcaptionbox{\label{fig:dis_vs_ang:lidar_model}}{\includegraphics[width=0.47\textwidth]{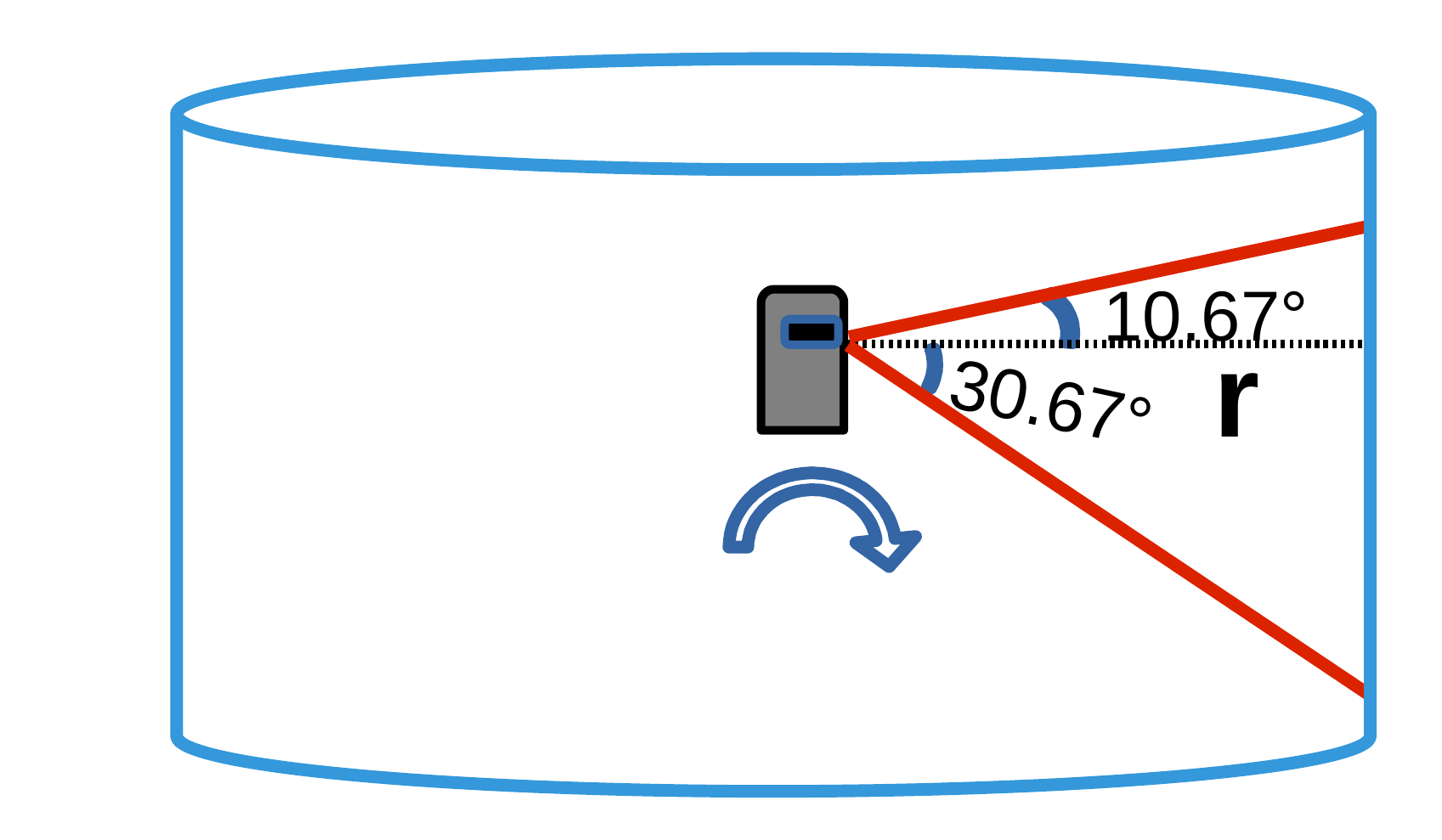}} \\ 
    \subcaptionbox{\label{fig:interval:h}}{\includegraphics[width=0.47\textwidth]{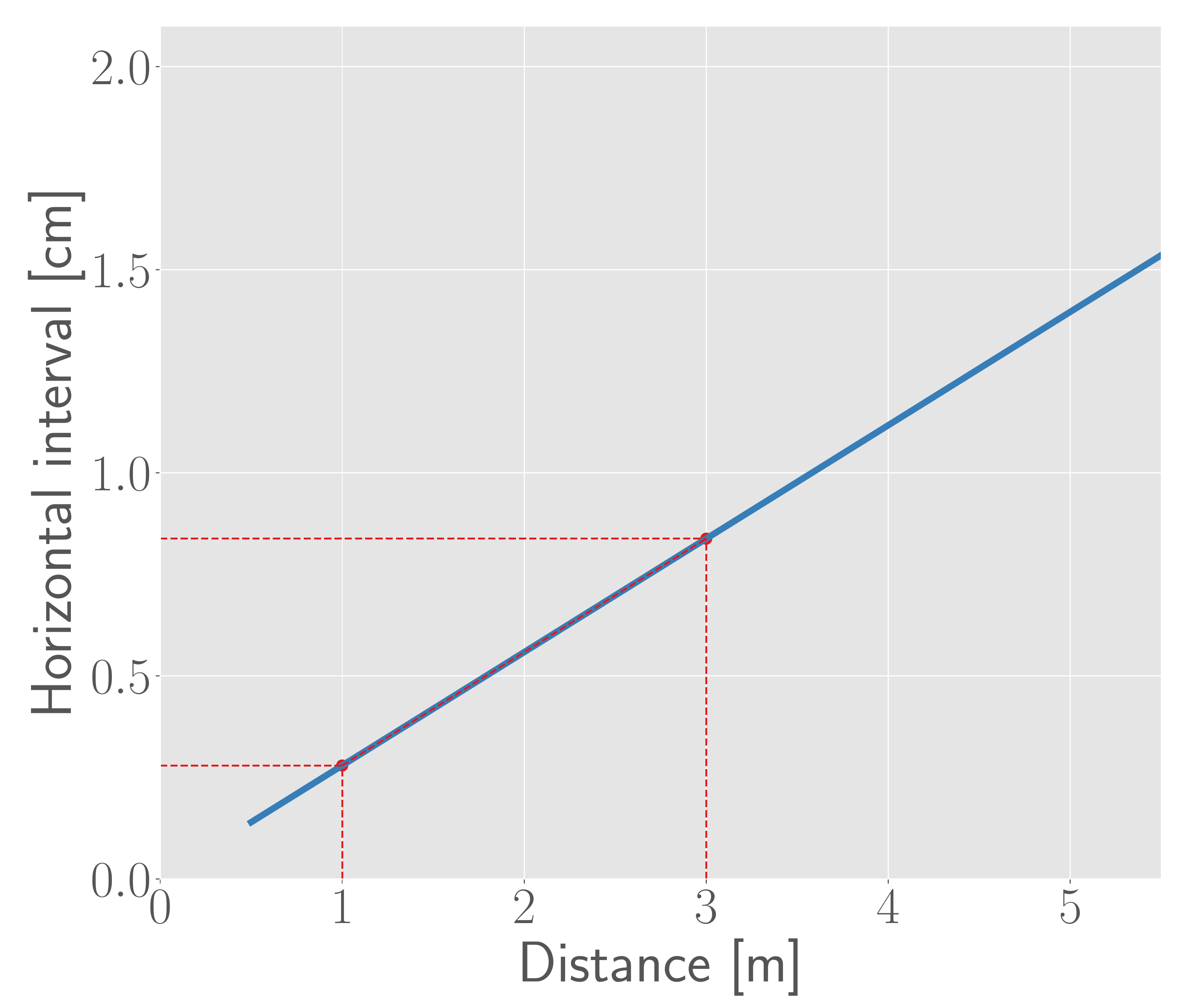}}
    \subcaptionbox{\label{fig:interval:v}}{\includegraphics[width=0.48\textwidth]{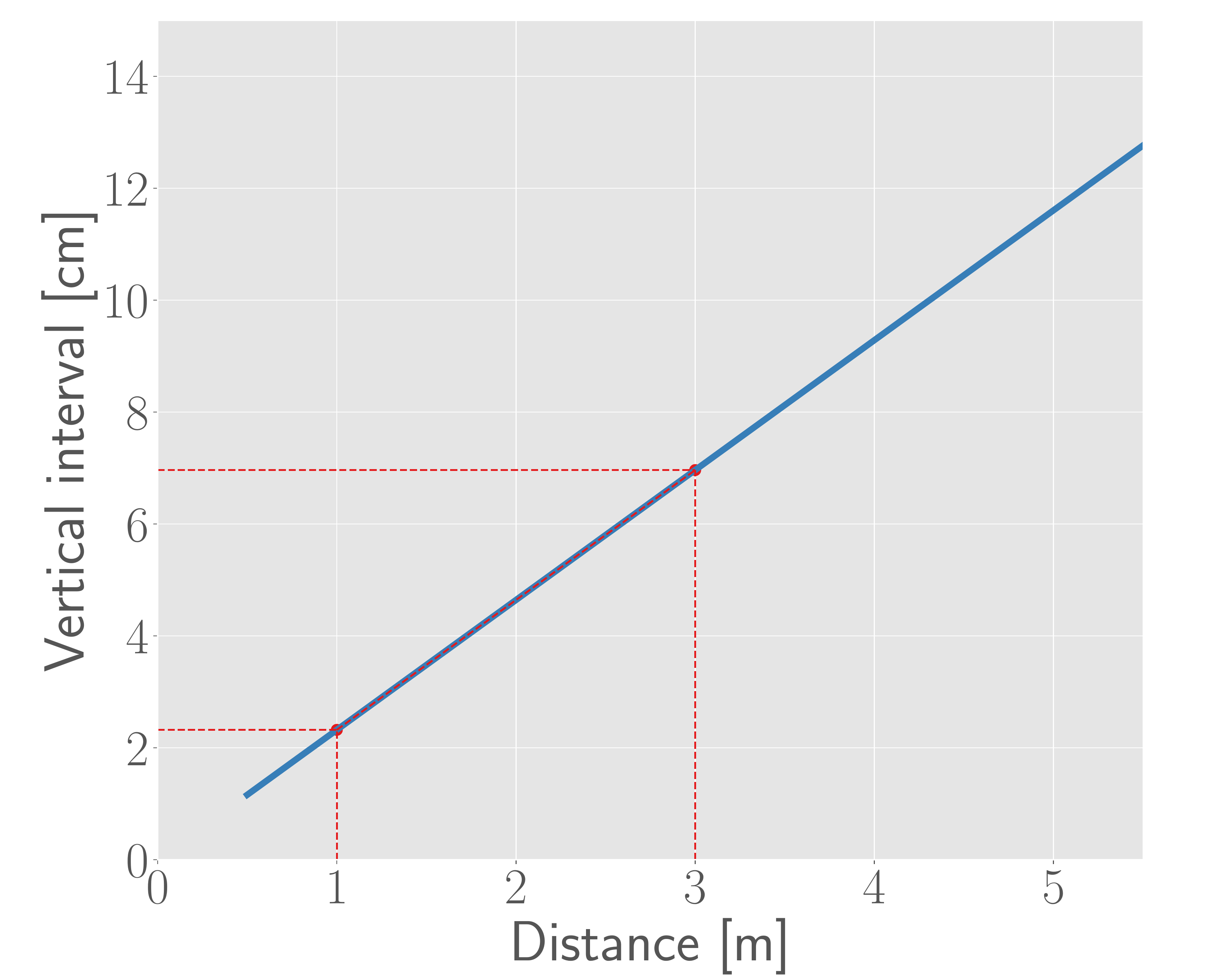}}   \vspace{-12pt}
  \caption{Vertical field view of Velodyne HDL-32e and the relationship between interval and distance. (\textbf{a}) Vertical angles of Velodyne HDL-32e; (\textbf{b}) Vertical field of view; (\textbf{c}) Relationship between the~ horizontal interval of two adjacent lasers and noise of the point cloud; (\textbf{d}) Relationship between the interval of two successive points of the scanline and the distance of the chessboard. Red lines in (\textbf{c},\textbf{d}) show the range of chessboard's distance we place in this work.}
  \label{fig:dis_vs_ang}
\end{figure}

There are many aspects that affect the measured distance and intensity  of the points reflected from the chessboard, such as distance, noise, divergence of the laser beam and the pose of the chessboard. The theoretical interval between two laser beams and two successive points along the scanning are calculated based on the distance and the LiDAR laser distribution, as shown in Figure \ref{fig:dis_vs_ang}. The noise is added to the points along $x$-, $y$-, $z$-axis independently. The probability model of the noise is considered as a Gaussian distribution. The base line of the deviation $\sigma$ for noise is empirically set to 0.0016, 0.0016 and 0.01 m with the mean $\mu=0$ for each axis. Simulated points are transformed with a rotation matrix and translation vector.

\begin{figure}[h!]\centering
    \
    \mbox{
        \begin{subfigure}[h]{.4\textwidth}
        \centering
            \includegraphics[height=3.5cm]{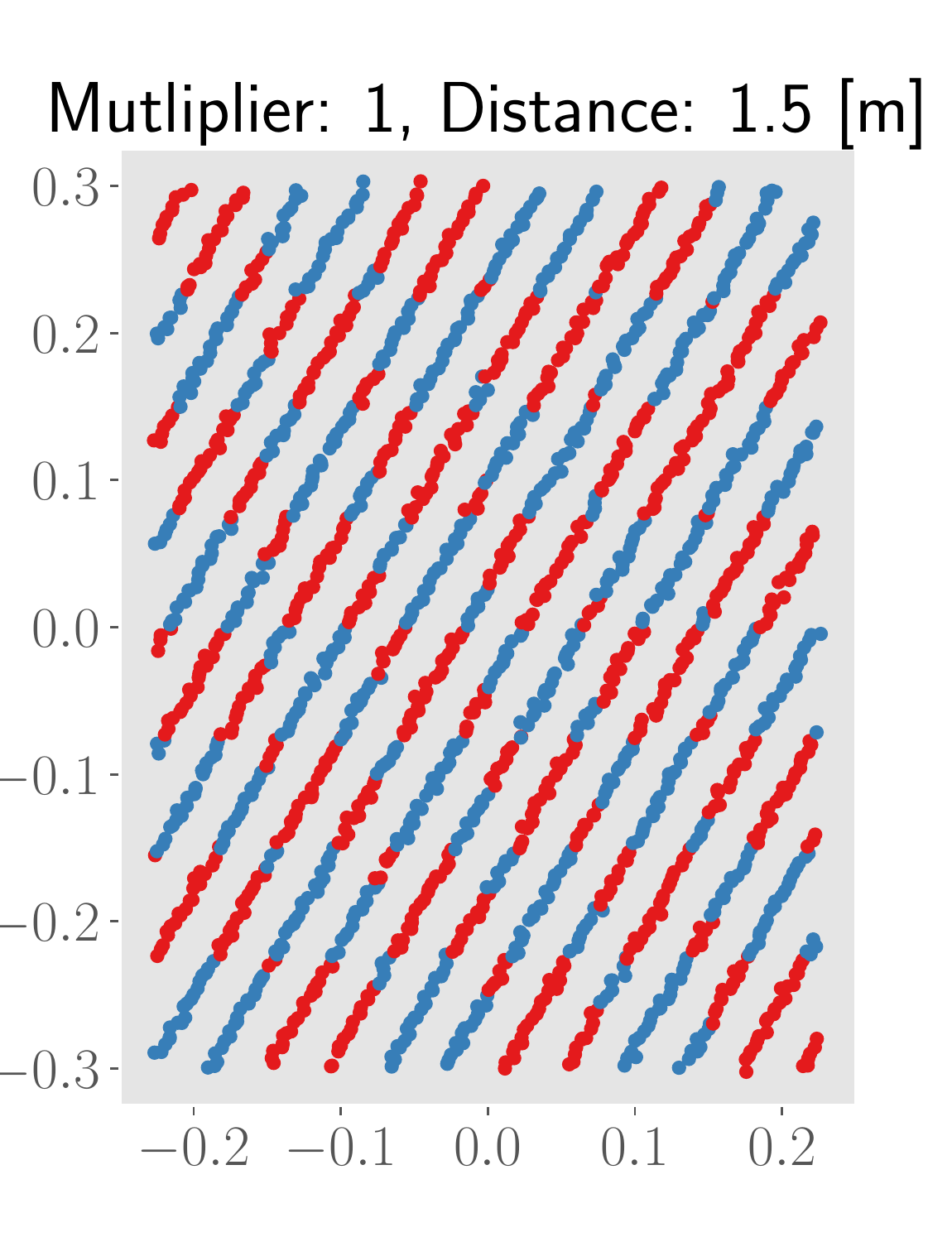}
            \hspace{-0.25cm}
            \includegraphics[height=3.5cm]{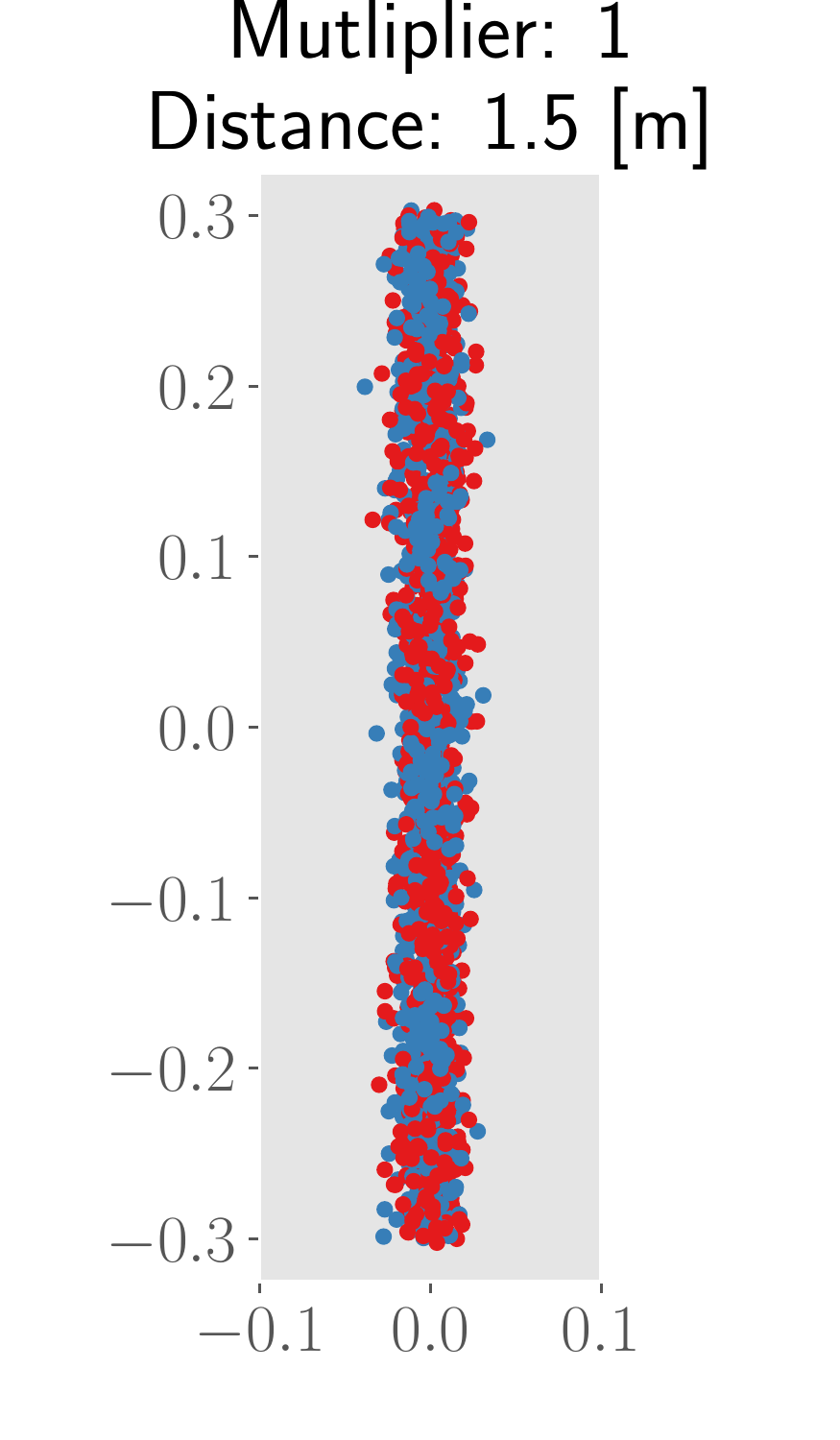}
        \caption{}
        \label{fig:fig:sim_chessboard:a}
        \end{subfigure}
        \hspace{-1.7cm}
        \begin{subfigure}[h]{.4\textwidth}
        \centering
            \includegraphics[height=3.5cm]{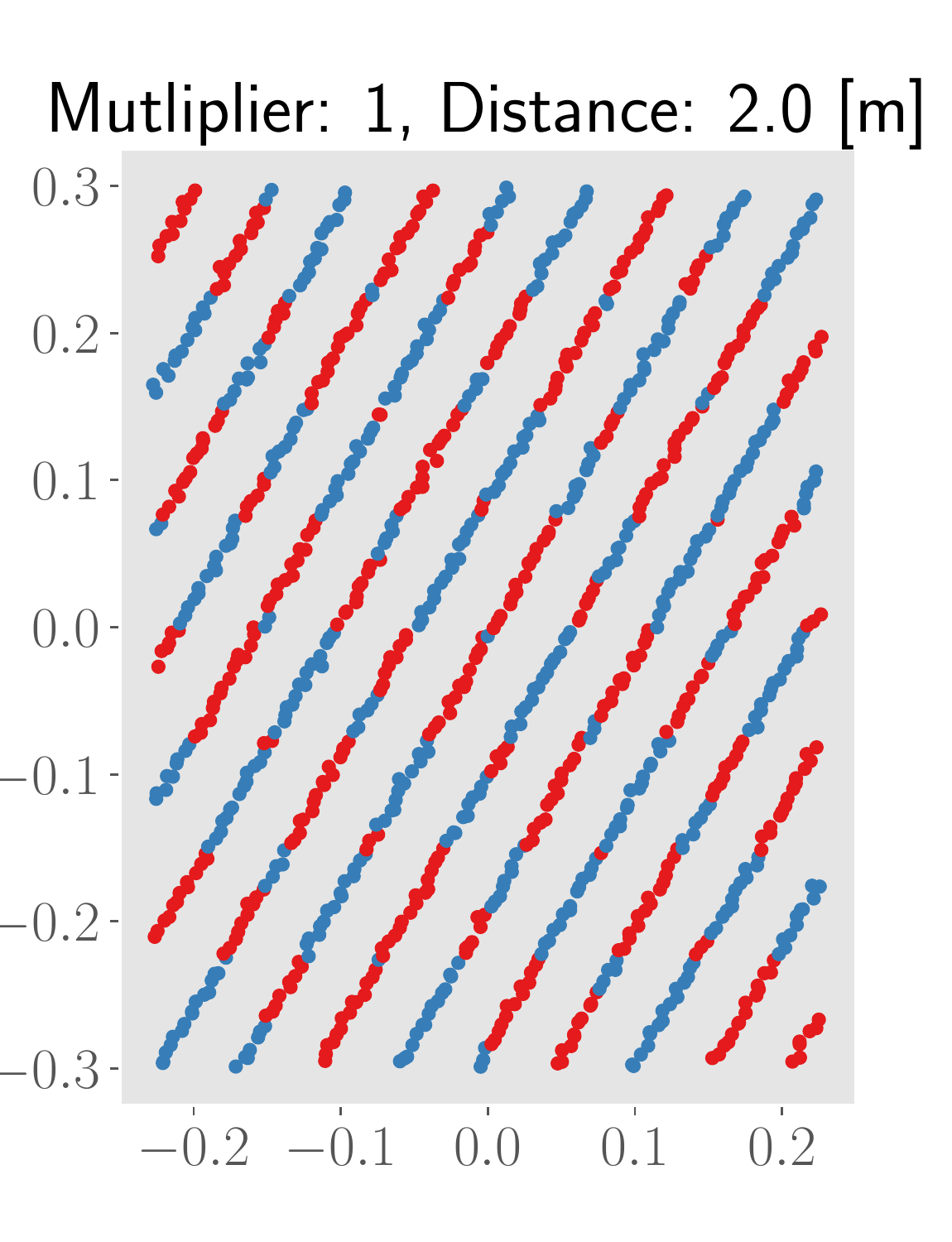}
            \hspace{-0.25cm}
            \includegraphics[height=3.5cm]{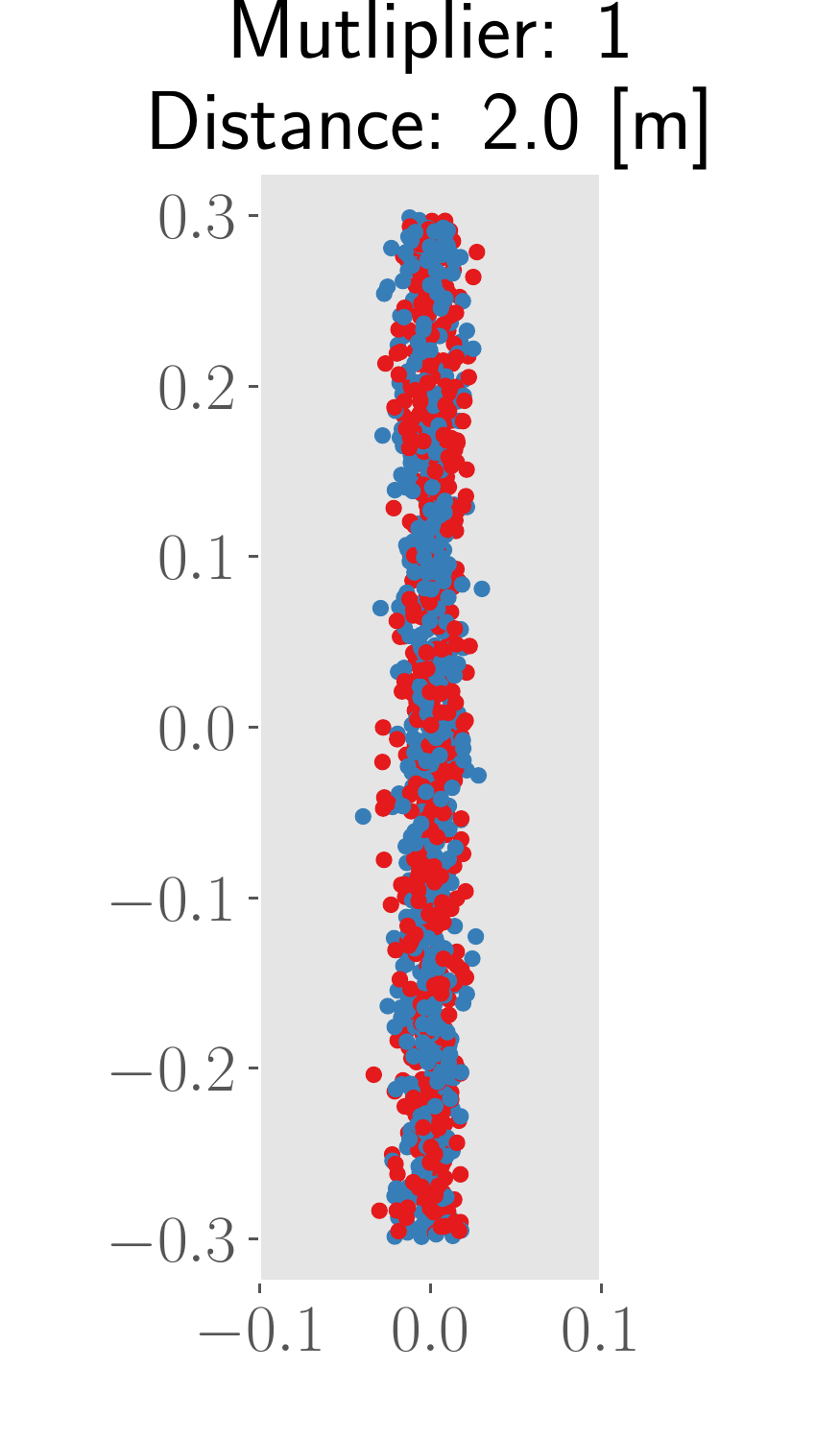}
        \caption{}
        \label{fig:fig:sim_chessboard:b}   
        \end{subfigure}
        \hspace{-1.7cm}
        \begin{subfigure}[h]{.4\textwidth}
        \centering
            \includegraphics[height=3.5cm]{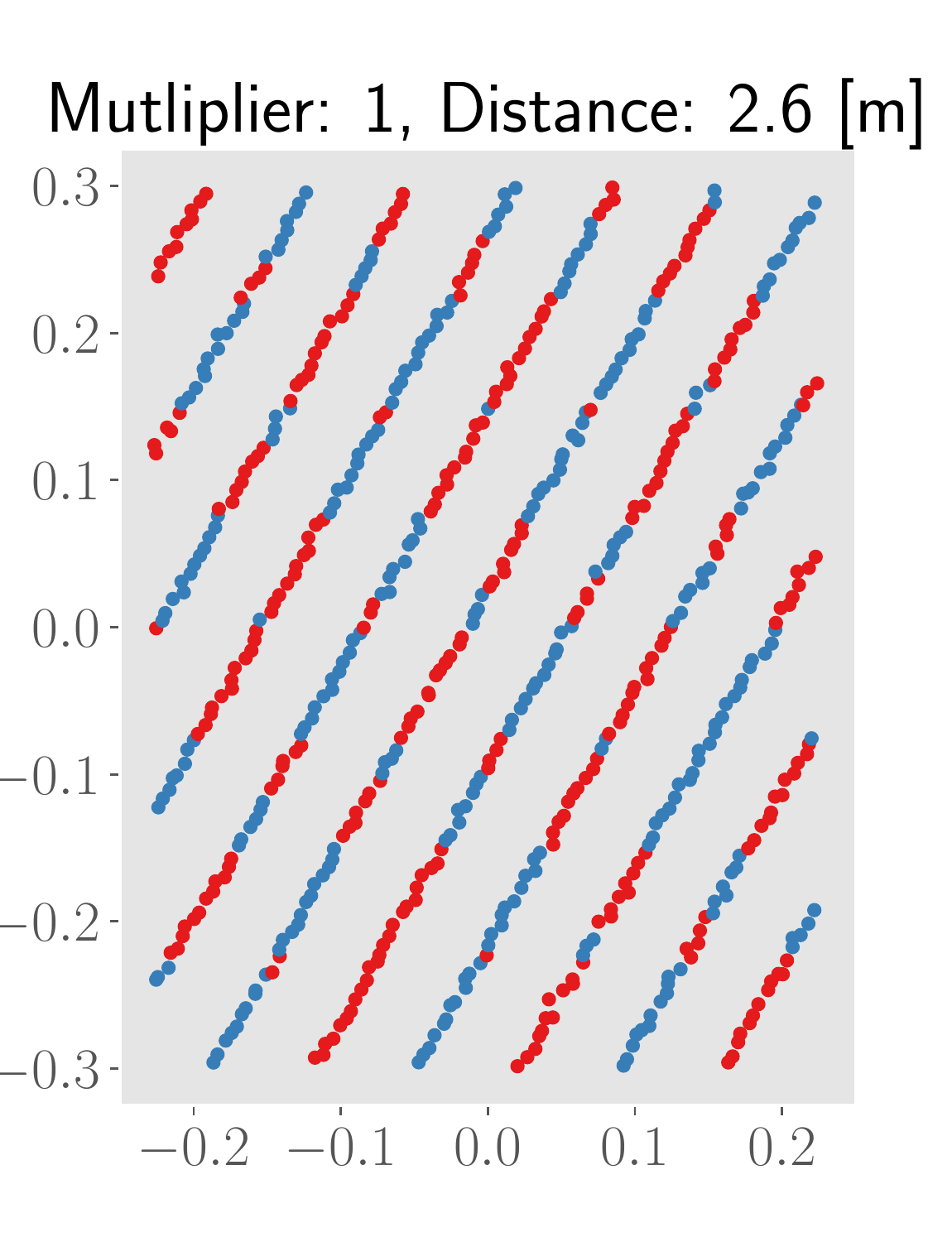}
            \hspace{-0.25cm}
            \includegraphics[height=3.5cm]{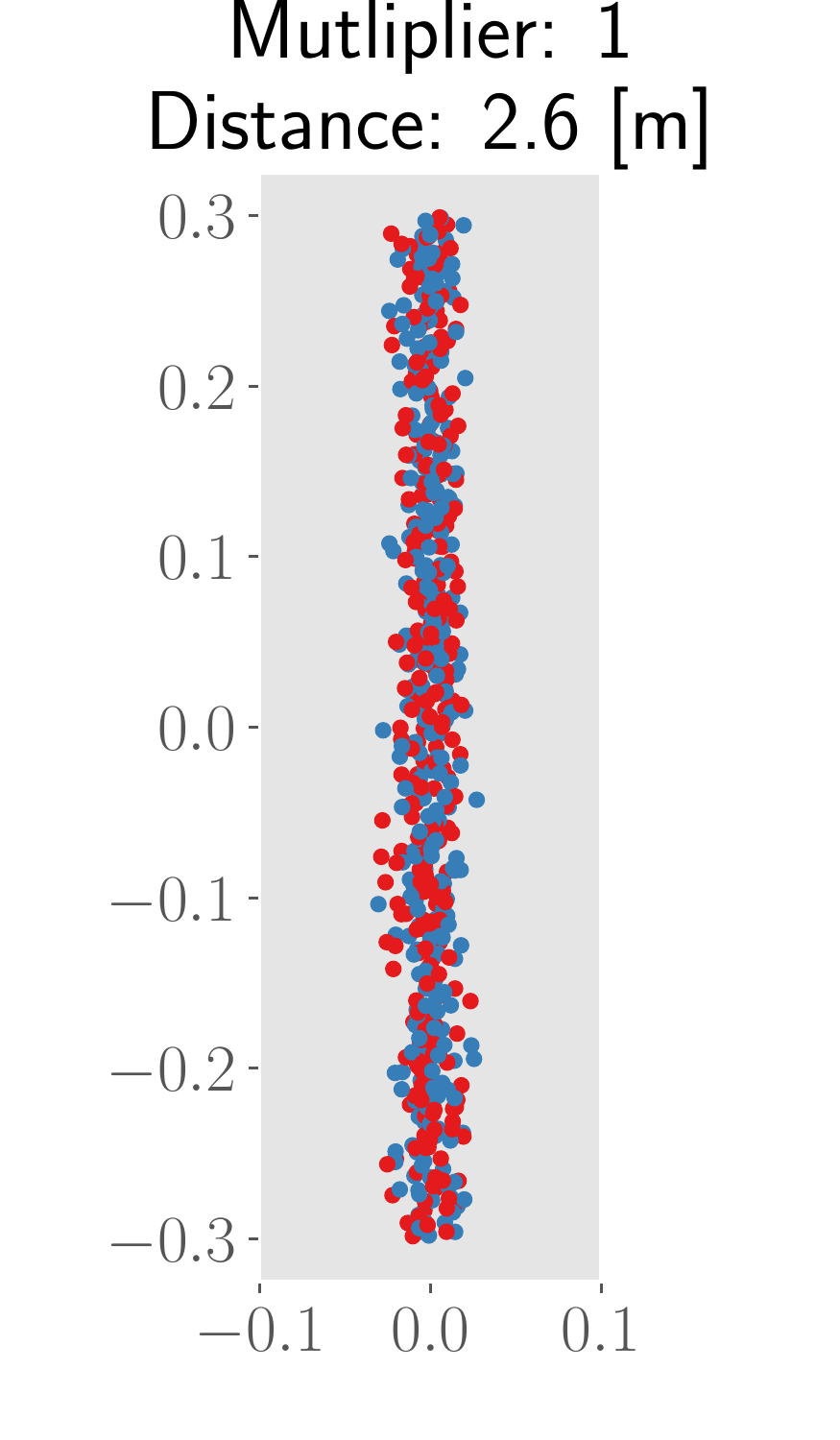}
        \caption{}
        \label{fig:fig:sim_chessboard:c}
        \end{subfigure}
        }     
         \mbox{
        \begin{subfigure}[h]{.4\textwidth}
        \centering
            \includegraphics[height=3.5cm]{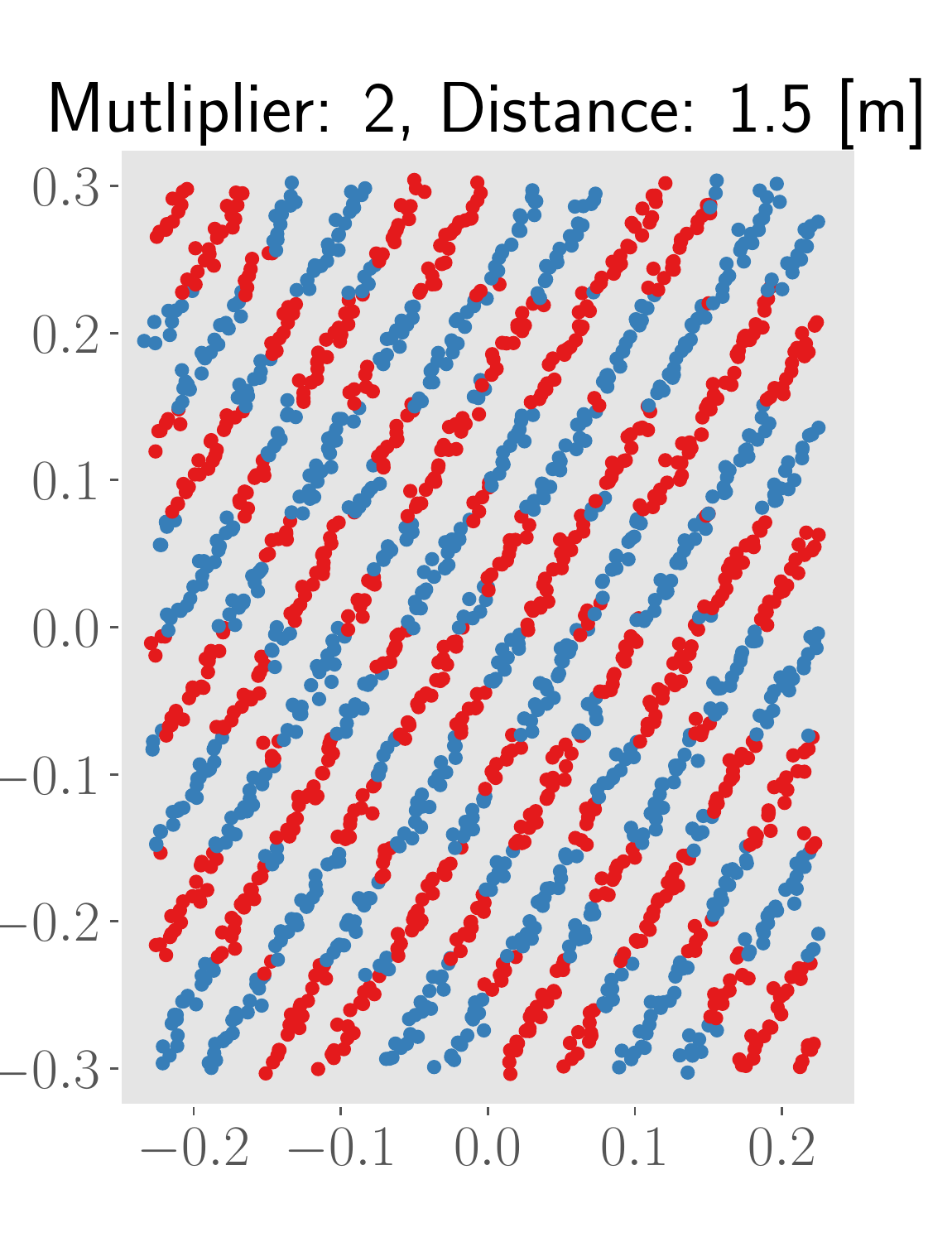}
            \hspace{-0.25cm}
            \includegraphics[height=3.5cm]{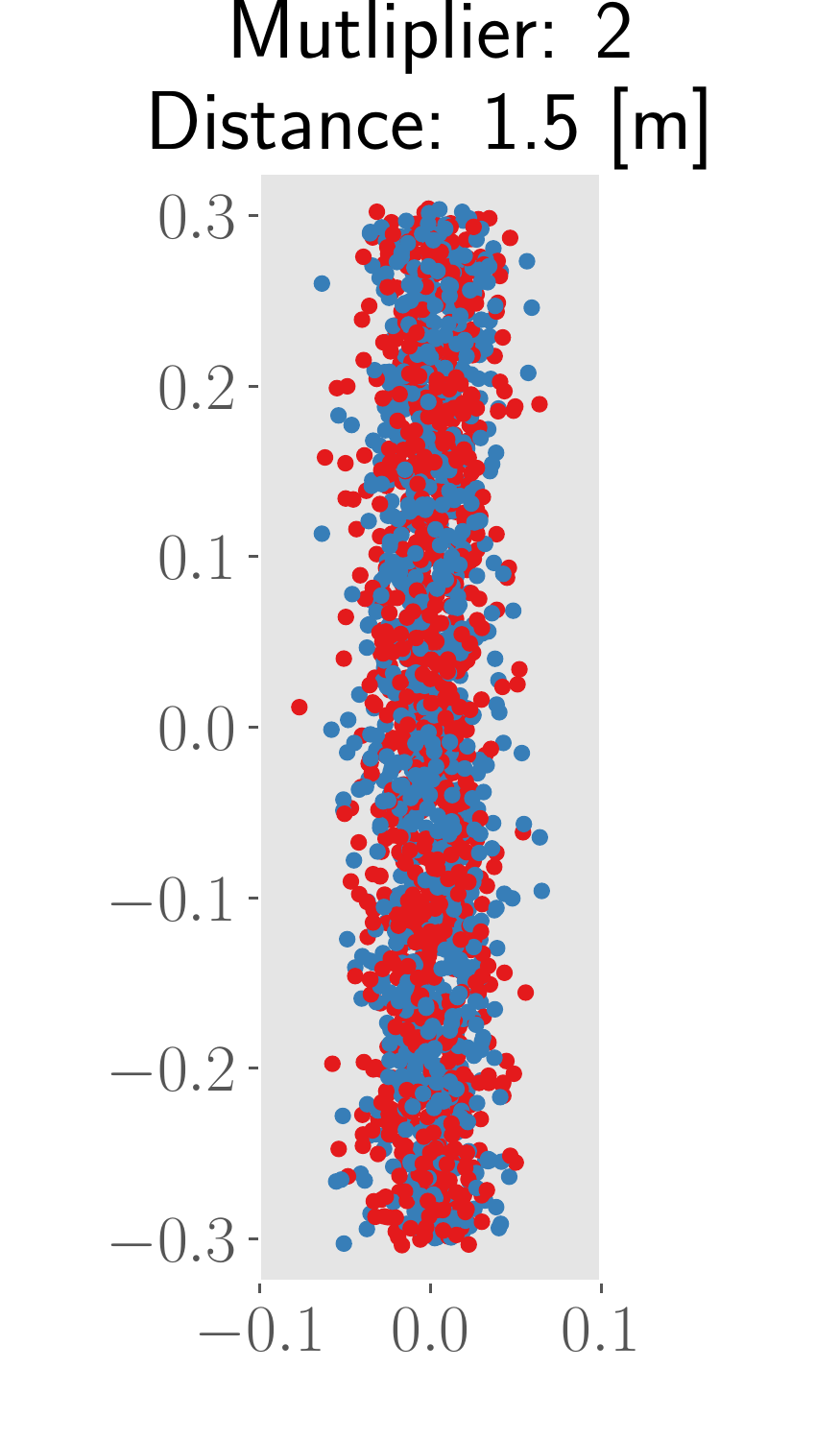}
        \caption{}
        \label{fig:fig:sim_chessboard:d}
        \end{subfigure}
        \hspace{-1.7cm}
        \begin{subfigure}[h]{.4\textwidth}
        \centering
            \includegraphics[height=3.5cm]{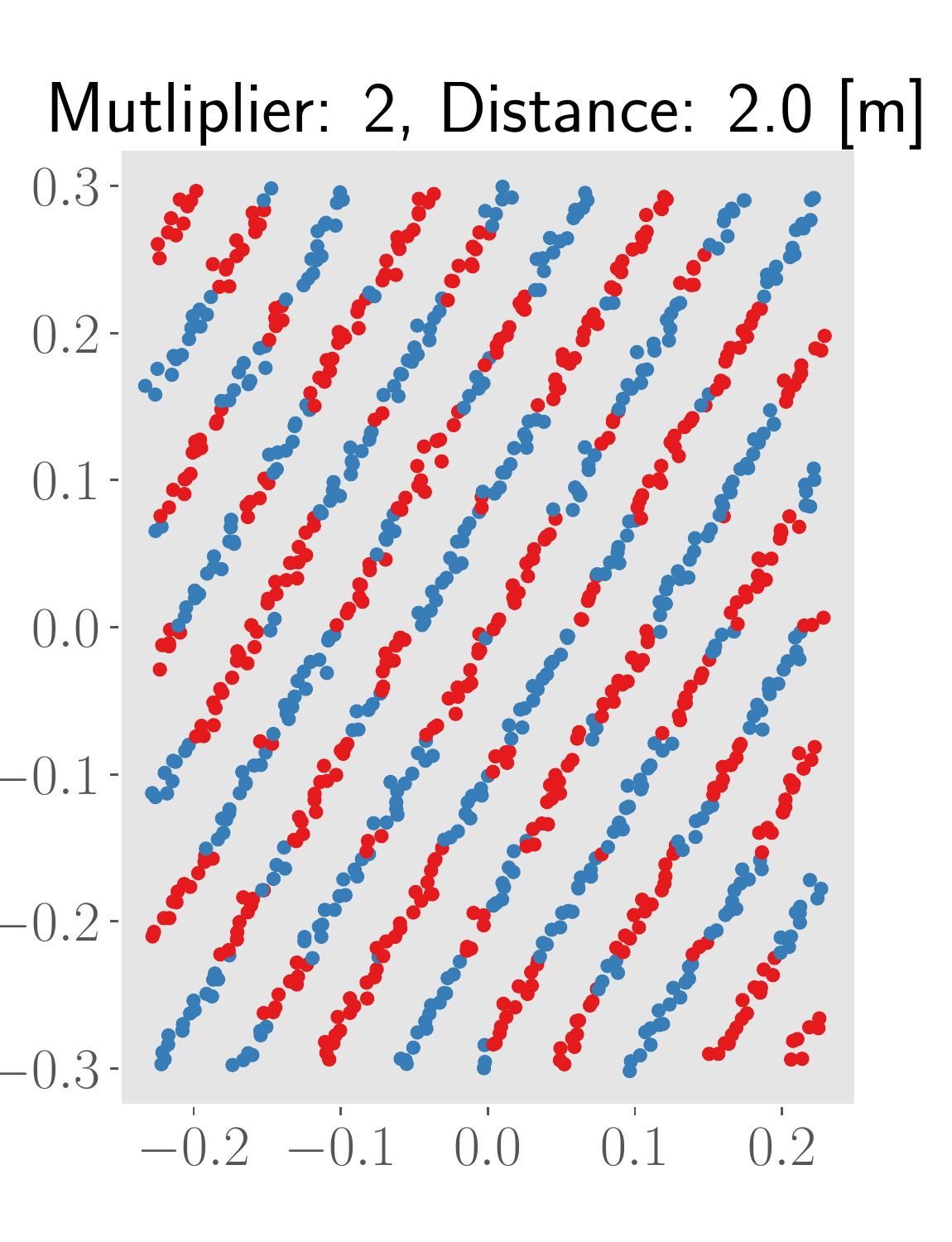}
            \hspace{-0.25cm}
            \includegraphics[height=3.5cm]{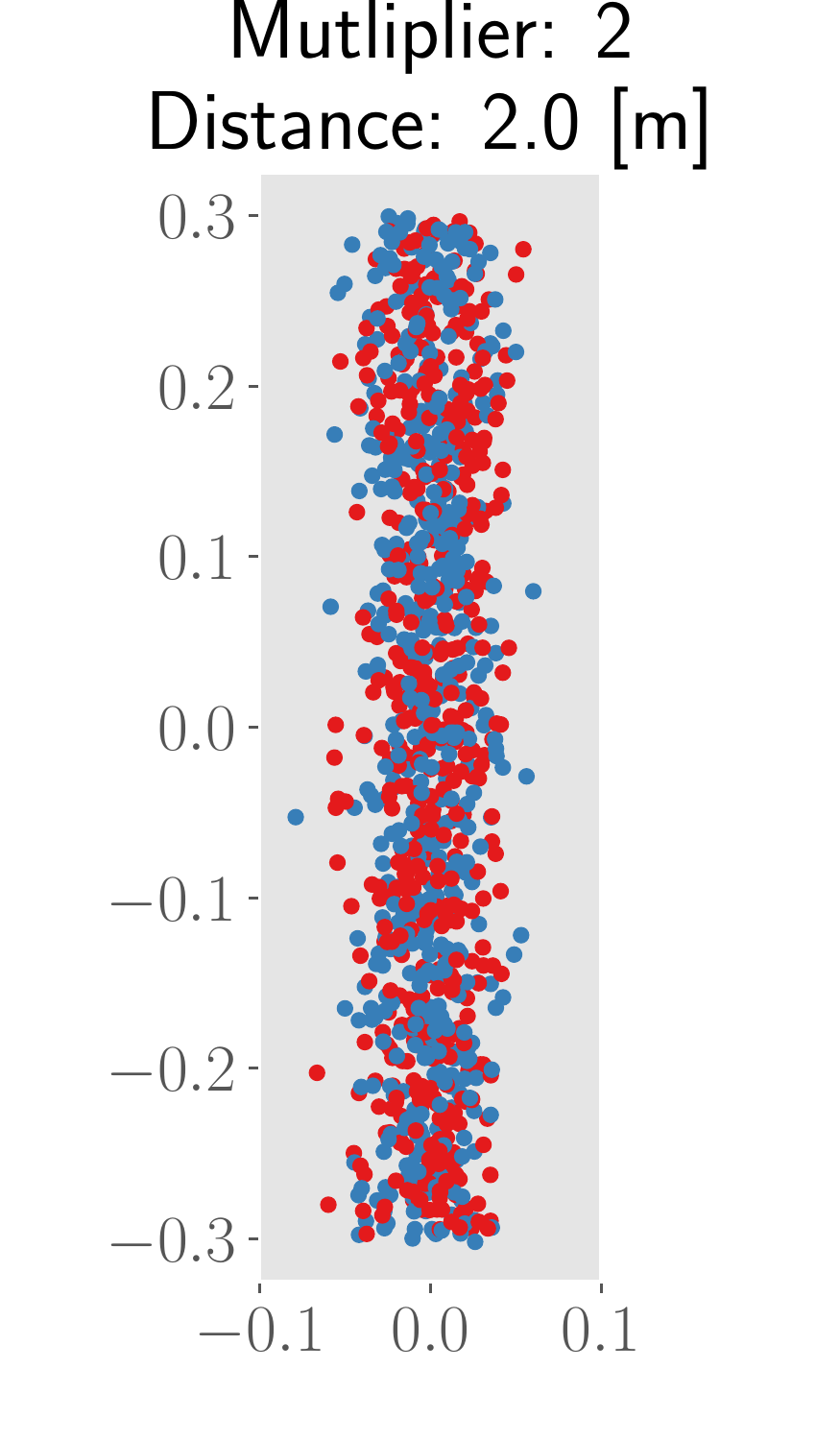}
        \caption{}
        \label{fig:fig:sim_chessboard:e}
        \end{subfigure}
        \hspace{-1.7cm}
        \begin{subfigure}[h]{.4\textwidth}
        \centering
            \includegraphics[height=3.5cm]{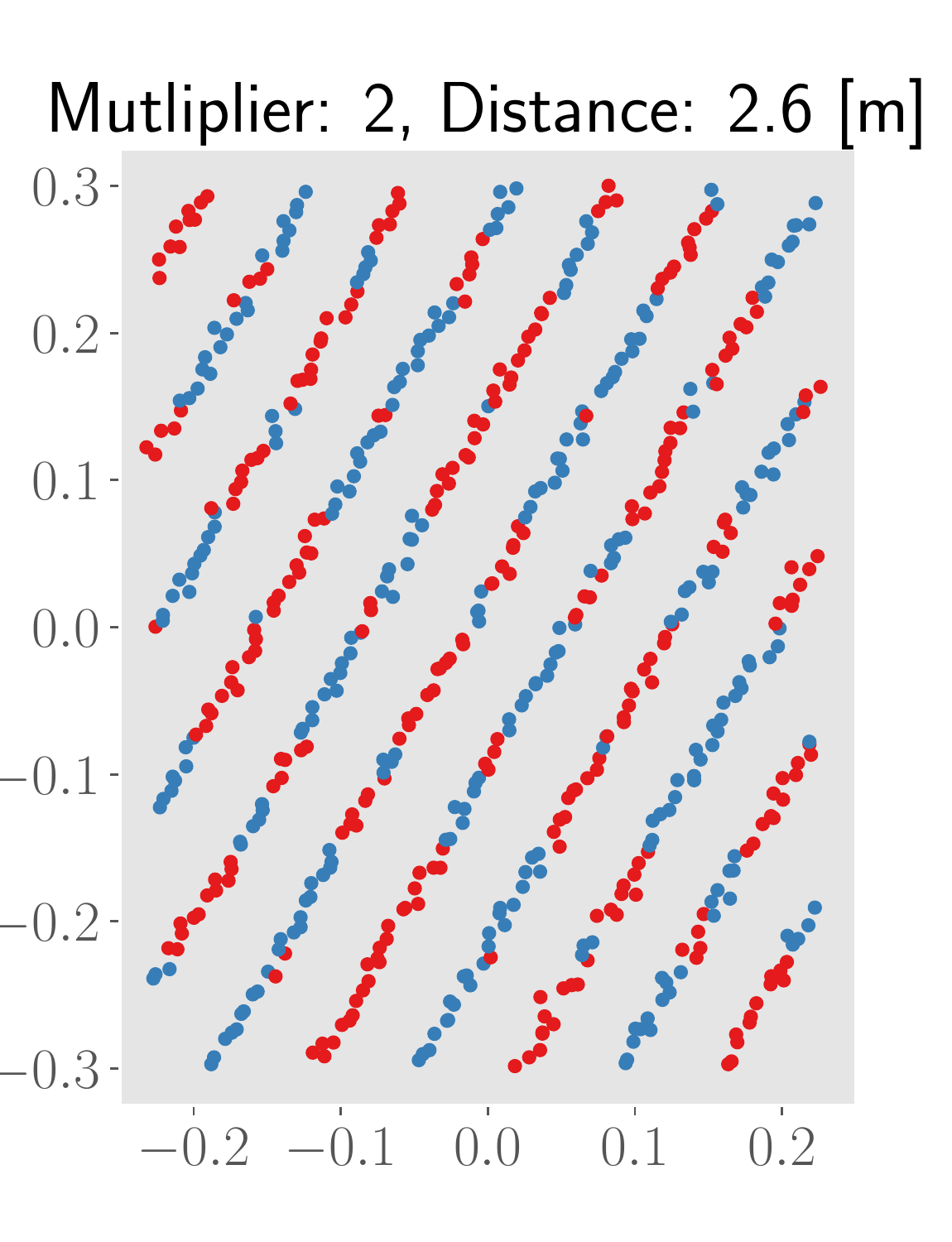}
            \hspace{-0.25cm}
            \includegraphics[height=3.5cm]{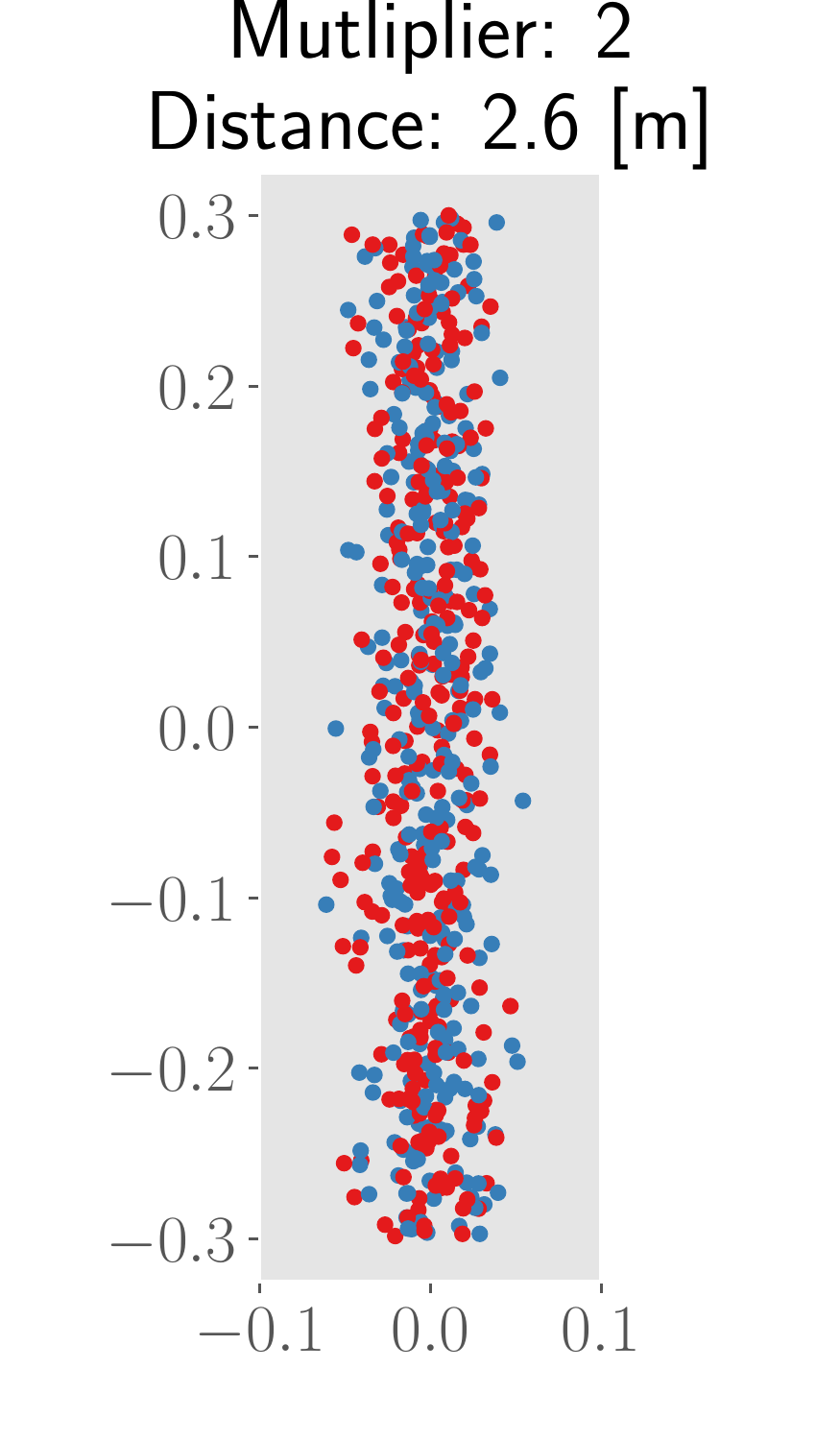}
        \caption{}
        \label{fig:fig:sim_chessboard:f}
        \end{subfigure}
        }
      \mbox{
        \begin{subfigure}[h]{.4\textwidth}
        \centering
            \includegraphics[height=3.5cm]{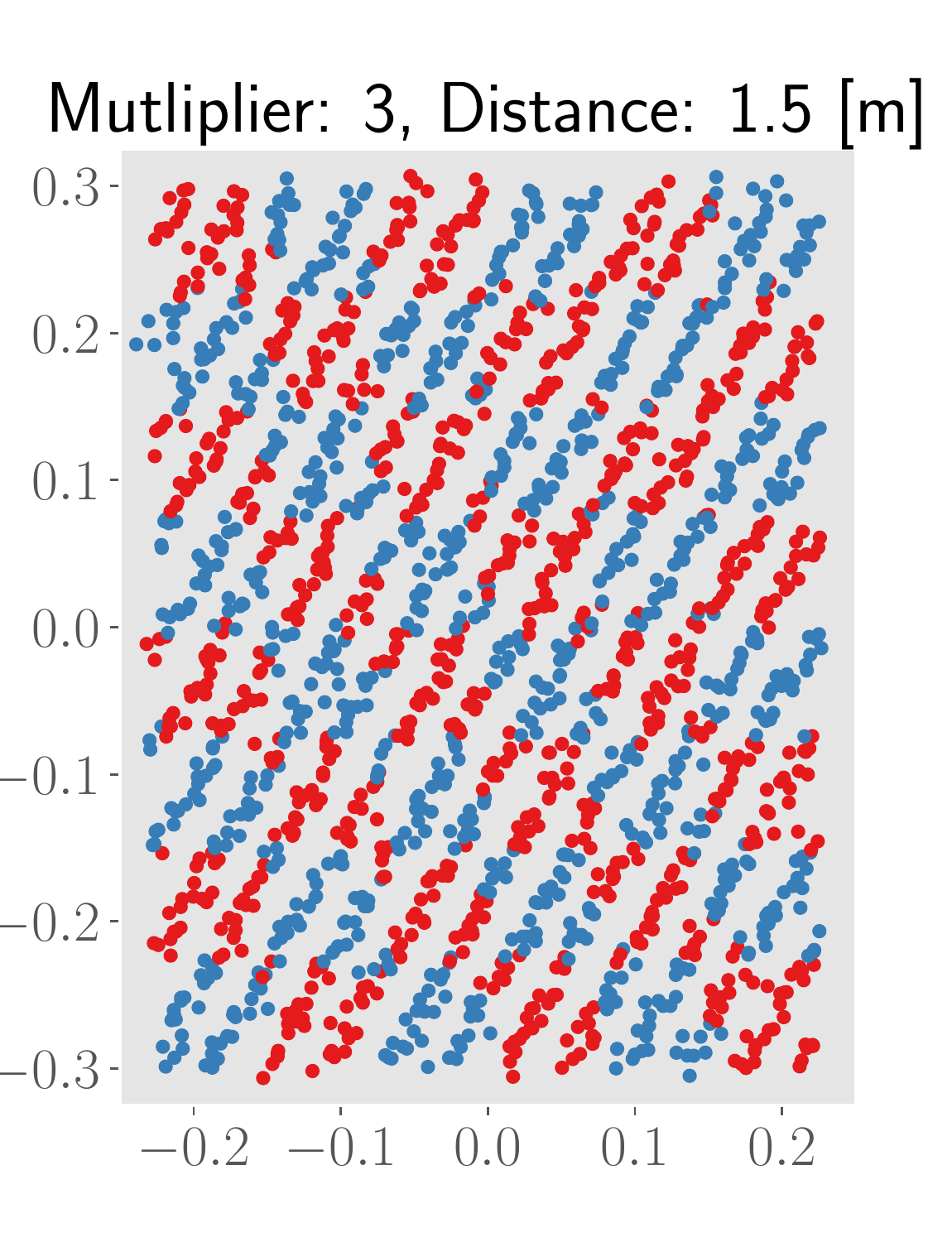}
            \hspace{-0.25cm}
            \includegraphics[height=3.5cm]{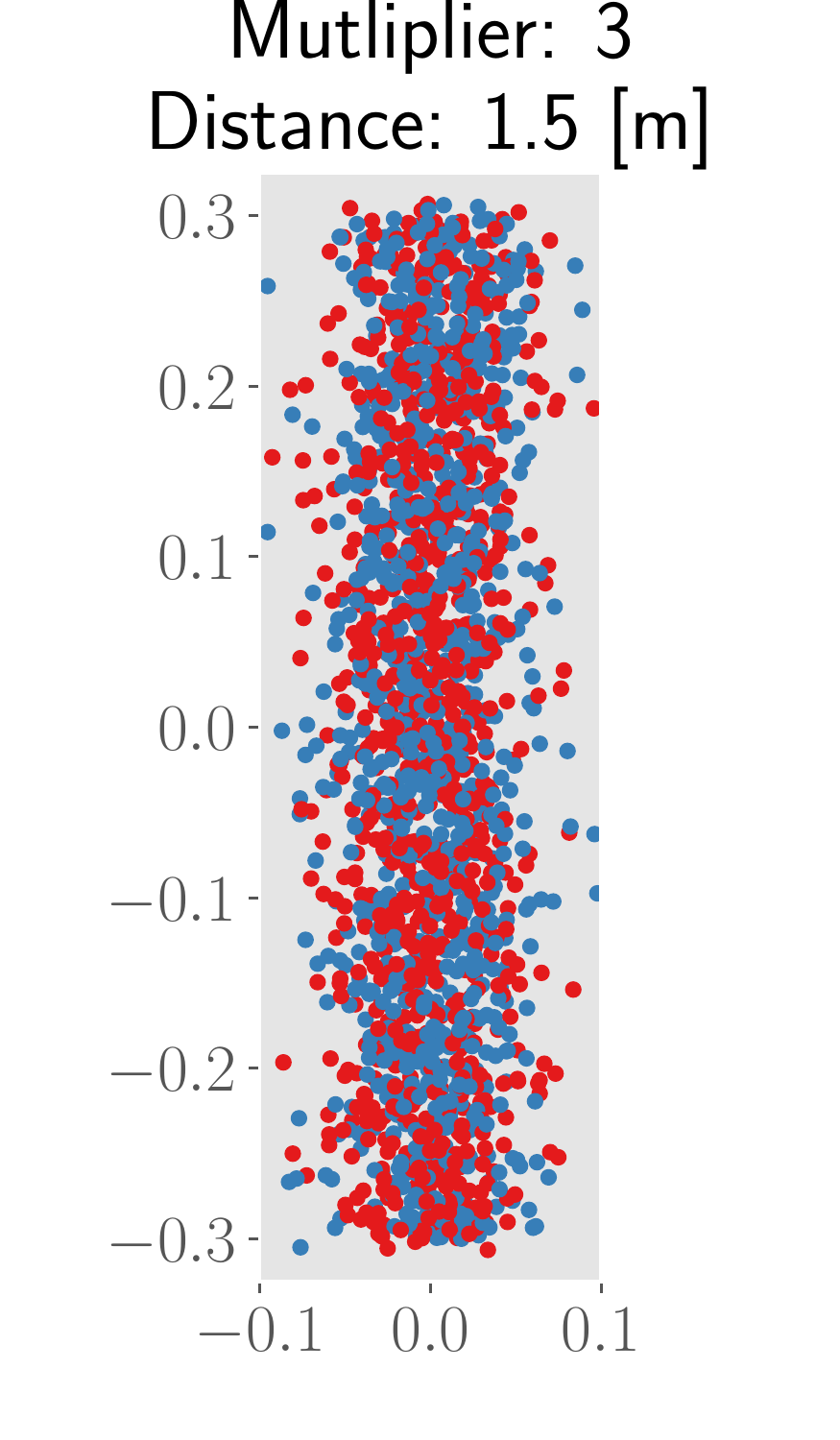}
        \caption{}
        \label{fig:fig:sim_chessboard:g}
        \end{subfigure}
        \hspace{-1.7cm}
        \begin{subfigure}[h]{.4\textwidth}
        \centering
            \includegraphics[height=3.5cm]{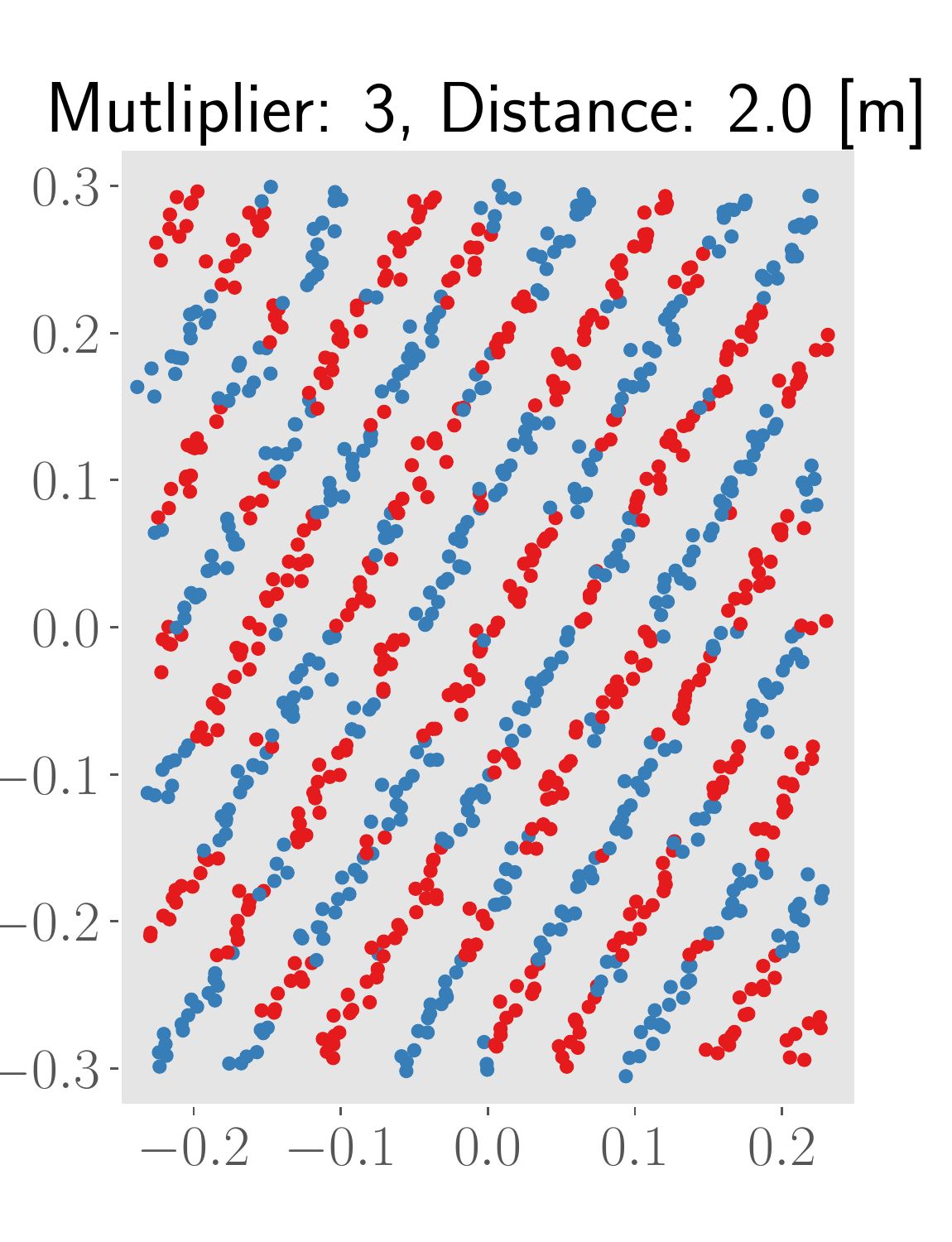}
            \hspace{-0.25cm}
            \includegraphics[height=3.5cm]{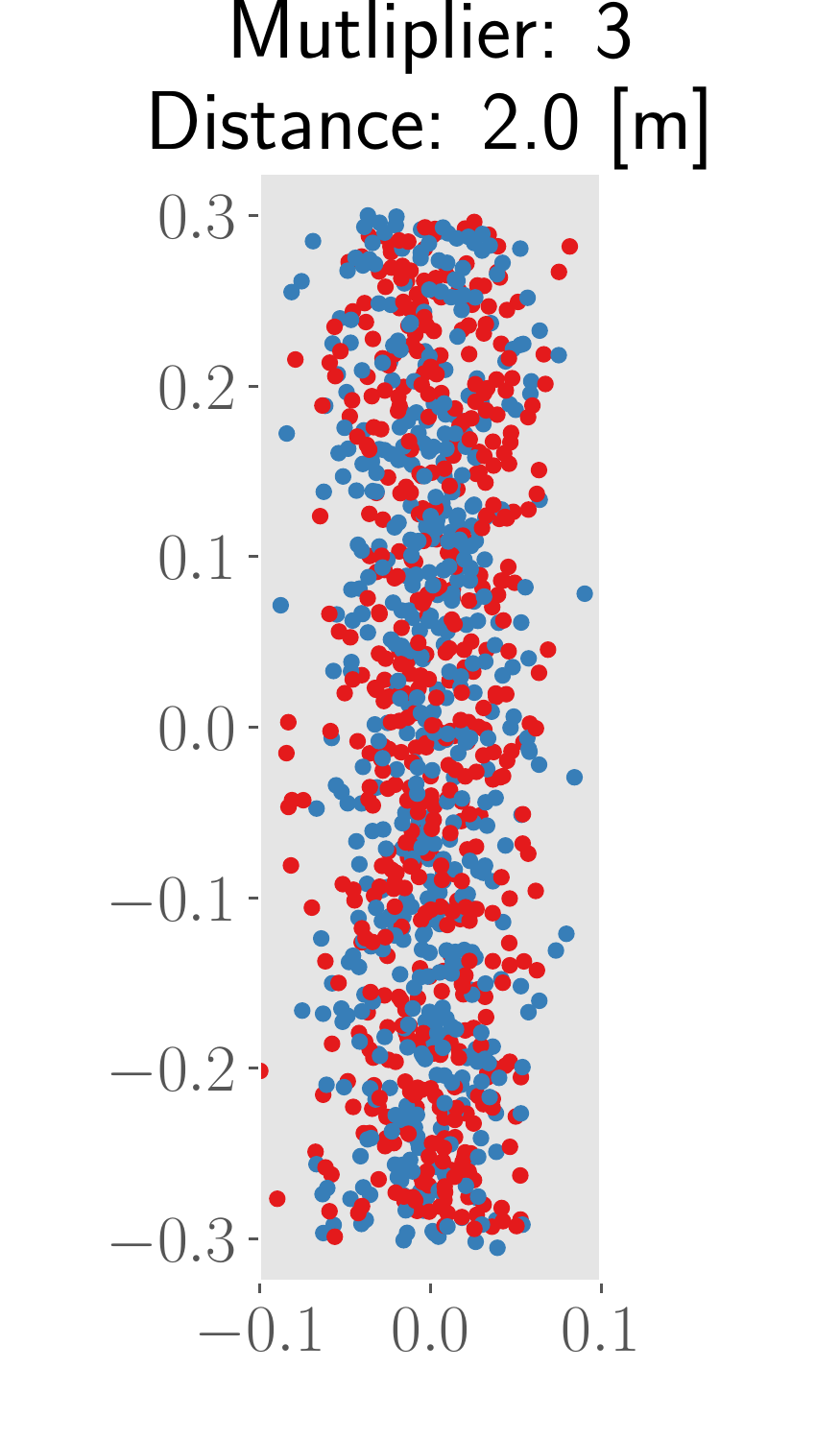}
        \caption{}
        \label{fig:fig:sim_chessboard:h}
        \end{subfigure}
        \hspace{-1.7cm}
        \begin{subfigure}[h]{.4\textwidth}
        \centering
            \includegraphics[height=3.5cm]{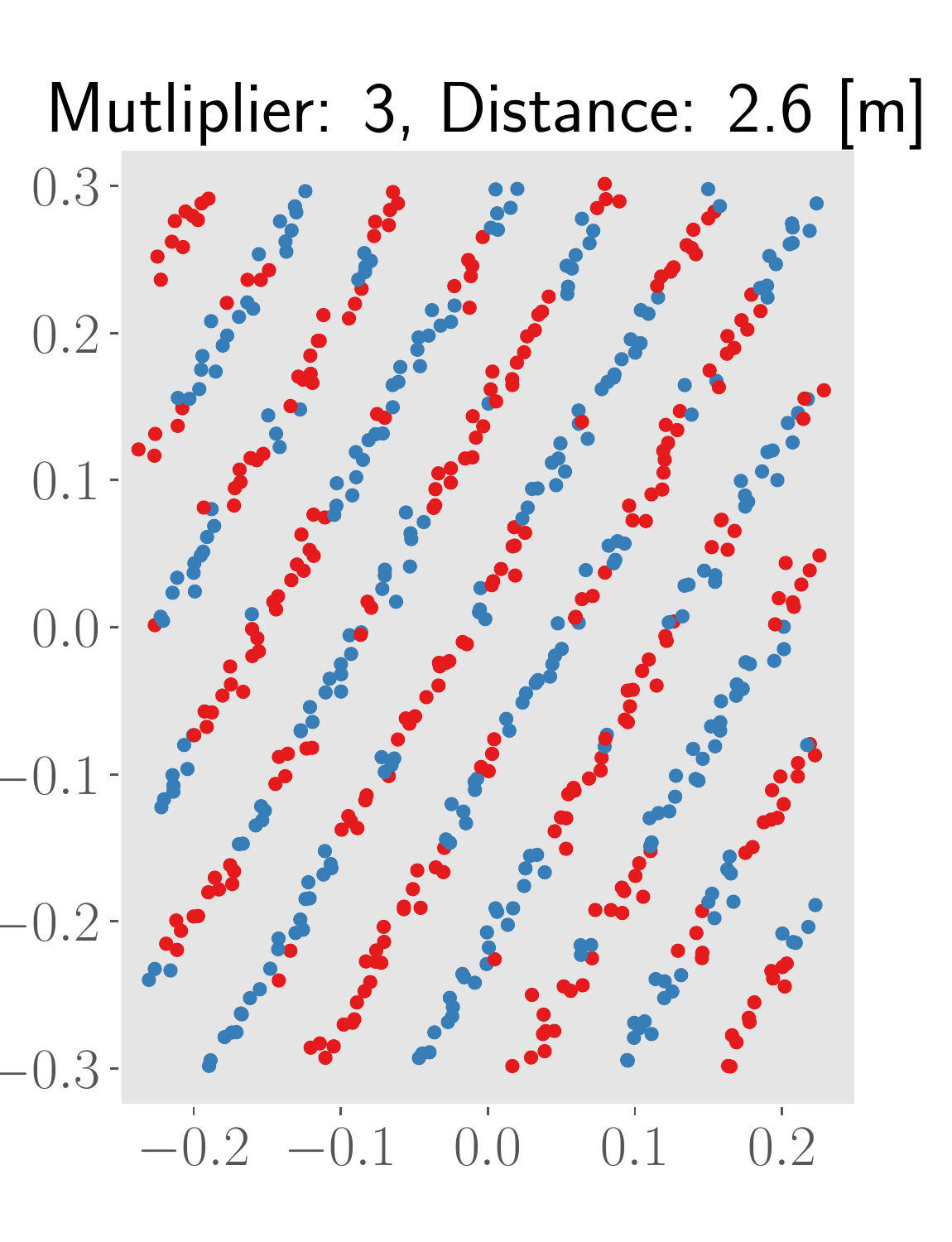}
            \hspace{-0.25cm}
            \includegraphics[height=3.5cm]{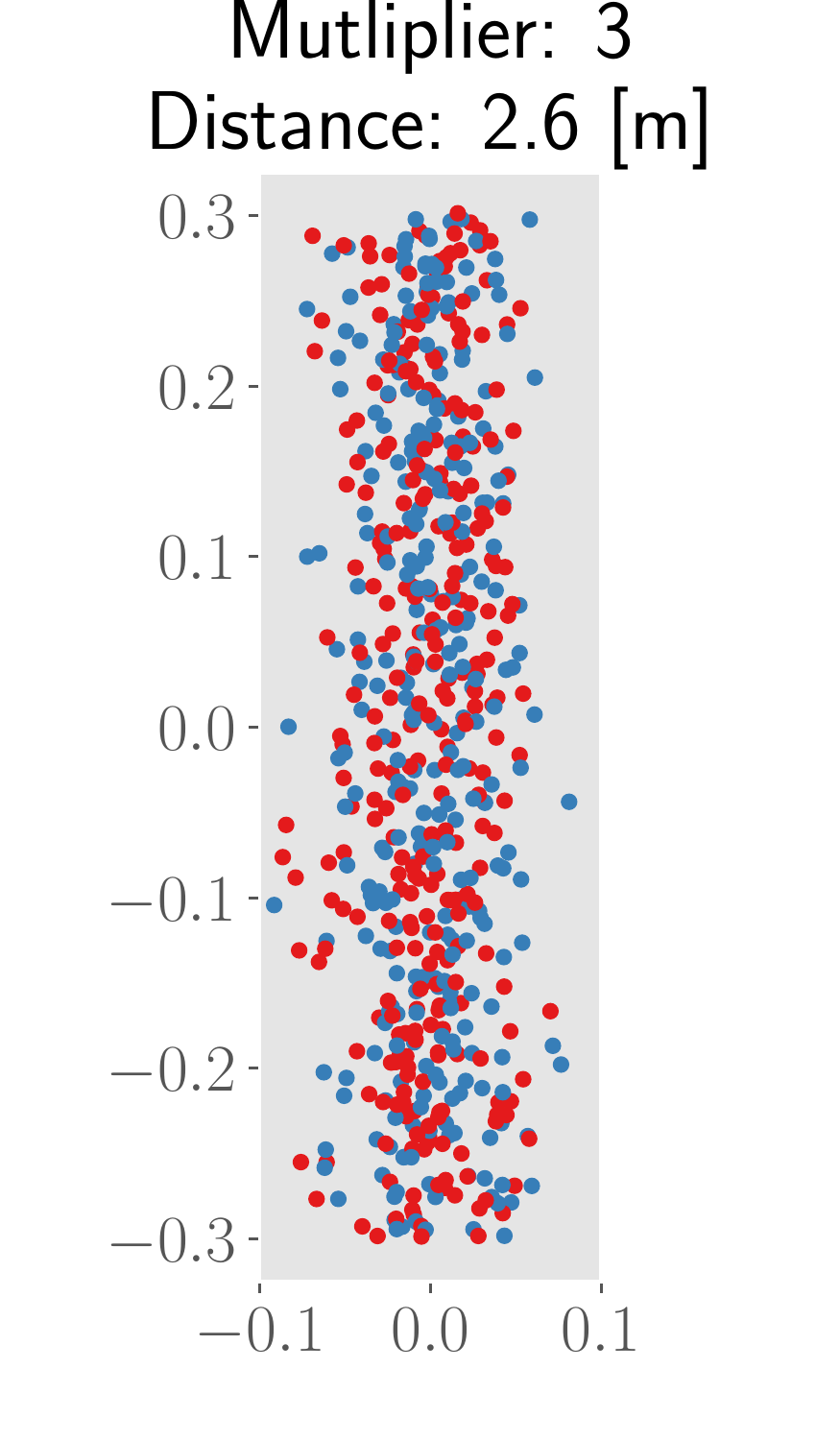}
        \caption{}
        \label{fig:fig:sim_chessboard:i}
        \end{subfigure}
        }
            \mbox{
        \begin{subfigure}[h]{.4\textwidth}
        \centering
            \includegraphics[height=3.5cm]{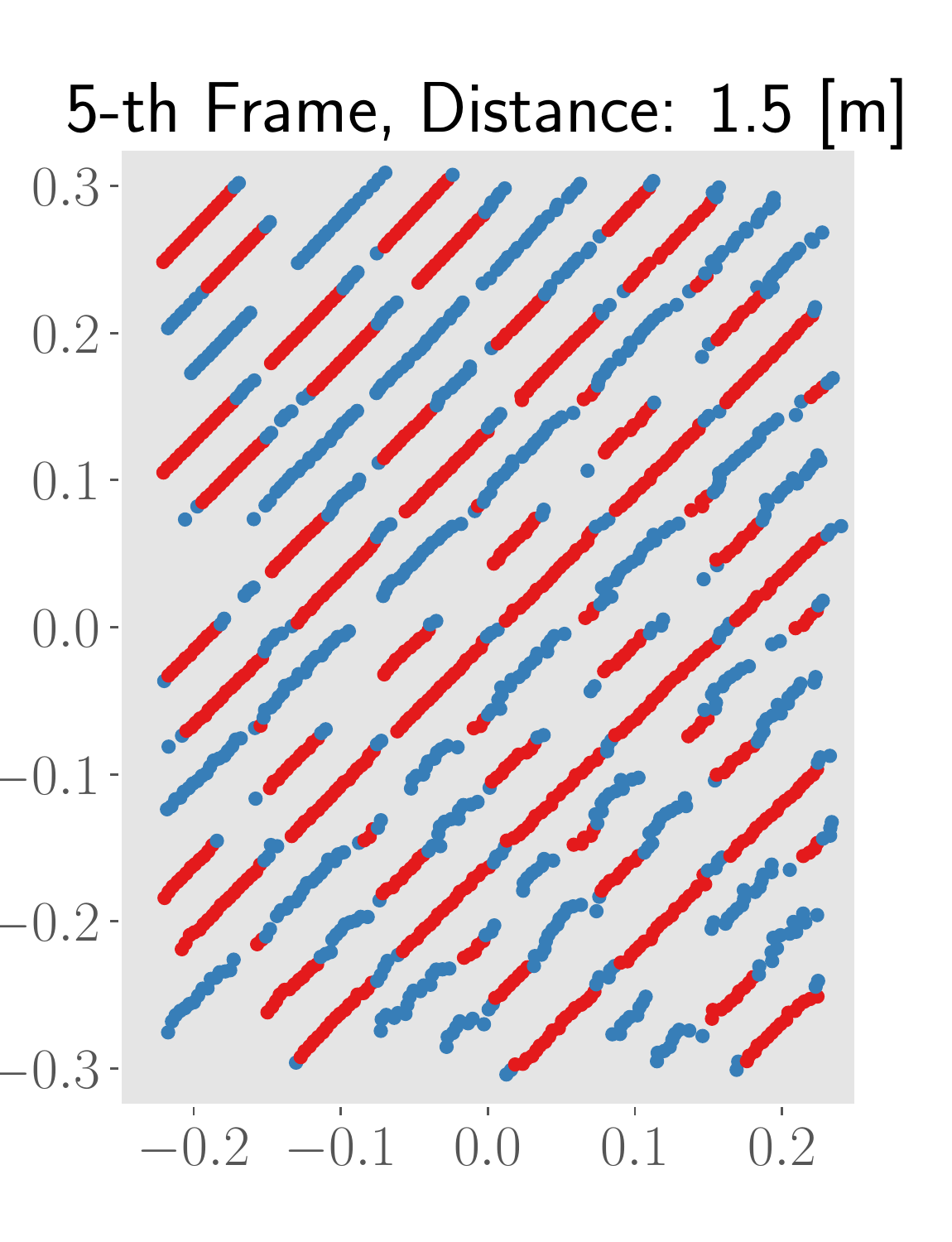}
            \hspace{-0.25cm}
            \includegraphics[height=3.5cm]{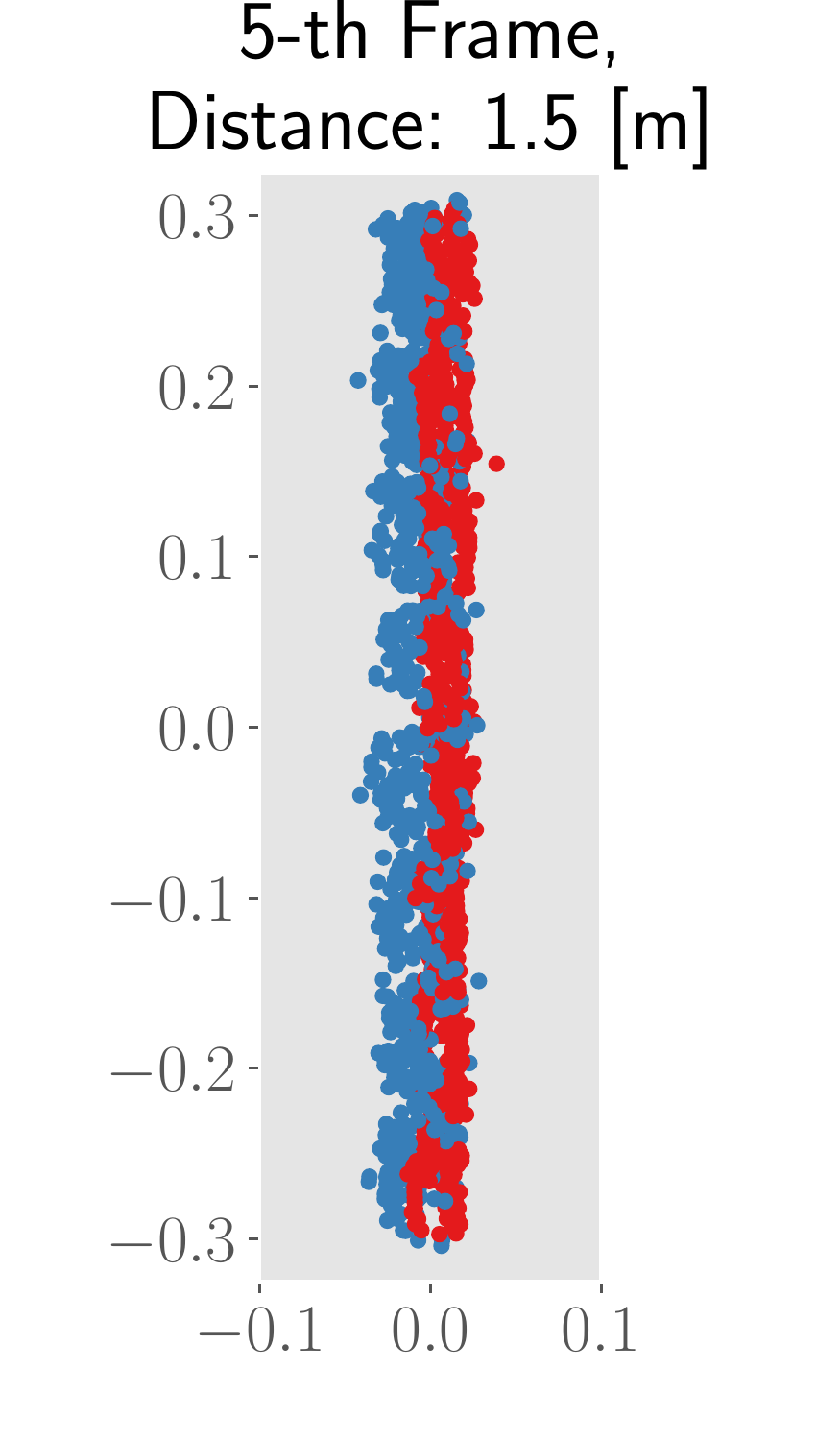}
        \caption{}
        \label{fig:fig:sim_chessboard:j}
        \end{subfigure}
        \hspace{-1.7cm}
        \begin{subfigure}[h]{.4\textwidth}
        \centering
            \includegraphics[height=3.5cm]{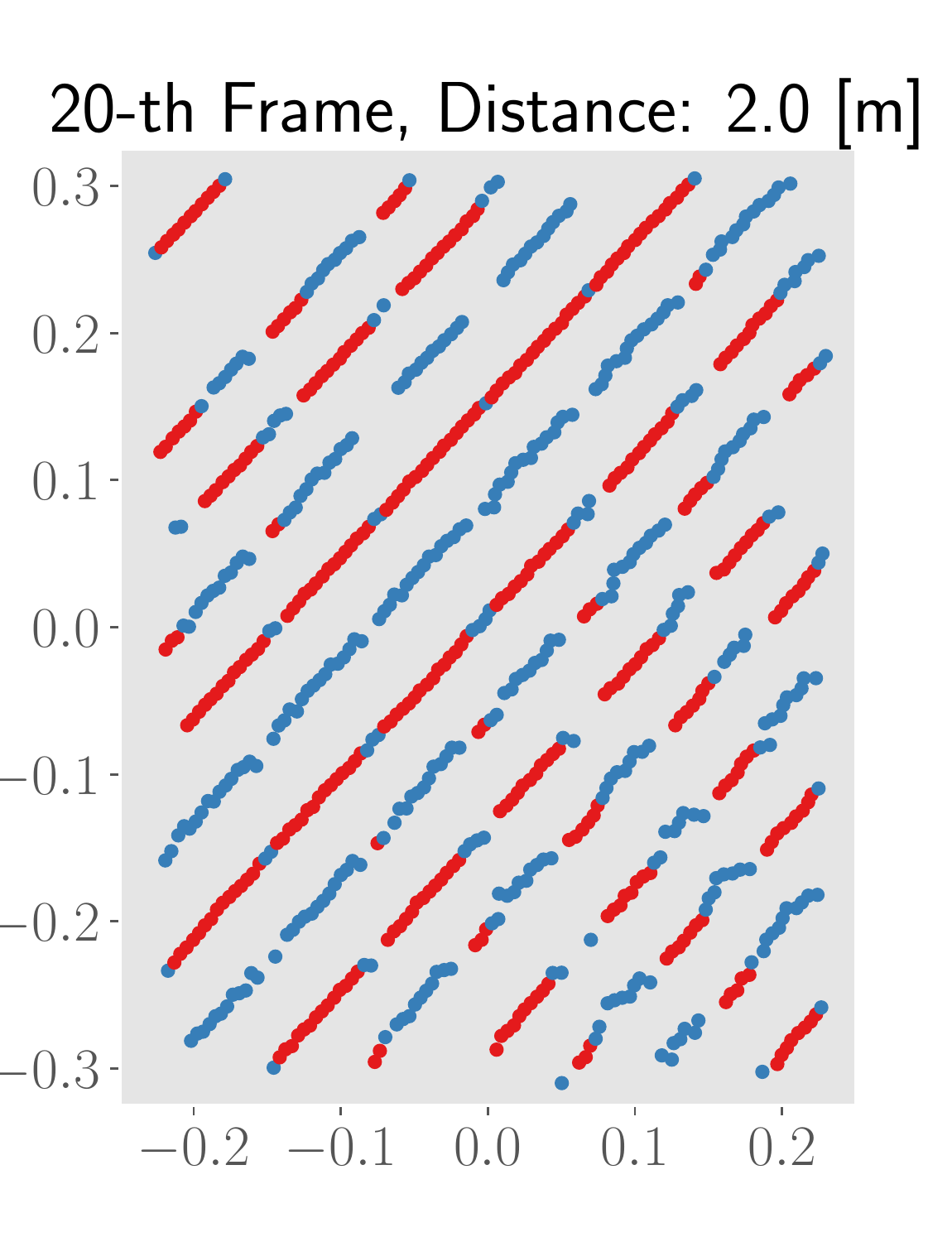}
            \hspace{-0.25cm}
            \includegraphics[height=3.5cm]{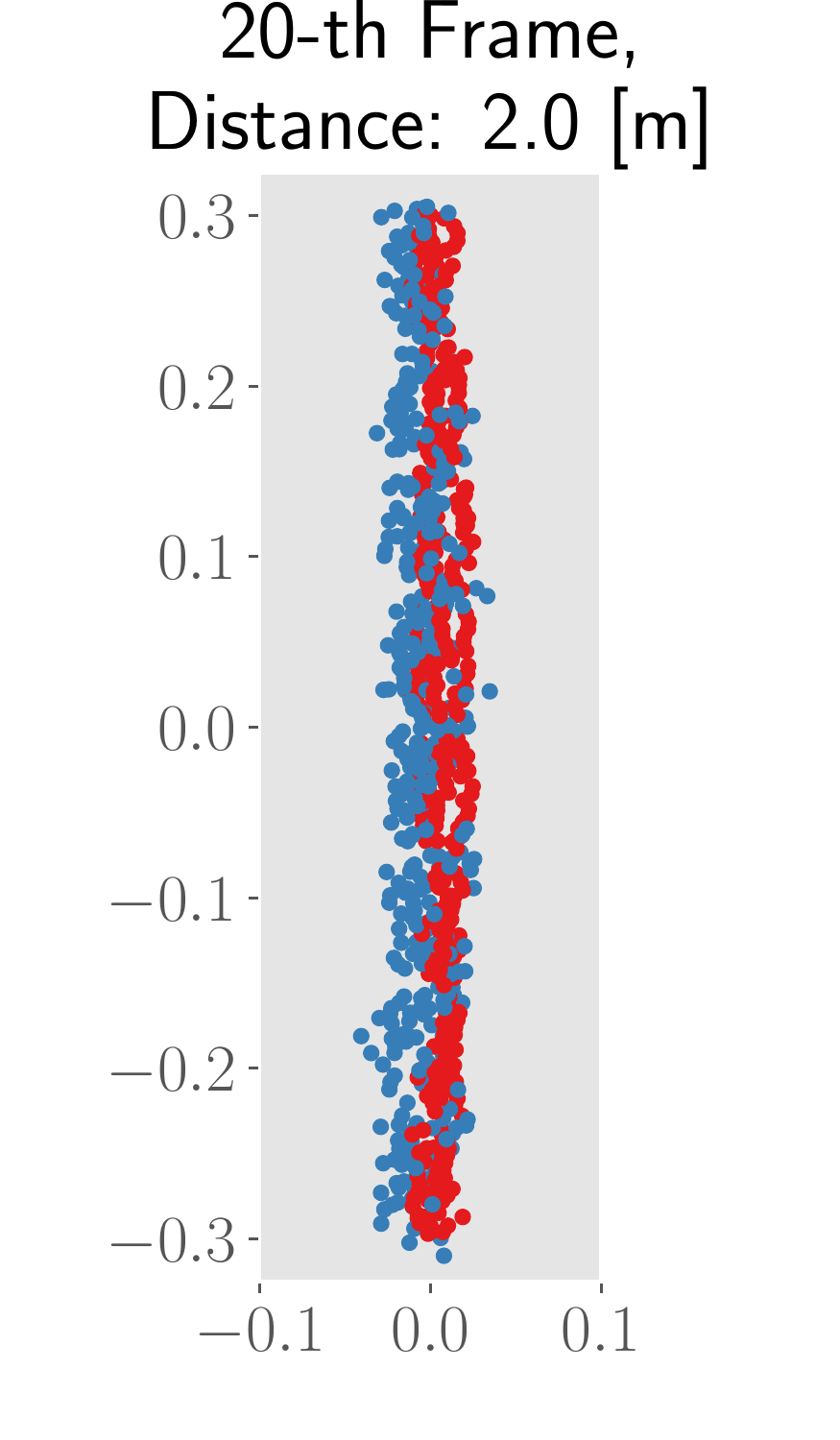}
        \caption{}
        \label{fig:fig:sim_chessboard:k}
        \end{subfigure}
        \hspace{-1.7cm}
        \begin{subfigure}[h]{.4\textwidth}
        \centering
            \includegraphics[height=3.5cm]{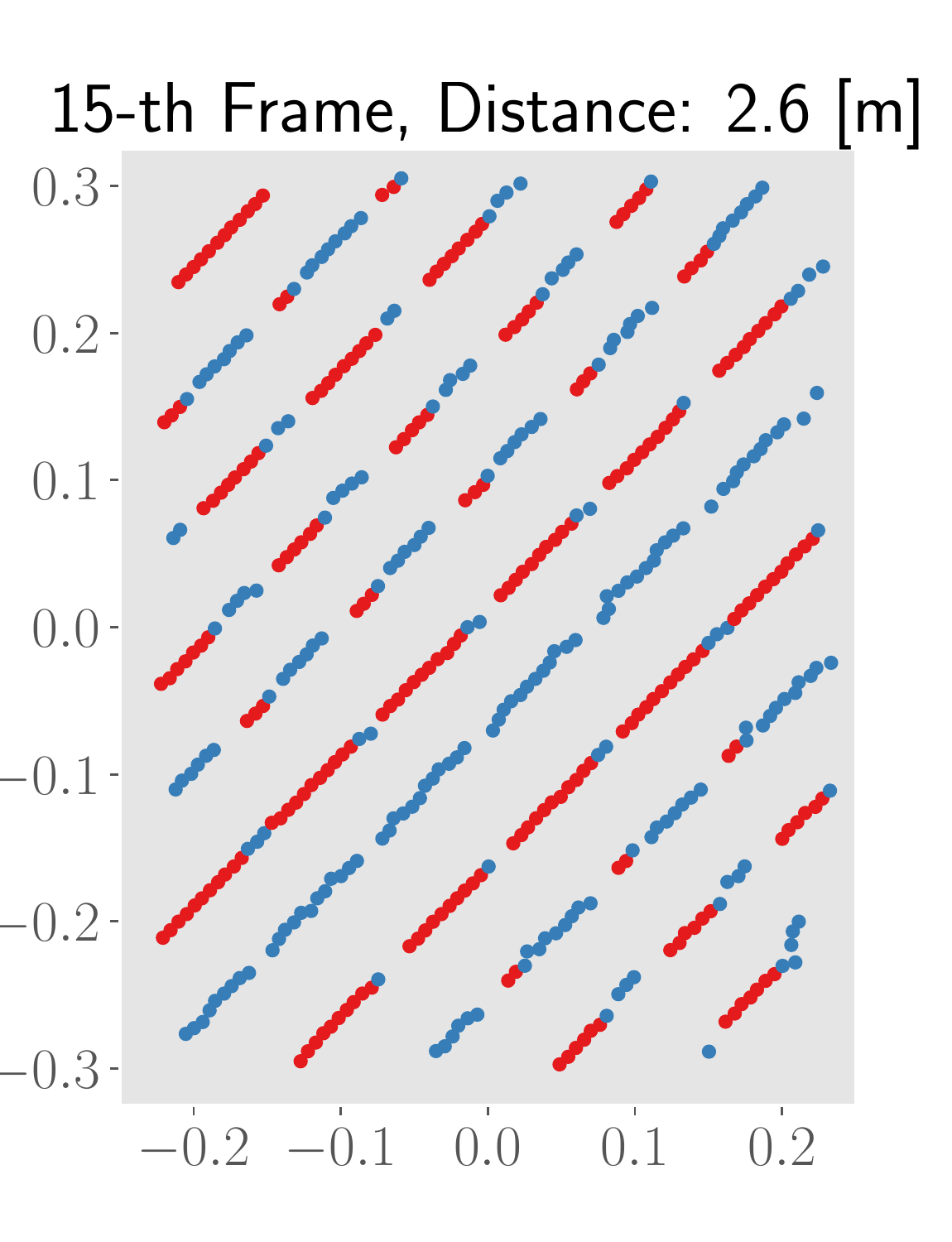}
            \hspace{-0.25cm}
            \includegraphics[height=3.5cm]{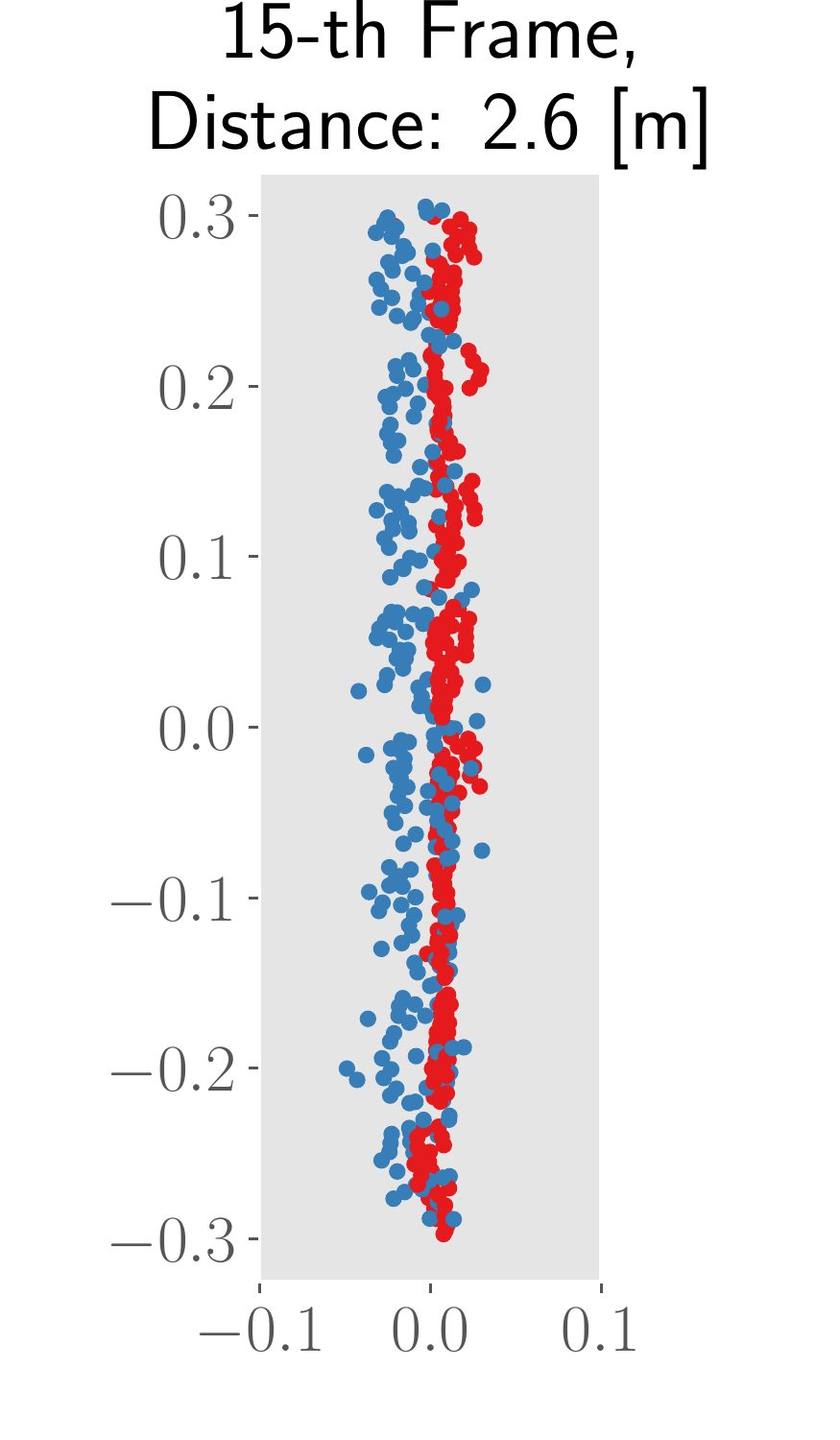}
        \caption{}
        \label{fig:fig:sim_chessboard:l}
        \end{subfigure}
        }  
\caption{Front view and side view of chessboard's point clouds from simulation results and real data. (\textbf{a}--\textbf{c}) Simulated point clouds with the multiplier 1 at different distances; (\textbf{d}--\textbf{f}) Simulated point clouds with the multiplier 2 at different distances; (\textbf{g}--\textbf{i}) Simulated point clouds with the multiplier 3 at different distances; (\textbf{j}--\textbf{l}) Real point clouds at different distances. }
\label{fig:sim_chessboard}
\end{figure}

Figure \ref{fig:sim_chessboard} shows some simulated point clouds for different noise and distance conditions. The~ last row shows the points from real data for comparison. By comparing the front view of the points clouds between the first row and the last row in Figure \ref{fig:sim_chessboard}, we can see that the points with baseline noise are similar to the real data. As for the side view of real data, we see that blue points and red points are generally distributed separately. This indicates that the noise of distance differs for different intensity for real measurements. However, as explained in Section \ref{subsubsec:cost_fun_corner}, since the proposed corners estimation method project all points onto the XOY plane, the noise along $z$-axis can be ignored. Thus, the difference between the simulated data and real data along $z$-axis will not affect the error evaluation of the proposed corner estimation method.

\subsubsection{Error Results from the Simulation}
\begin{figure}[h!]\centering
    \subcaptionbox{\label{fig:sim_error:noise}}{\includegraphics[width=0.49\textwidth]{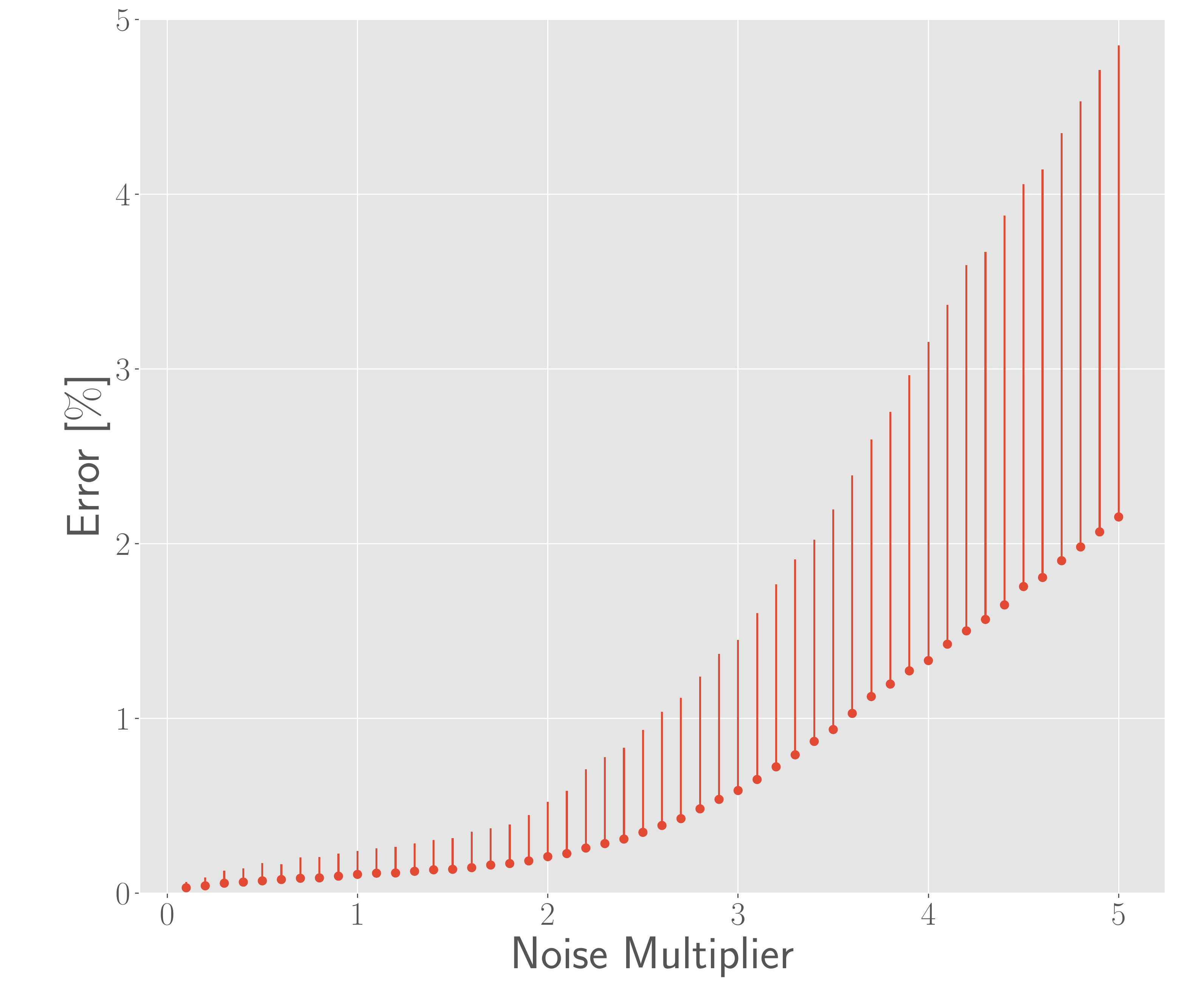}}
    \subcaptionbox{\label{fig:sim_error:dis}}{\includegraphics[width=0.49\textwidth]{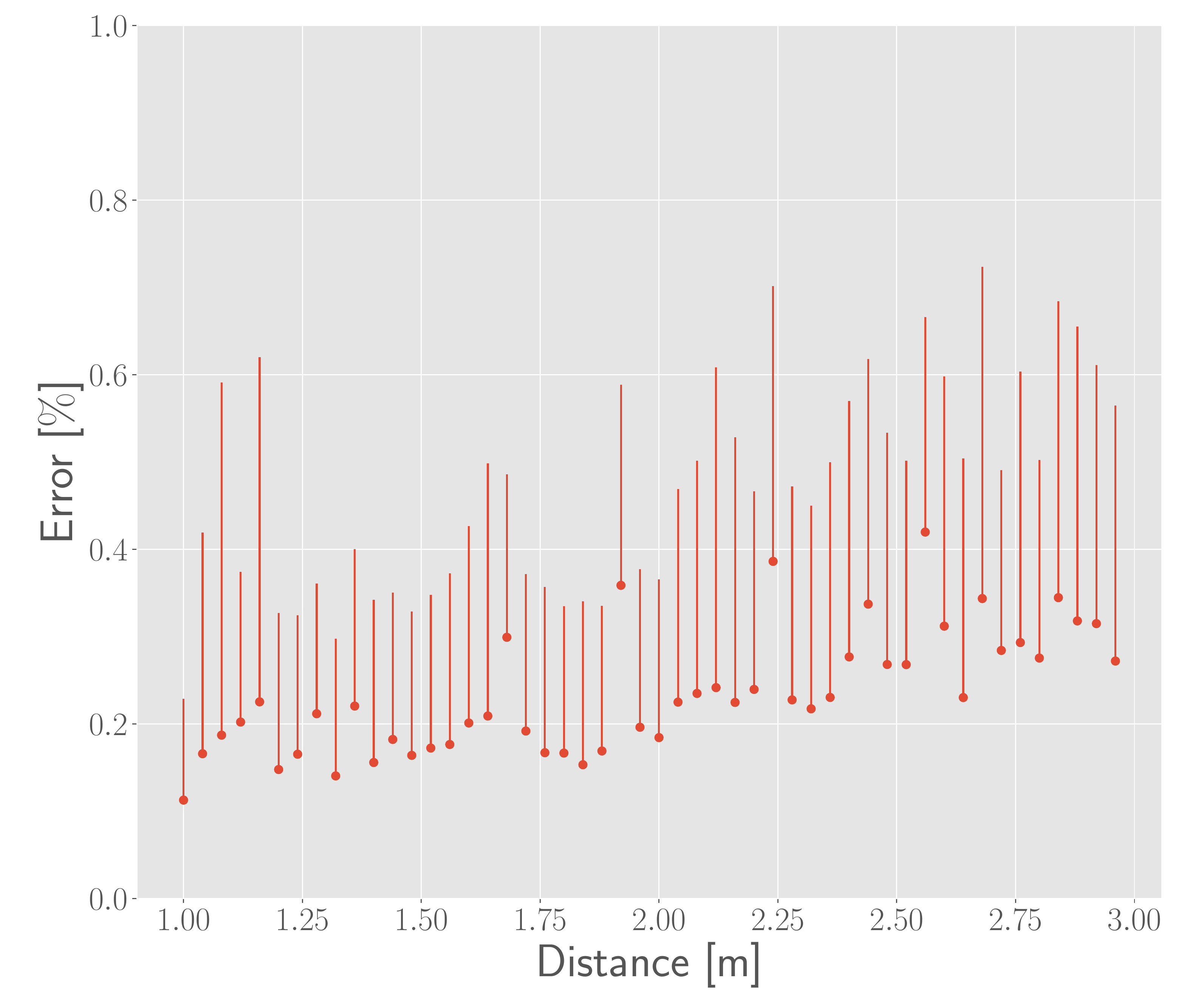}}  
  \caption{Corner detection error by simulation.  The horizontal axes represent different simulation conditions and the vertical axes represent the relative error. Red points represent the mean value and the vertical lines represent the $3\sigma$ range of the results simulated by 100 times at each simulation condition. (\textbf{a}) Relationship between the error and noise of the point cloud at 1 m. The $x$ axis represent the multiplier for the noise baseline; (\textbf{b}) Relationship between the error and the distance of the chessboard with the baseline noise.}
  \label{fig:sim_error}
\end{figure}

We applied the proposed method to simulated points with varying noise and distance conditions to estimate the coordinates of the point cloud. The distance error is calculated $\frac{\sqrt{\sum_{i=1}^{n}{\left | \hat{p}_i-p_i \right |}^2}}{n}$, where $\hat{p}_i$ depicts coordinates of the estimated corner and $p_i$ is the ground truth. Then the relative error is calculated with the unit of percentage by dividing the side length (7.5 cm  in this work) of a pattern in the chessboard. For each noise and distance condition, we repeat the simulation with different random seeds 100 times and calculate the average and standard deviation. Error results are shown in Figure \ref{fig:sim_error}. The horizontal axes represent the simulation conditions and the vertical axes represent the relative error. From Figure \ref{fig:sim_error}a, we can see that the error and uncertainty of the estimation increases drastically as the noise of measurement increases.  The influence of distance on the error is relatively lower than that of noise, as shown in Figure \ref{fig:sim_error}b. 

\subsection{Detected Corners}
\label{sec:sub:corners}
 \vspace{-6pt}

\subsubsection{From the Image}
\begin{figure}[h!]\centering
    \subcaptionbox{\label{fig:detected_corners:1}}{\includegraphics[height=3.8cm]{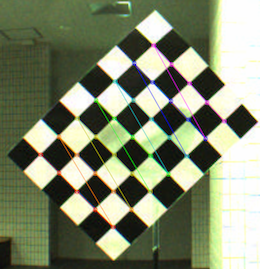}} 
    \subcaptionbox{\label{fig:detected_corners:2}}{\includegraphics[height=3.8cm]{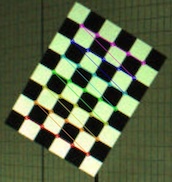}}
    \subcaptionbox{\label{fig:detected_corners:3}}{\includegraphics[height=3.8cm]{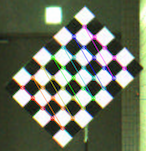}}
  \caption{{Detected corners} from the panoramic images. (\textbf{a}--\textbf{c}) Example results of detected corners from images with different poses and distances. }
  \label{fig:detected_corners}
\end{figure}

We use the method in \cite{Geiger_2012} to detect the corners in the panoramic image. Some example results are shown in Figure \ref{fig:detected_corners}. We can see that all corners are robustly detected. Compared to conventional vertex-based methods, it is difficult to detect vertices accurately and automatically when the background color is similar to the color of the planar board (lower left of the chessboard in Figure~\ref{fig:detected_corners}c). However, it can be accurately and automatically detected if we use the corners as feature points. For the correspondence with the corners from the point cloud, corners are counted starting from the lower left board.

\subsubsection{From the Point Cloud}
\begin{figure}[h!]\centering
    \subcaptionbox{\label{ested_grid_p9_all}}{\includegraphics[height=4cm]{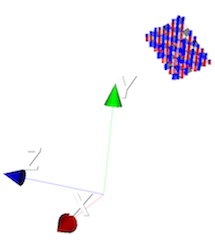}} 
    \hspace{0.3cm}
    \subcaptionbox{\label{ested_grid_p9_front}}{\includegraphics[height=4cm]{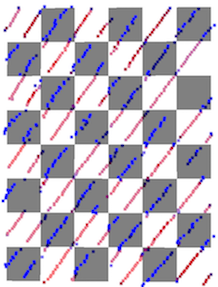}}
    \hspace{0.3cm}
    \subcaptionbox{\label{ested_grid_p9_side}}{\includegraphics[height=4cm]{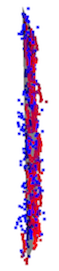}}
  \caption{Estimated corners of chessboard. (\textbf{a}) The fitted chessboard model of the point cloud in the real Velobug LiDAR coordinate system; (\textbf{b}) The front view of the zoom checkerboard; (\textbf{c}) The side view of the zoom checkerboard.}
  \label{ested_corners_fig}
\end{figure}

We applied the proposed corner detection method from the point cloud to real measured data. An example of the estimated corners is shown in Figure \ref{ested_corners_fig}. Figure \ref{ested_corners_fig}a shows the position of the chessboard's point cloud and the estimated chessboard model in the LiDAR coordinate system. The~zoomed-in image of the chessboard is shown in Figure \ref{ested_corners_fig}b. The corners of the chessboard are the estimated corners, which can be calculated with $\textit{M}_{XOY}^M, t_{XOY}^M$ in Section~ \ref{subsubsec:transform_xoy} and $T_r({\bm{\theta}}^M,\bm{t}^M)$ in  Section~ \ref{subsubsec:cost_fun_corner}. The counting order also starts from the lower left, which is the same as that in the corners from the correspondence with corners from the image. From Figure \ref{ested_corners_fig}b, we can see that most blue points (low intensity) are mapped on the black patterns, and red points (high intensity) are mapped on the white patterns. This means that the chessboard fits the points well.  

\subsection{Estimated Extrinsic Parameters}
\label{sec:sub:extrinsic_paras}

\begin{figure}[h!]\centering
    \subcaptionbox{\label{fig:x_ang}}{\includegraphics[width=0.34\textwidth]{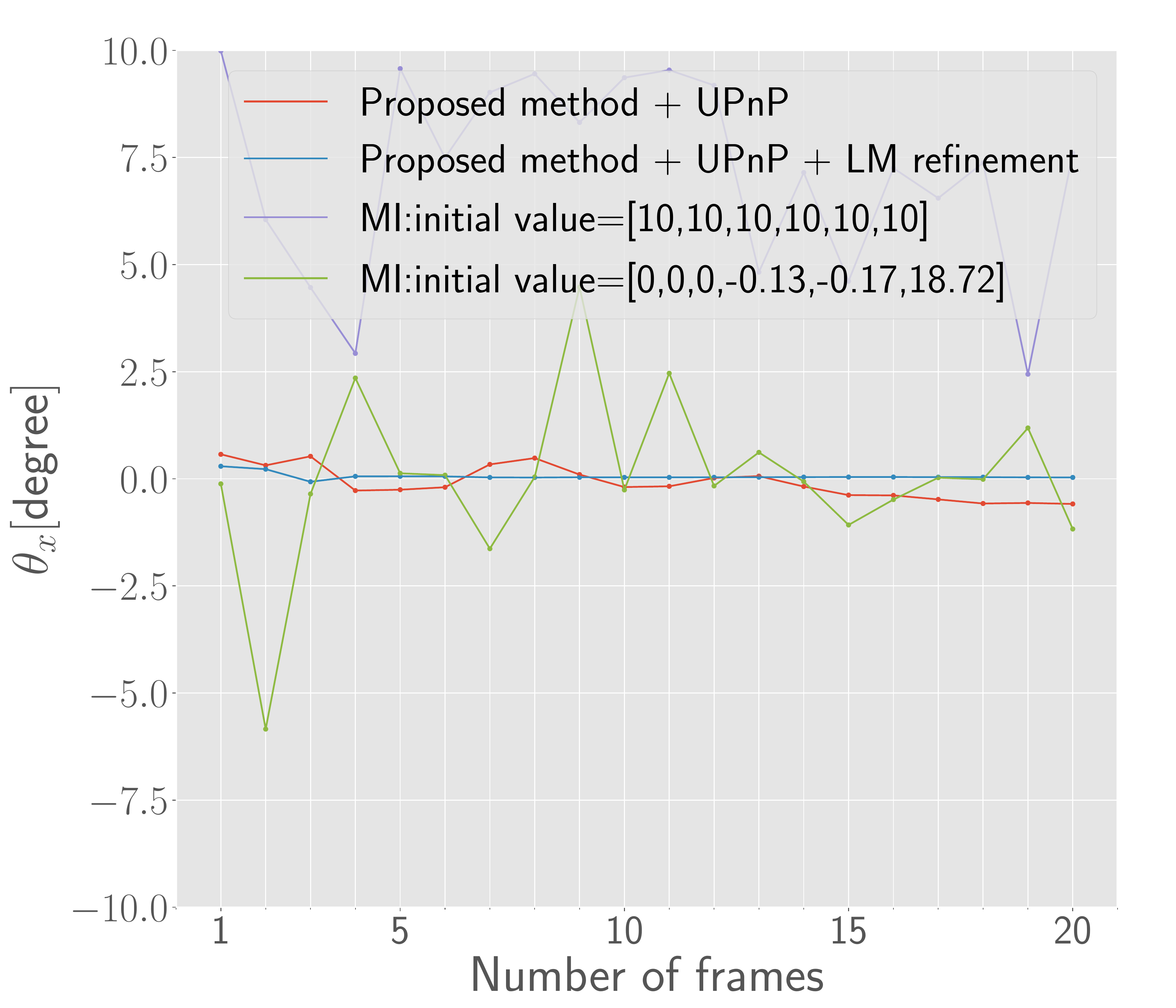}} \hspace{-0.6cm}
    \subcaptionbox{\label{fig:y_ang}}{\includegraphics[width=0.34\textwidth]{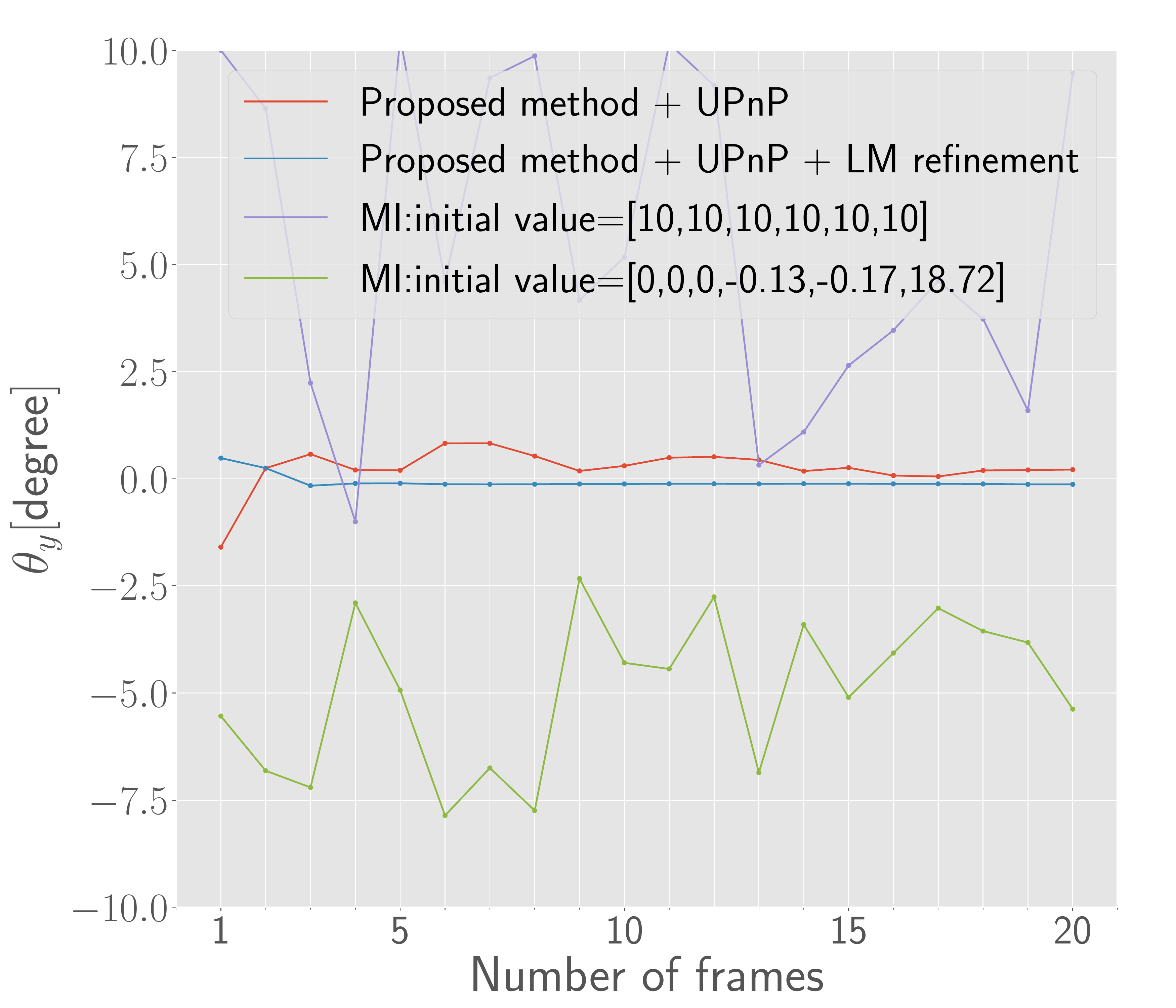}} \hspace{-0.6cm}
    \subcaptionbox{\label{fig:z_ang}}{\includegraphics[width=0.34\textwidth]{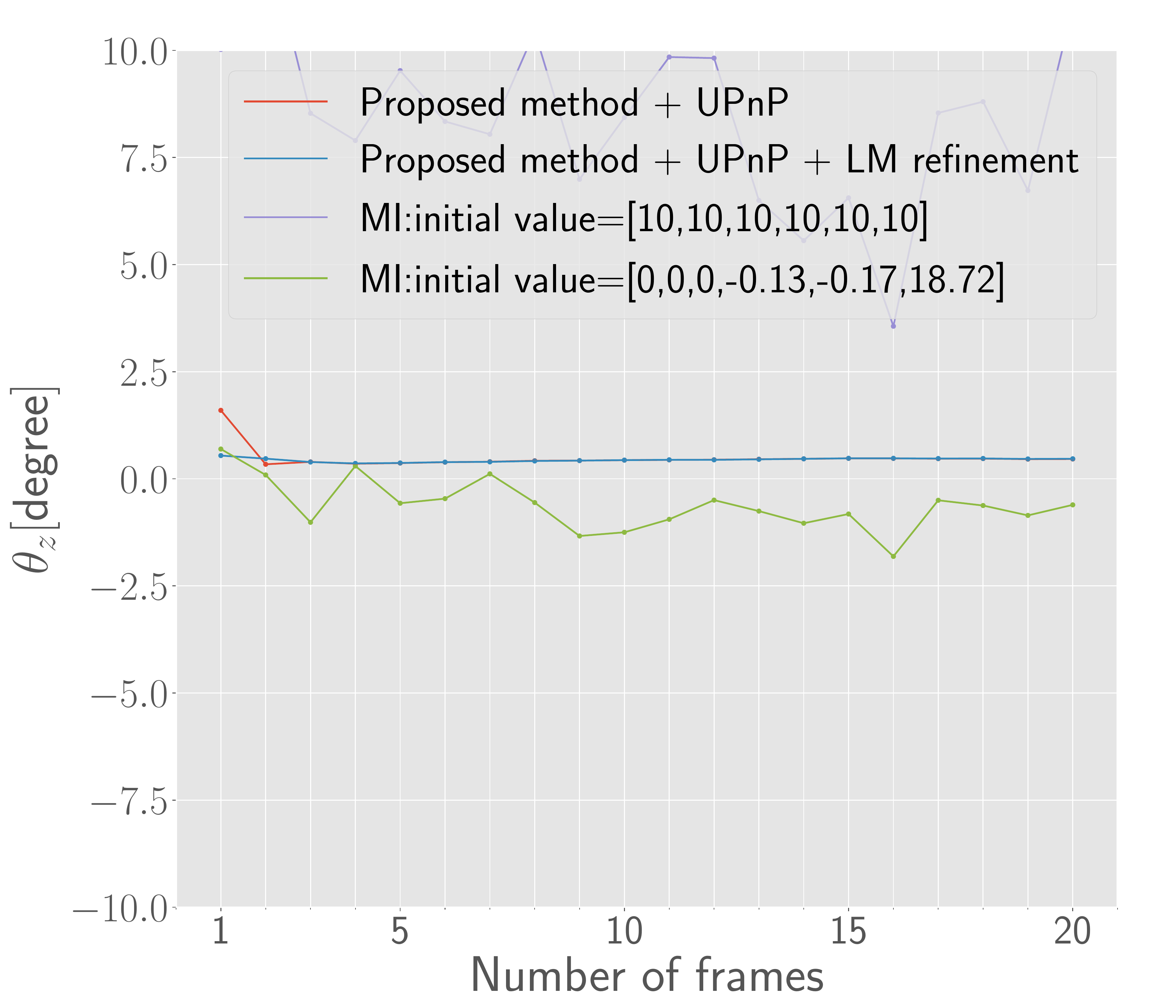}}\\
    \subcaptionbox{\label{fig:t_x}}{\includegraphics[width=0.34\textwidth]{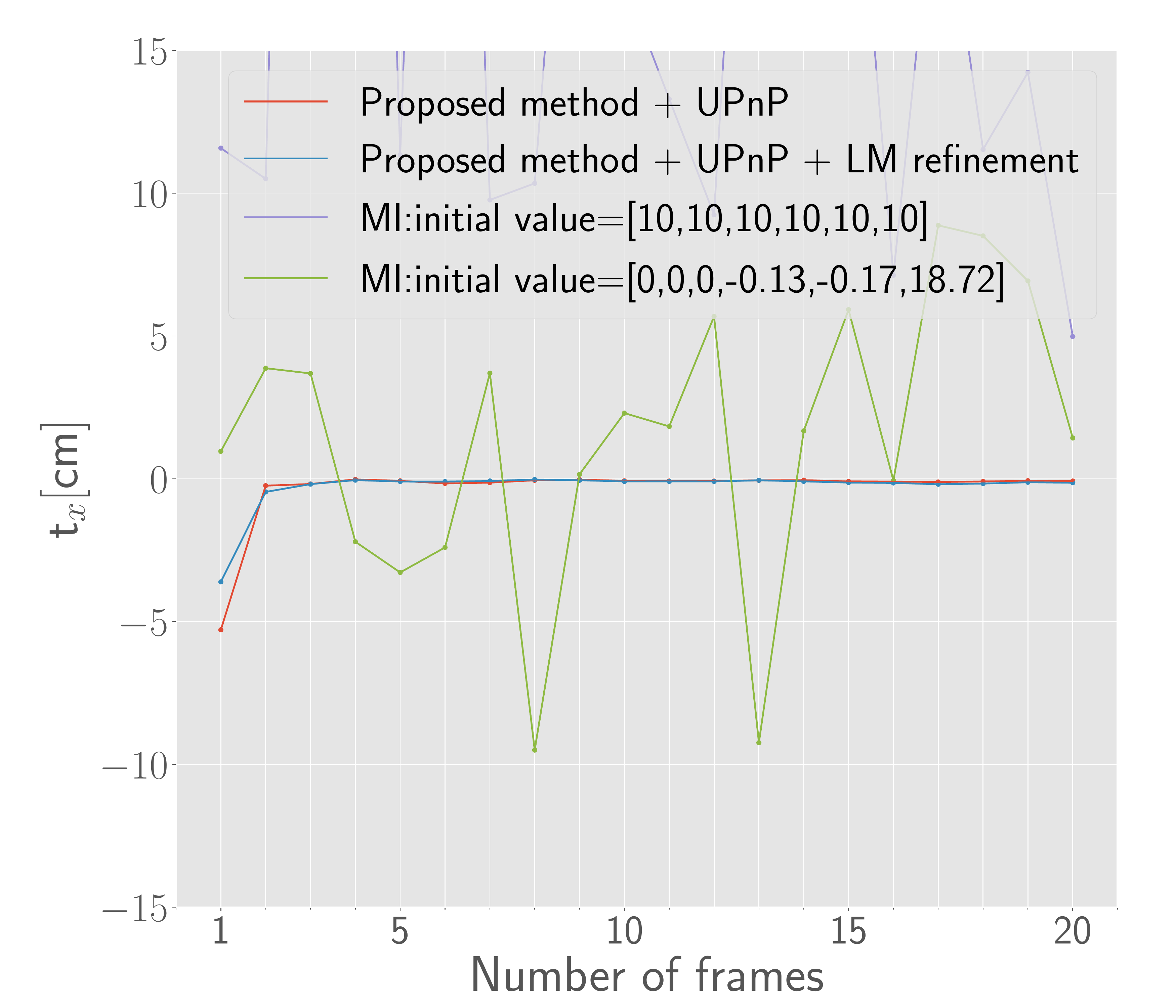}}  \hspace{-0.6cm}
    \subcaptionbox{\label{fig:t_y}}{\includegraphics[width=0.34\textwidth]{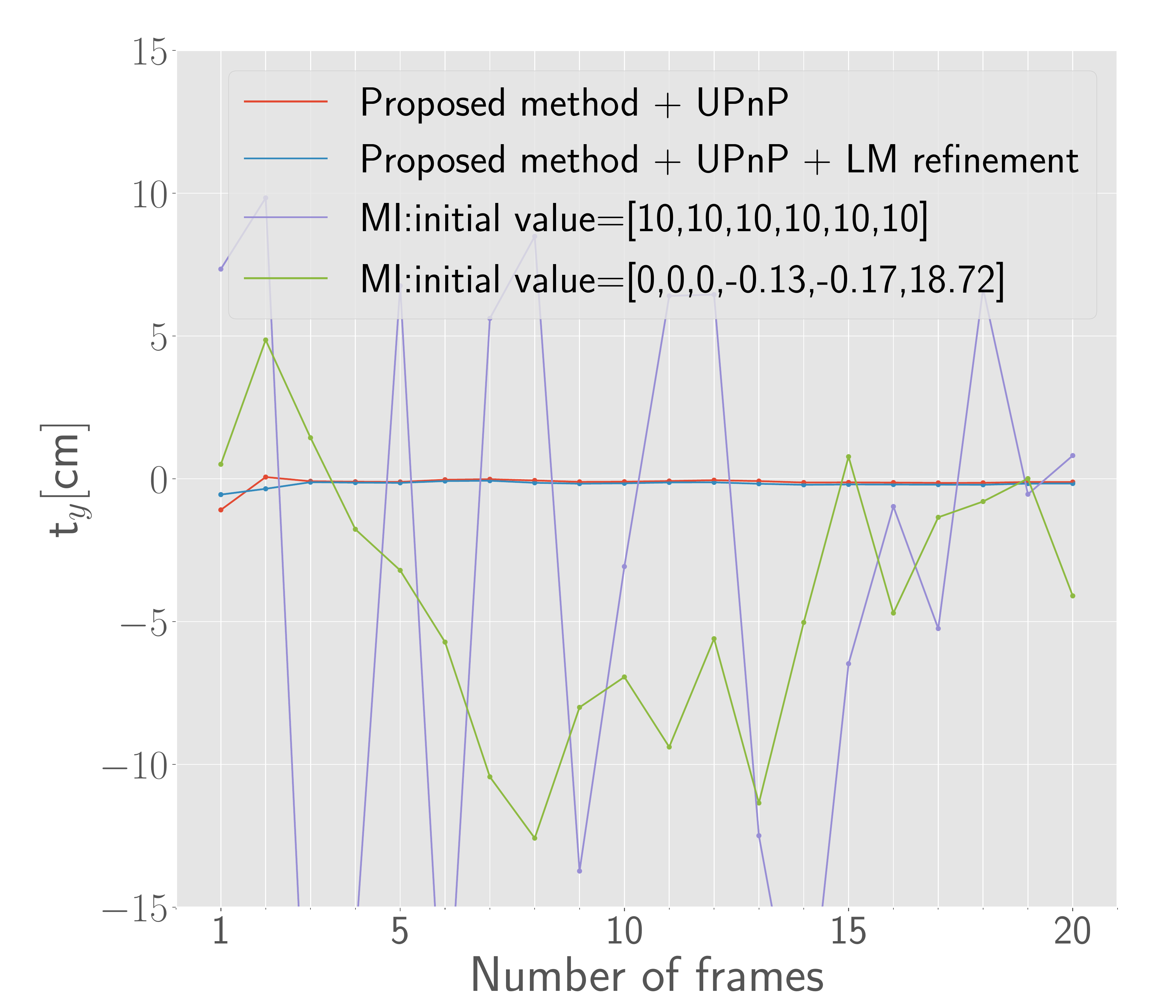}}\hspace{-0.6cm}
    \subcaptionbox{\label{fig:t_z}}{\includegraphics[width=0.34\textwidth]{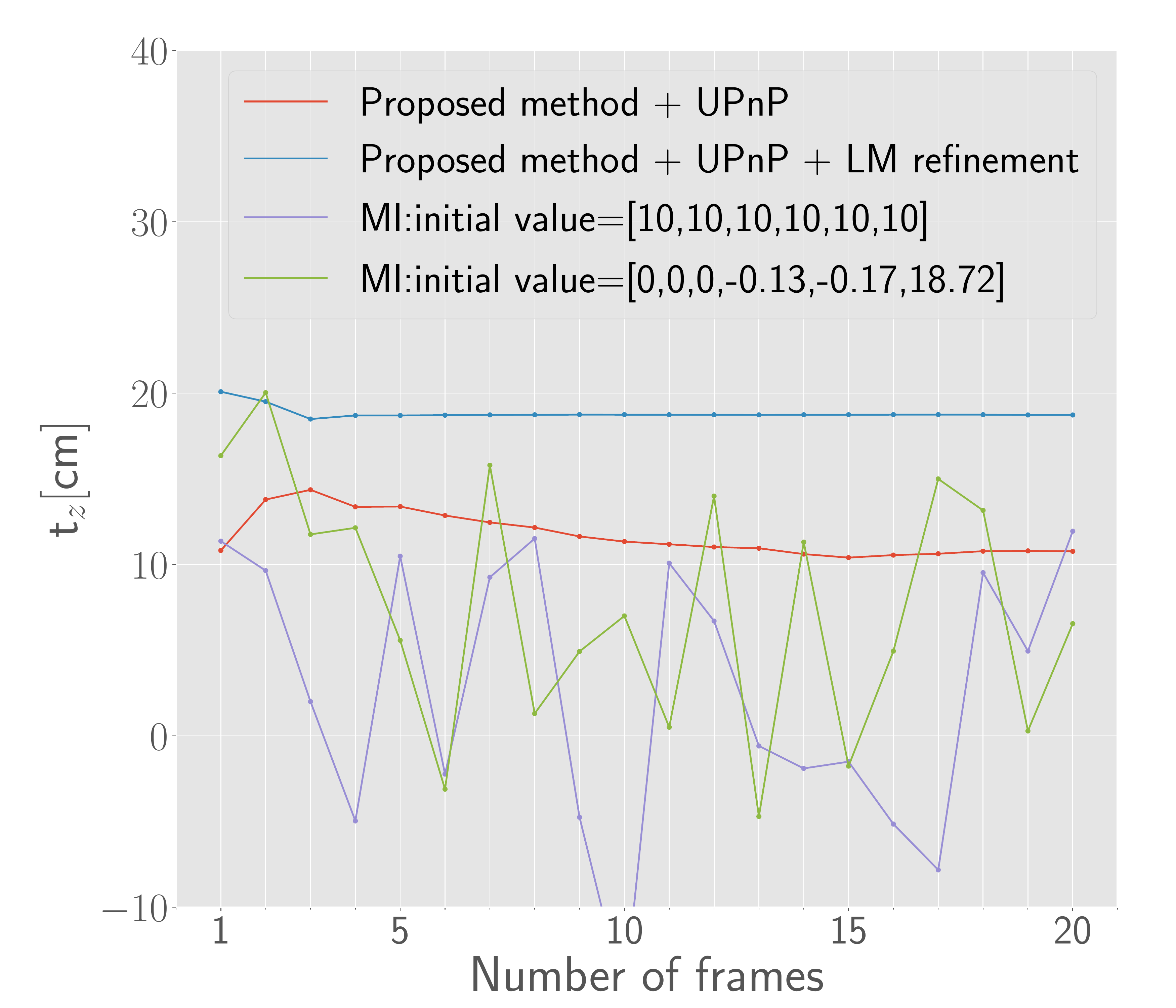}}  
    \caption{Estimated parameters by the proposed method and Pandey's Mutual Information (MI) method \cite{Pandey_2014} with different initial values as the numbers of frame increases. (\textbf{a}--\textbf{c}) Rotation angle along each axis; (\textbf{d}--\textbf{f}) Translation along each axis. }
    \label{fig:paras}
\end{figure}

The initial and refined transformation parameters with different numbers of frames are shown in Figure \ref{fig:paras}. As mentioned above, 20 frames are used for this evaluation. The $x$-axis represents the number of frames used for the estimation. For example, 5 indicates that the first five frames \mbox{frame\#1 $\sim$ frame\#5} are used for the extrinsic calibration. The frames are indexed as in Table \ref{tab:enumerate} to make the variation of the chessboard's point cloud increase in the horizontal direction as the number of frame increase for global optimization. The index of each camera is defined in \cite{ladybug3}.


Figure \ref{fig:paras} depicts the state from when the parameters start to become stable after the nonlinear refinement with more than four frames, while some parameters estimated by UPnP are erratic as the number of frames changes. Generally, there is no significant difference between the results obtained by UPnP and by nonlinear refinement. The translation value along the $z$-axis estimated by UPnP is 11.7~cm, which differs from 18.7~cm obtained using nonlinear refinement. We manually measure the offset between the LiDAR and the panoramic camera of the set. It is difficult to accurately find the center of each sensor, the range of the offset is supposed to be  in the range between 17 and 20~cm. Thus, the refined result is more accurate.

As a comparison, we also applied Pandey's online method \cite{Pandey_2014}, which also uses laser's reflectance intensity, to the data in this work. The results under different initial guesses are shown in Figure~\ref{fig:paras}. One initial guess is [10$^\circ$, 10$^\circ$, 10$^\circ$, 10 cm, 10 cm, 10 cm] and the other is [0$^\circ$, 0$^\circ$, 0$^\circ$, $-$0.13 cm, $-$0.17~cm, 18.72 cm]  which are almost the accurate parameters. From the comparison of the two methods, we~can conclude that the proposed method is more stable and accurate in terms of both rotation and translation. The main reason why Pandey's method performed not well with the data in the work is considered that the surrounding environment is almost static and the increment of the frames does not increase the variance of the data for the optimization.
However, as stated before, the original problem settings of two methods are different. Online calibration methods like Pandey's are very convenient and  can estimate an approximate results if a good initial value is provided while the proposed methods is target-based, but it can provide accurate and stable extrinsic calibration result.

\begin{table}[h!]
\centering
\begin{tabular}{|l|l|l|l|l|l|}
\hline
Camera Index & \multicolumn{1}{c|}{0} & \multicolumn{1}{c|}{1} & \multicolumn{1}{c|}{2} & \multicolumn{1}{c|}{3} & \multicolumn{1}{c|}{4} \\ \hline
Frame Index  & 1,6,11,16              & 2,7,12,17              & 3,8,13,18              & 4,9,14,19              & 5,10,15,20             \\ \hline
\end{tabular}
\caption{Relationship between the index of the frame and the index of the camera.}
\label{tab:enumerate}
\end{table}

\subsection{Re-Projection Error}
\label{sec:sub:reproj_err}

\label{subsec:back_proj_error}


To quantitatively evaluate two groups of  estimated parameters, we define a re-projection error metric based on intensity, which is similar to Equation \eqref{marker_cost_func} except that the error metric is defined on the image plane. To increase the confidence, only the points that are contained within the polygon of the detected corners on the images are counted. Moreover, the gray zone of intensity is increased for the same consideration and $\epsilon_g$ in Equation \eqref{eq:grayzone} is set to 4.

\begin{figure}[h!]
    \centering
    \begin{tabular}{cc}
    \adjustbox{valign=b}
        {\begin{tabular}{@{}c@{}}
            \subcaptionbox{\label{fig:errors:black_pattern}}{\includegraphics[width=0.25\textwidth]{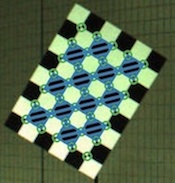}} \vspace{-0.2cm}\\
            \subcaptionbox{\label{fig:erros:white_pattern}}{\includegraphics[width=0.25\textwidth]{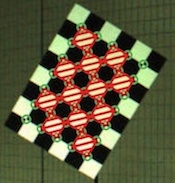}}
        \end{tabular}}
    &
    \adjustbox{valign=b}
        {\subcaptionbox{\label{fig:errros:res}}{\includegraphics[width=0.55\textwidth]{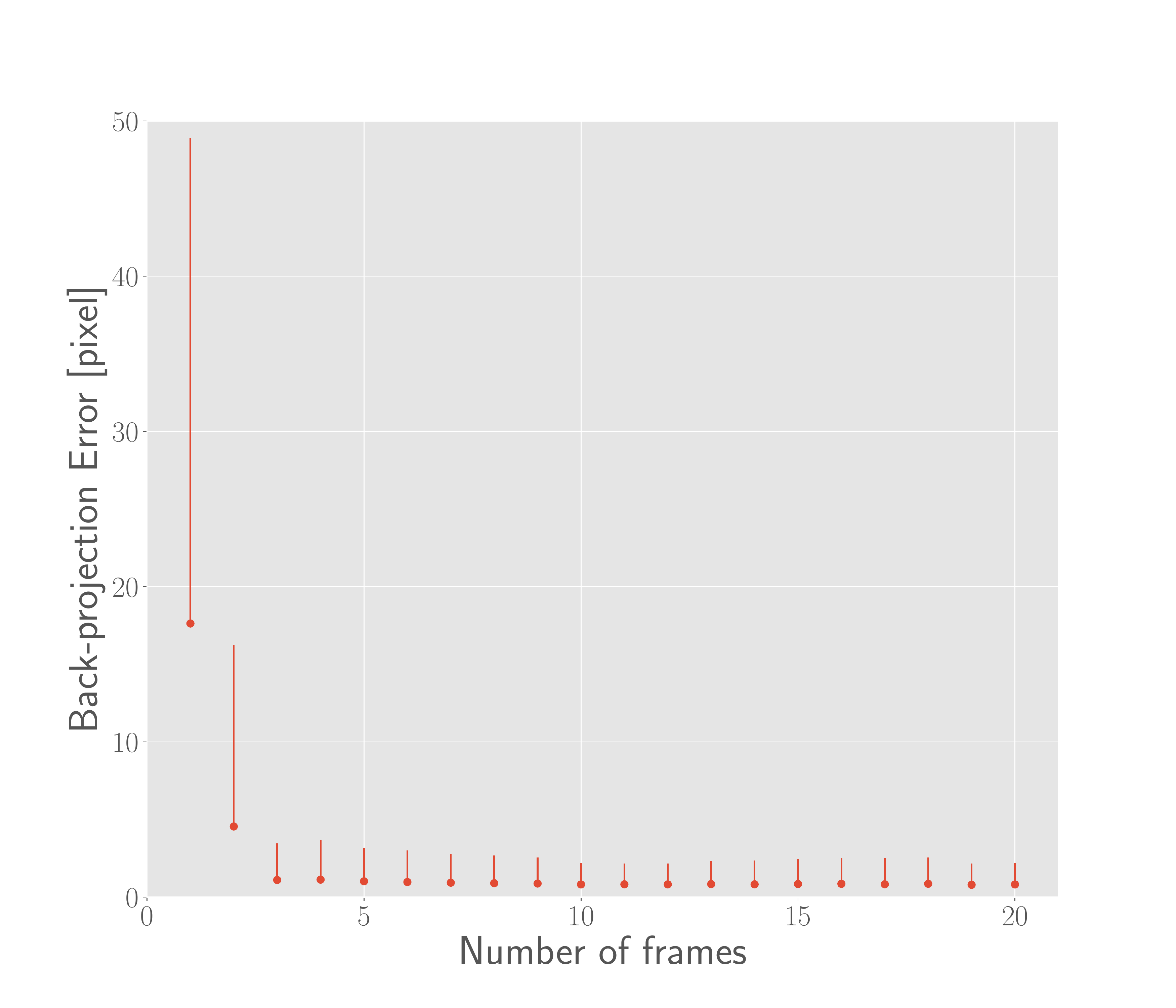}}}\\
    \end{tabular}
    \caption{Re-projection error calculation and results. (\textbf{a},\textbf{b}) Shaded quadrilaterals show the regions of black and white patterns respectively. Points mapped into these regions are counted for error calculation; (\textbf{c}) The errors for parameters estimated by different numbers of frames. The point and vertical line represent the mean and $3\sigma$ range of all errors calculated by applying the estimated parameters to the different number of frames.}
    \label{fig:errors}
\end{figure}

For the error calculation, we first transform the point cloud of the chessboard with estimated extrinsic parameters. After mapping all 3D points of the transformed chessboard point cloud at the panoramic image based on an equirectangular projection model, we calculate the error on the chessboard plane of the panoramic image. Only the point that is mapped within the quadrilateral region constructed by the four detected corners on the image is counted, as shown in Figure \ref{fig:errors}a,b. As the gray zone is identified, we can classify the mapped points into white or black colors. If the estimated color of a mapped point differs from the color of the quadrilateral it falls in, the error of this point is calculated using Equation \eqref{fd1}.

 In the measurement, the farther the chessboard is, the smaller the occupied pixels of chessboard on the image will be. Because the re-projection error is measured with absolute pixel, the error of farther chessboard will be relatively smaller. To ensure that the error metric is unaffected by distance, we normalize the error by multiplying the value of the distance with the consideration that the size of an object on the image is generally proportional to the reciprocal of the distance. Moreover, the~``attendance rate'' of the point, which indicates the rate at which points are counted for error calculation, is also taken into account for the overall re-projection performance.  The final formula is defined in~ Equation \eqref{eq:error_calc}
\begin{equation}\label{eq:error_calc}
    e=\frac{C_m }{N_{c}}\cdot r^M\cdot\frac{ P_{c}  N_{a}}{P_{a}{N_{c}}}
\end{equation}

Here, $C_m$ is the cost function defined in Equation \eqref{marker_cost_func}, $r_M$ is the Euclidean distance of the chessboard, $P_a$ is the total number of the patterns in the chessboard, $P_c$ is the  number of the quadrilaterals constructed by detected corners for calculating the re-projection error, $N_{a}$ is the points number of chessboard's point cloud, $N_{c}$ is the number of points that fall into the quadrilaterals constructed by detected corners, $r^M$ is a factor of the error caused by the distance and $\frac{ P_{c}  N_{a}}{P_{a}N_{c}}$ is the penalty factor by dividing  the ``attendance rate''. The ``attendance rate'' is derived from $\frac{N_{c}}{P_{c}}\div\frac{N_{a}}{P_{a}}$, which is the ratio of  the average number of point in each pattern of the region (Figure \ref{fig:errors}a,b) only counted for error calculation and the average number of point in each pattern of the whole chessboard. 

The re-projection error with the metric defined above is shown in Figure \ref{fig:errors}c. The point and vertical line represent the mean and $3\sigma$ range of all errors calculated by applying the estimated parameters to 20 frames. The length of one grid of the chessboard in the 4000 $\times$ 8000 panoramic image at 1 m is approximately 100 pixels.  The back projection error is 0.8 pixel when the optimization begins to converge after 4 frames, and the relative error to the side length is 0.8\%.
\subsection{Re-Projection Results}
\label{sec:sub:reproj_res}

\begin{figure}[h!]\centering

            \includegraphics[width=0.24\textwidth]{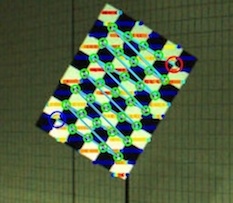}
            \includegraphics[width=0.24\textwidth]{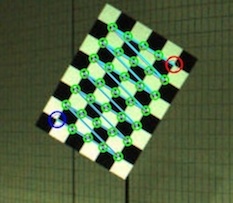}
            \includegraphics[width=0.24\textwidth]{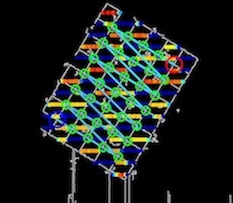}
            \includegraphics[width=0.24\textwidth]{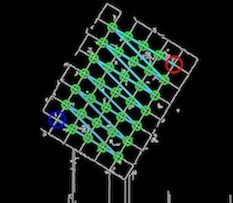}
\caption{Re-projected corners and points of the chessboard (best viewed when zoomed in). Big green circles and cyan lines indicate the detected corners. Small pink circles in big green circles and  pink lines indicate re-projected corners estimated from the point cloud. Big blue and red circles represent the start and end for counting the corners. Blue points indicate low reflectance intensity and red points indicate high reflectance intensity.}
\label{fig:back_proj_obj:corner}
\end{figure}

For qualitative evaluation, we apply the final extrinsic transformation matrix estimated by the proposed method to rotate point cloud and project it to the image.  The result of re-projected corners is shown in Figure \ref{fig:back_proj_obj:corner}. The large green circles and cyan lines indicate the detected corners. The small pink circles in large green circles and lines indicate re-projected corners estimated from the point cloud. The large blue and red circles represent the start and end for counting the corners. We can see that the re-projected corners estimated from the point cloud almost coincide with the corners detected in the image.

\begin{figure}[h!]\centering
    \subcaptionbox{\label{p9_back_proj_intens}}{\includegraphics[width=0.8\textwidth]{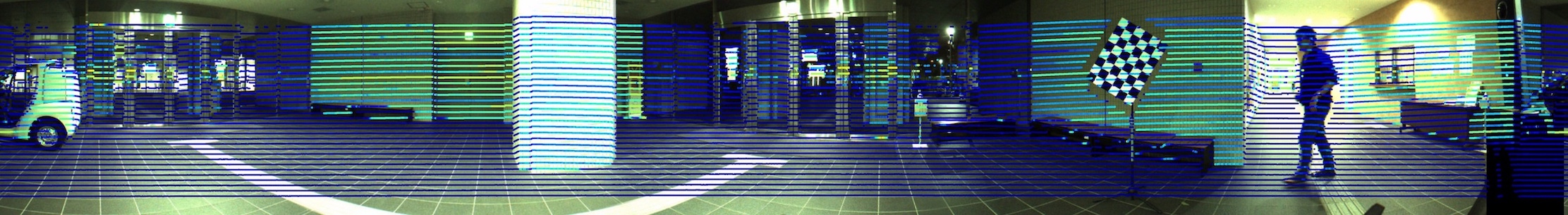}}\\
    \subcaptionbox{\label{p9_back_proj_dis}}{\includegraphics[width=0.8\textwidth]{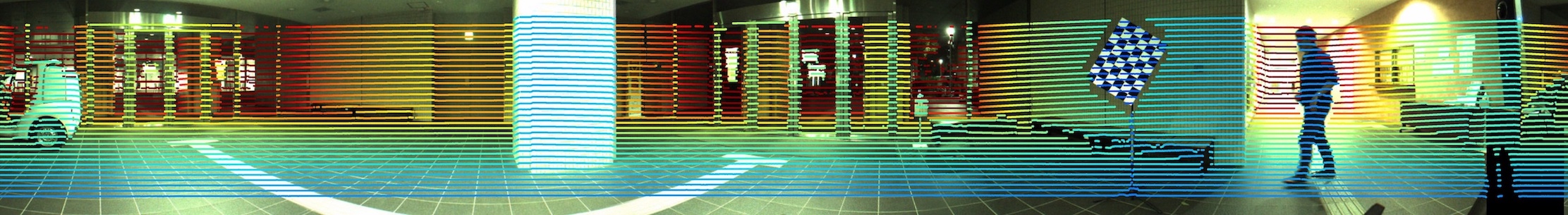}}\\
    \subcaptionbox{\label{back_proj_intens_edge}}{\includegraphics[width=0.8\textwidth]{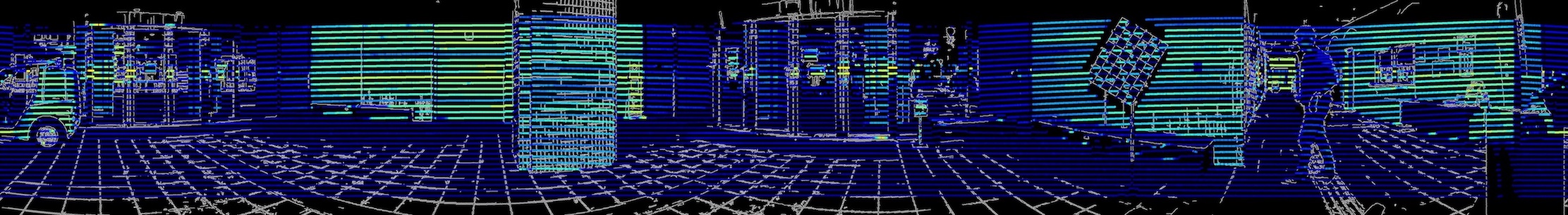}}\\
    \subcaptionbox{\label{back_proj_dis_edge}}{\includegraphics[width=0.8\textwidth]{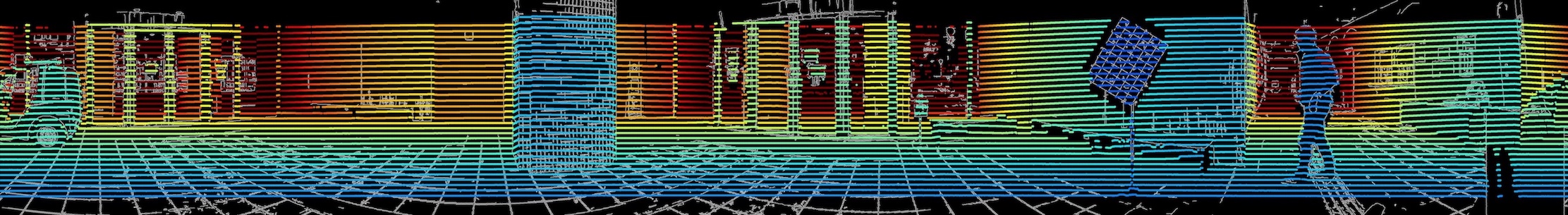}}\\
  \caption{Re-projection results. (\textbf{a}) All re-projected points on the color panoramic image and colored by  intensity;  (\textbf{b}) All re-projected  points on the color panoramic image and colored by distance; (\textbf{c})~Re-projected result on edge extracted image of all points colored by intensity; (\textbf{d}) Re-projected result on edge extracted image of all points  colored by distance. }
\label{backproj_fig}
\end{figure}

\begin{figure}[h!]\centering
    \mbox{
        \begin{subfigure}[h!]{1\textwidth}
        \centering
            \includegraphics[width=0.23\textwidth]{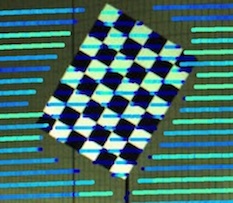}
            \includegraphics[width=0.23\textwidth]{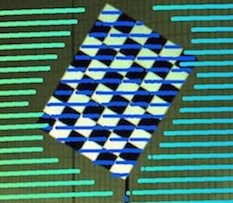}
            \includegraphics[width=0.23\textwidth]{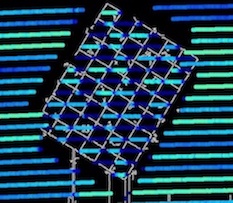}
            \includegraphics[width=0.23\textwidth]{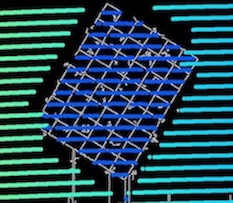}
        \caption{}
        \label{fig:back_proj_obj:marker}
        \end{subfigure}
    }
    \vspace{0.1cm}
    \mbox{
        \begin{subfigure}[h!]{1\textwidth}
        \centering
            \includegraphics[width=0.23\textwidth]{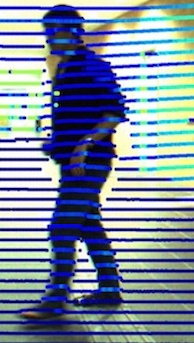}
            \includegraphics[width=0.23\textwidth]{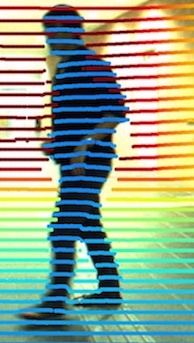}
            \includegraphics[width=0.23\textwidth]{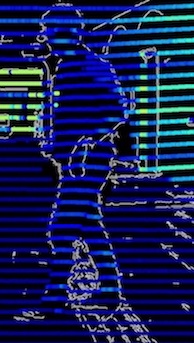}
            \includegraphics[width=0.23\textwidth]{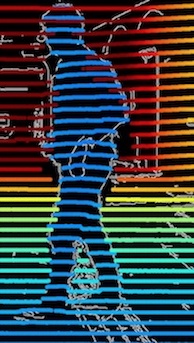}
        \caption{}
        \label{fig:back_proj_obj:me}
        \end{subfigure}
    }
    \vspace{0.1cm}
    \mbox{
        \begin{subfigure}[h!]{1\textwidth}
        \centering
            \includegraphics[width=0.23\textwidth]{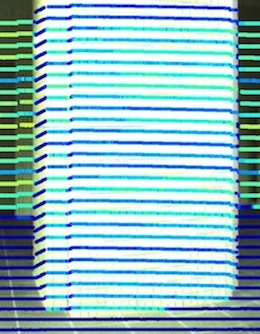}
            \includegraphics[width=0.23\textwidth]{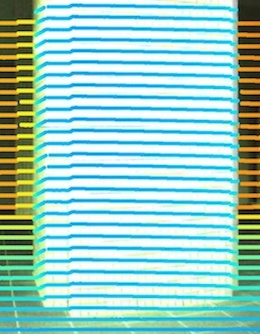}
            \includegraphics[width=0.23\textwidth]{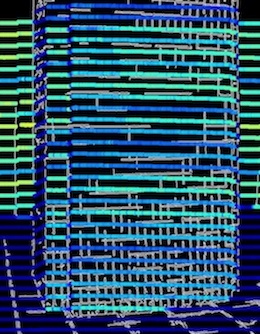}
            \includegraphics[width=0.23\textwidth]{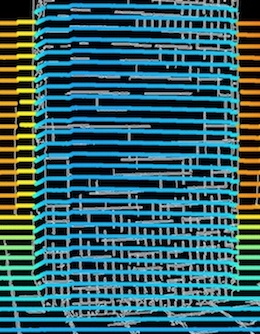}
        \caption{}
        \label{fig:back_proj_obj:pillar}
        \end{subfigure}
    }
    \vspace{0.1cm}
    \mbox{
        \begin{subfigure}[h!]{1\textwidth}
        \centering
            \includegraphics[width=0.23\textwidth]{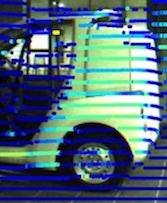}
            \includegraphics[width=0.23\textwidth]{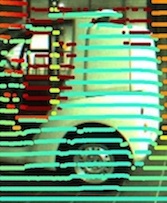}
            \includegraphics[width=0.23\textwidth]{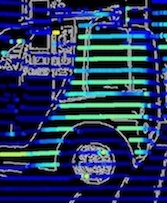}
            \includegraphics[width=0.23\textwidth]{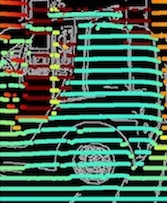}
        \caption{}
        \label{fig:back_proj_obj:car}
        \end{subfigure}
    }
\caption{{Zoomed details of} re-projected points. (\textbf{a}--\textbf{d}) Re-projected results on chessboard, human, pillar and car respectively. Each column represents re-projected points colored by intensity and distance on original RGB images and edges-extracted images. Blue indicates low value and red indicates high value. }
\label{fig:back_proj_obj}
\end{figure}

\begin{figure}[h!]\centering
    \subcaptionbox{\label{colored_pcd}}{\includegraphics[height=3.2cm]{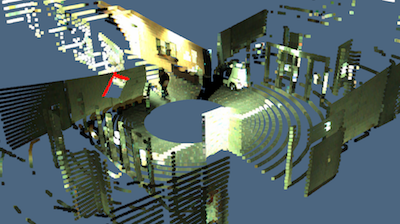}}
    \subcaptionbox{\label{colored_pcd_zoomed_1}}{\includegraphics[height=3.2cm]{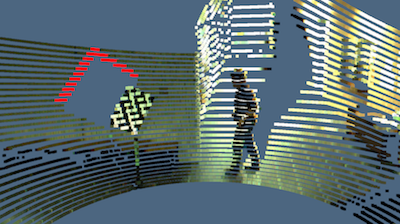}}
    \subcaptionbox{\label{colored_pcd_zoomed_1}}{\includegraphics[height=3.2cm]{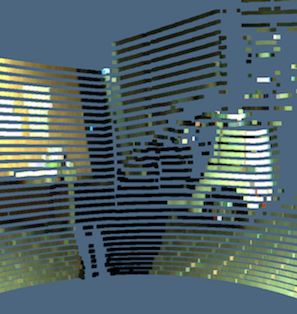}}
  \caption{Projection of the RGB information from the image to the point cloud with the estimated extrinsic parameters. (\textbf{a}) An example of the colored point cloud;  (\textbf{b}) The zoomed view of the chessboard in (\textbf{a}). The red points in (\textbf{a},\textbf{b}) are occluded region caused by the chessboard; (\textbf{c}) The zoomed view of the car in (\textbf{a}). }
  \label{fig:colored_pcd}
\end{figure}

Re-projection results of all points within one frame are as listed in Figures \ref{backproj_fig}--\ref{fig:colored_pcd}. For better visualization, points colored according to the intensity and distance are mapped to the original RGB images and the edges-extracted images, respectively. Figure \ref{backproj_fig} shows the overall results on the panoramic image. Details of individual objects are shown in Figure \ref{fig:back_proj_obj}. We can see that the bounds generated by the change of re-projected 3D points with different intensities and distance fit the edges of the image exactly, for example the fourth images in Figure \ref{fig:back_proj_obj}a,b. Some inconsistencies occur due to the occlusion, for example the lower left and upper right parts of the chessboard in Figure \ref{fig:back_proj_obj}a and the upper part of the human in Figure \ref{fig:back_proj_obj}b. The final colored point cloud is shown in Figure \ref{fig:colored_pcd}. Red points indicate the region occluded by the chessboard.


\section{Discussions}
\label{sec:dis}
\begin{itemize}
\item \textit{Automatic segmentation.}  
As the first step of the proposed method, automatic segmentation is performed. The current segmentation method is only based on the distance information, which needs the chessboard to be spatially separated from the surrounding objects. Nevertheless, slight under-segmentation caused by the stand of the chessboard or over-segmentation caused by the measurement noise may still occur. The degree of mis-segmentation generated by the segmentation method used in this work is experimentally shown to be negligible for the corners estimation with the overall optimization of the proposed method.

\item \textit{Simulation.}
To evaluate the performance for the corner estimation with the proposed method, we approximately simulated the points by considering the probability model of the distance as Gaussian distribution. However, the probability model for the noise of reflectance intensity, which is an aspect for corners estimation, is not considered. Under the influence of reflectance intensity, the real error of corner estimation is supposed to be higher than the simulated results in this work. This is one of the reasons why the relative error for corners estimation is about 0.2\%, as~shown in Figure \ref{fig:sim_error}b, and the final re-projection error increased to 0.8\% in Section \ref{subsec:back_proj_error}.
For a more precise simulation, the probability model of the reflectance value related to the incidence angle, the distance and the divergence of laser beam needs to be formulated.

\item \textit{Chessboard.}
As shown in Figure \ref{fig:dis_vs_ang}, both the horizontal and vertical intervals increase as the distance increases. To gather enough information for corner estimation, the side length of one grid in the chessboard is suggested to be greater than 1.5 times of the theoretical vertical interval at the farthest place. In addition, the intersection angle between the diagonal line of the chessboard and the $z$-axis of the LiDAR is suggested to be less than $15^\circ$ to enable the scanning of as many patterns as possible.

We use the panoramic image for calibration, therefore, to remain unaffected by the stitching error, it is better to place the chessboard in the center of the field of view for each camera.

\item \textit{Correspondence of 3D and 2D corners.}
In this work, a chessboard with 6$\sim$8 patterns is used and the counting order is defined as starting from the of the chessboard for automatic correspondence. To make the ``lower left'' identified correctly, the chessboard should be captured to make the ``lower left'' of the real chessboard be same with that of chessboard in the image during the data acquisition. Also, the direction of $z$-axis of the two sensors should be almost consistent shown as in Figure \ref{setup}b. However, these restrictions can be released with the introduction of asymmetrical patterns in practical use.
\end{itemize}

\section{Conclusions and Future Works}
\label{sec:con}
In this work, we proposed a novel and fully automatic reflectance intensity based calibration method for LiDAR and panoramic camera system. Compared to existing methods, we make use of the corners information of the chessboard's point cloud instead of the edges information. 
Corners of the sparse and noisy chessboard's point cloud are estimated by solving the optimization problem with the intensity information. After the correspondence of corners both on the image and in the point cloud, the extrinsic transformation matrix is generated by solving the nonlinear optimization problem. To evaluate the performance of the corner estimation from the point cloud, we simulated the points and compared the estimated corners with the ground truth. We applied the proposed method to an equipment set consisting of a Velodyne LiDAR sensor and a Ladybug3 panoramic camera. Quantitative evaluation of the accuracy of the proposed method was performed with a proposed intensity-based re-projection error metric. The re-projected results were  verified by qualitative evaluation. The Python implementation of the proposed method can be downloaded from \href{https://github.com/mfxox/ILCC}{https://github.com/mfxox/ILCC}. 

In this work we evaluate the proposed method only with the Velodyne LiDAR HDL-32e sensor, we will evaluate on more types of sensor in the future. In practical 3D color mapping applications with the fusion of two sensor modalities , occlusion due to the difference in views must be addressed in the future. In addition, we aim to implement an online calibration without any information about the correct matrix based on the segmentation of point cloud according to both scanline discontinuity and reflectance intensity. 

\vspace{6pt}



\bibliographystyle{ieeetr}
\bibliography{journal2}
\end{document}